Electronics and Computer Science
Faculty of Physical Sciences and Engineering
University of Southampton

Jakub J. Dylag
Victor Suarez
James Wald
Aneesha Amodini Uvaraj

22/11/2022

Automatic Geo-alignment of Artwork in Children's Story Books

Project supervisor: Shoaib Jameel
Second examiner: Daniel Clark

A Group Design Project Report submitted for the Award of

Master of Computer Science and Artificial Intelligence

Master of Computer Science

Master of Electrical and Electronic Engineering

Master of Electrical and Electronic Engineering

# Abstract

This report explores the work carried out by Group #11 for the 4th year group design project: Automatic Geo-alignment of Artwork in Children's Story Books. A study was conducted to prove AI software could be used to translate and generate illustrations without any human intervention. This was done with the purpose of showing and distributing it to the external customer, Pratham Books. The project aligns with the company's vision by leveraging the generalisation and scalability of Machine Learning algorithms; offering significant cost-efficiency increases to a wide range of literary audiences in varied geographical locations. A comparative study methodology was utilised to determine the best performant method out of the 3 devised: Prompt Augmentation using Keywords, CLIP Embedding Mask, and Cross Attention Control with Editorial Prompts. A thorough evaluation process was completed using both quantitative and qualitative measures. Each method had its own strengths and weaknesses, but through the evaluation, method 1 was found to have the best yielding results. Promising future advancements may be made to further increase image quality by incorporating Large Language Models and personalised stylistic models. The presented approach can also be adapted to Video and 3D sculpture generation for novel illustrations in digital webbooks.

# Contents

# List of Abbreviations

| | |
|---|---|
| AI | Artificial Intelligence |
| APM | Agile Project Management |
| CFG | Classifier-free Diffusion Guidance |
| CLIP | Contrastive Language–Image Pre-training |
| DDIM | Denoising Diffusion Implicit Models |
| DDPM | Denoising diffusion probabilistic models |
| FID | Frechet Inception Distance |
| GAN | General Adversarial Networks |
| LDM | Latent Diffusion Models |
| NeRF | Neural Radiance Fields |
| POS | Parts of Speech |
| RAIL | Responsible use of AI License |
| RfP | Reading For Pleasure |
| SD | Stable Diffusion |

# Acknowledgements

To our computers, who put up with all this.

**Statement of Originality**

- I have read and understood the ECS Academic Integrity information and the University's Academic Integrity Guidance for Students.
- I am aware that failure to act in accordance with the Regulations Governing Academic Integrity may lead to the imposition of penalties which, for the most serious cases, may include termination of programme.
- I consent to the University copying and distributing any or all of my work in any form and using third parties (who may be based outside the EU/EEA) to verify whether my work contains plagiarised material, and for quality assurance purposes.

*You must change the statements in the boxes if you do not agree with them.*

We expect you to acknowledge all sources of information (e.g., ideas, algorithms, data) using citations. You must also put quotation marks around any sections of text that you have copied without paraphrasing. If any figures or tables have been taken or modified from another source, you must explain this in the caption and cite the original source.

**I have acknowledged all sources, and identified any content taken from elsewhere.**

If you have used any code (e.g., open-source code), reference designs, or similar resources that have been produced by anyone else, you must list them in the box below. In the report, you must explain what was used and how it relates to the work you have done.

**I have not used any resources produced by anyone else.**

You can consult with module teaching staff/demonstrators, but you should not show anyone else your work (this includes uploading your work to publicly accessible repositories e.g., Github, unless expressly permitted by the module leader), or help them to do theirs. For individual assignments, we expect you to work on your own. For group assignments, we expect that you work only with your allocated group. You must get permission in writing from the module teaching staff before you seek outside assistance, e.g., a proofreading service, and declare it here.

**I did all the work myself, or with my allocated group, and have not helped anyone else.**

We expect that you have not fabricated, modified or distorted any data, evidence, references, experimental results, or other material used or presented in the report. You must clearly describe your experiments and how the results were obtained, and include all data, source code and/or designs (either in the report or submitted as a separate file) so that your results could be reproduced.

**The material in the report is genuine, and I have included all my data/code/designs.**

We expect that you have not previously submitted any part of this work for another assessment. You must get permission in writing from the module teaching staff before re-using any of your previously submitted work for this assessment.

**I have not submitted any part of this work for another assessment.**

If your work involved research/studies (including surveys) on human participants, their cells or data, or on animals, you must have been granted ethical approval before the work was carried out, and any experiments must have followed these requirements. You must give details of this in the report and list the ethical approval reference number(s) in the box below.

**My work did not involve human participants, their cells or data, or animals.**

*ECS Statement of Originality Template, updated August 2018, Alex Weddell aiofficer@ecs.soton.ac.uk*

# 1 Introduction

Pictures have played a vital role in human communications. Illustrations within story books visualise the presented narrative, regardless of textual confusion, which in this case might aid readers with limited literacy in their understanding of the story [1]. In a study done by Stewig in 1972, children in Grades 1 to 3 were approached on what kind of books they liked, with them saying that they preferred books in which 25% of the space was covered in pictures . It can be said that seeing these illustrations can peak child interest, making them more likely to pick up and read the book. By producing reading-for-pleasure books (RfP), which in a definition by the National Literacy Trust, defines as reading as per our own free will, in order to obtain the satisfaction, in storybooks where the pictures and text are intertwined, can improve their reading ability. By having access to relevant RfP books and other reading material, the child is more likely to read more, and in doing so, they strengthen their attitude towards reading, which further increases their comprehension skills and improves their grammar, as well as their general knowledge[2].

Integrating their own cultural practices or even pictures from aspects of their own heritage are important for children, as this helps shape and solidify their cultural identity once they see themselves represented in the books they read [3]. This can be as simple as seeing characters that look alike to their ethnicity, wearing traditional clothing, or even familiar inanimate objects that they see in their everyday life. As stated by Sugen, this is quite important to those in colonised countries, where some of these traditions can be lost [4]. In an article published in 2015 about a Jordanian writer, Taghreed Najjar, she mentions that when her kids read books sourced from other countries, they were unable to relate to clothes and houses the characters wore, and lived in. This also led them unable to practice their mother tongue, and ultimately led to them becoming less likely to read as a whole [5].

A challenge for illustrators is how to quickly create relevant engaging artwork to make content creation efficient. Inspired by the concept of geographically (geo)-localised web search results, one possible way is to automatically geo-align the RfP book artwork. For instance, a page in a Spanish book depicted with two friends wearing "*traje de gitana*" standing near a bull would be more suitable for a Spanish audience; however, for a reader in Poland the same content might not be very relevant; the reader might better engage with two friends wearing "s*troje ludowe*" standing near a "European bison".

This project aims to geographically localize, or geo-align, existing artworks in children's books using the concept of machine learning and computer vision, resulting in a different look for the artwork depending on the geographical location of its audience, while at the same time, maintaining the semantic information. Although this topic has not really been tackled before in this field of Artificial Intelligence (AI), the result is to eventually obtain a book that is comfortable for the audience to read, in a style that they recognize, and in some countries, more culturally sensitive. As mentioned by Najjar, the standards of modesty when publishing books in the Middle East must be met. However, manually realising this goal is a difficult task. There is also a need and resultant cost from hiring illustrators or graphic designers to do any modifications to artwork. There would be certain skills needed in editing softwares such as Adobe Photoshop and Lightroom. Additionally, this can be very time consuming, and thus, the end goal aims to autonomise and speed up this whole process, while producing quality picture books.

A comparative study of this novel concept will be conducted by analysing 3 varied methodologies. This project report will look at the work carried out to fulfil the project goal. Section 2 details any background research and information needed. Section 3 then details the 3 individual step processes, showcasing both the process and the results, with the last Sections 4 and 5 concluding the report and discussing future possibilities and improvements that can be done. There are a few risks such as unintentional misinterpretation or offence towards cultures, or even security risks when utilising the AI itself. These will be discussed further in Section 6. The output of this project is to present a proof of concept, adapting existing generative AI software to translate between higher level abstractions such as cultures, as opposed to existing styled translations. The concept specifically ascertains the best performant method requiring as little human intervention as possible. A varied collection of localised images will be procured and used in detailed evaluation of all approaches, highlighting the strengths and weaknesses of each. The following questions will be answered: Can we auto geo-localise artwork? From where we could obtain geographically relevant sample images? How to "replace" one object in an image with another?

## 2 Background

### 2.1 CLIP

#### 2.1.1 CLIP Background

Contrastive Language–Image Pre-training (CLIP) [6] was trained using zero-data learning *'where a model must generalize to classes or tasks for which no training data are available and only a description of the classes or tasks are provided.'* [7] CLIP is particularly suited to this project as it learns from *'unfiltered, highly varied, and highly noisy data'*. Since both the text and illustrations in children's books can be complex and contain many different objects, it is important that the embeddings can be used in a zero-shot manner because it is unlikely that any model will have been trained on the same, or even similar images/text as the ones used in the project. In order to train the model successfully, it was important to create a large enough dataset. This was done by constructing *'a new dataset of 400 million (image, text) pairs collected from a variety of publicly available sources on the Internet.'* [6] The CLIP model was also trained **using a novel approach** by taking text-image pairings from the internet and then creating a task, whereby the model must choose, given an image, which one of a possible 32,768 text snippets was paired with the image. *'To do this CLIP learns a multi-modal embedding space by jointly training an image encoder and text encoder to maximize the cosine similarity of the image and text embeddings'* [6] This also means that CLIP can map an image or piece of text into an embedding of size 1x786, since images and text are mapped into the same space. This becomes a useful tool for seed selection which will be explained in more detail in section 2.2.1.

In addition, a paper by Wolfe and Caliskan [8] analysed the performance of CLIP against GPT-2 [9] *'by comparing the geometry and semantic properties of contextualized English language representations formed'* by the 2. They also looked at the *'intra-layer self-similarity (mean pairwise cosine similarity) of CLIP'* and found that it is under 0.25 in all layers. The results showed that CLIP outperformed GPT-2 despite not being designed as a language model which produced high quality contextual word embeddings. Previous work has shown success in leveraging CLIP embeddings to edit and guide image generation [10]–[12]. A paper by Kim et al. [13] proposed *'a novel method, dubbed DiffusionCLIP, that performs text driven image manipulation using diffusion models.'* They used a model which had been pretrained on datasets containing images of faces, dogs, bedrooms and churches and then attempted to manipulate these by changing certain attributes or characteristics of the images. They found that text-guided manipulation using CLIP loss performed very well for in-domain and out-of-domain manipulation.

#### 2.1.2 Limitations

It is important when discussing novel models to understand the limitations so that the potential impacts can be mitigated or to add some context to some of the outputs. **One limitation of CLIP is that it is susceptible to biases introduced through the training dataset**, for example, they *'found discrepancies across gender and race for people categorized into the 'crime' and 'non-human' categories'* [6] Since the model is trained on such a large dataset, it is likely there will always be biases depending on the original data source. Attempting to create a training dataset without any bias would be infeasible at the scale required to train the model successfully. Furthermore, CLIP is very dependent on the class design as show in a paper by

Agarwal et al. [14] which looked at classifying 10,000 images from the FairFace dataset and included some additional classes to the ones provide by FairFace such as "criminal" and "gorilla". They found that 'Black' images were misclassified at a disproportionate rate as well as people aged 0-20. Since this project looks specifically at illustrations in children's books, often including human characters, it is important to be aware of these biases which are inherent in CLIP. Another paper by Struppek et al. [15] investigated how homoglyphs, which are visually similar non-Latin characters, can be used maliciously to trigger cultural biases which have been learned by the model. They concluded that *'models trained on large datasets from public sources capture the specifics of individual cultures implicitly from a relatively small number of samples containing non-Latin characters.'* and that *'By changing just a single character in the text description, we could either induce cultural distortions or obfuscate certain concepts in the generated images.* This demonstrates that it is important to verify the text which is being used to generate CLIP embeddings to ensure the characters are what they appear to be, and not used to exploit this method.

### 2.1.3 CLIP Embedding Similarity Across languages

To explore the CLIP embedding similarity across languages a text, a corpus of the 100 most common English words was translated into Hindi and Japanese. These words were then converted into their corresponding CLIP embeddings, and the cosine similarity was calculated between translated words. These values were then averaged to create the mean language vector, and these results are displayed in Figure 1. As shown, all mean language vectors are within a cosine similarity 0.008. This suggests all languages are not segregated into distinct areas but spread throughout common CLIP embedding space.

However, a second investigation was performed into translation of singular concepts across various languages, and these results are shown in Figure 2. **From this we can see that cultural concepts of "English" and "Japanese" and "Indian" do not translate between languages**, meaning there is little representation of other cultures in already deprived foreign language data. Therefore, the use of foreign language books would be severely detrimental to performance, as similar international concepts are not aligned within the CLIP embedding space across different languages. Although not exhaustive, due to time constraints, this brief investigation gave valuable insight for the dataset creation criteria.

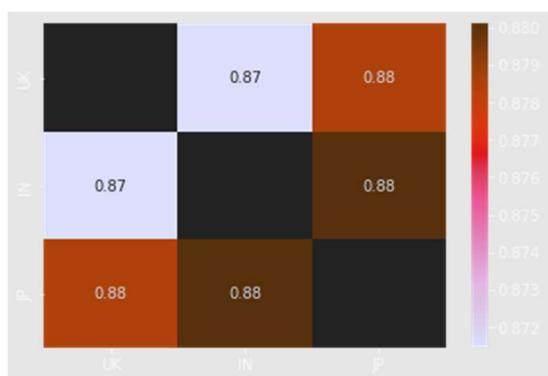

*Figure 1: Similarity of 100 most common words.*

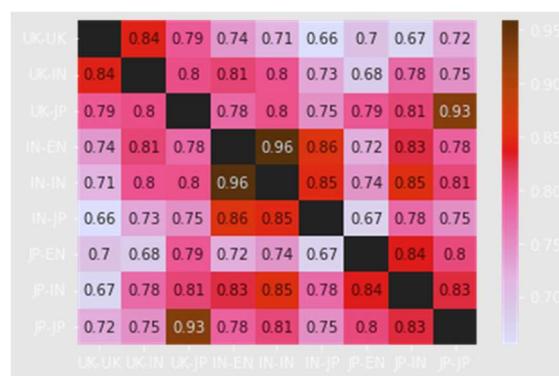

*Figure 2: Similarity of singular concepts across languages. Keys = ["England", "India", "Japan","इंगलैंड", "इंडिया", "जापान", "イングランド","インド"," 日本"]*

## 2.2 Latent Diffusion Models

A series of tasks exist within the whole scope of Image Generation. Firstly, the fundamental task is the **synthesis of an image conditioned on a textual prompt**. Secondly, synthesis can be alternatively performed without a text prompt, by conditioning on a class of images instead. This is known as class-conditioned [16] image generation. Highly variable **unconditional synthesis [16] is also possible when omitting all conditions**. To add to that, a singular image may be transformed in a desired way, as dictated by the prompt in image-to-image Translation [17], [18] tasks. Similarly, only selected parts of image may be altered with image inpainting [19]. Finally, unseen details within and outside of an image may be inferred by a model in image extrapolation [20] and super resolution tasks [20].

Several diffusion-based generative models such as diffusion probabilistic models [21] and noise-conditioned score network [22] have formed the basis for **Denoising diffusion probabilistic models (DDPM)** [23]. This approach was inspired by Nonequilibrium Thermodynamics [21], given the use of a defined Markov chain increasingly adding random noise to the data, destroying its structure. The model learns to reverse the diffusion process to reconstruct desired data samples from the noise. DDPMs have achieved **state-of-the-art performance** in a wide variety of generative tasks including image generation [111, 112, 113, 114], audio synthesis [24], molecular generation [25], and likelihood estimation [26]. It has also been demonstrated to **outperform General Adversarial Networks (GAN)** [27], including StyleGAN [51], which forms the basis of the aforementioned StyleCLIP [50] model.

The DDPM are based on the principal workings of the Diffusion process. **A Forward Diffusion process can be viewed as the addition of small amounts of Gaussian noise to a sample image ($x_0$), producing increasingly distorted images ($x_T$)**, whereas reverse diffusion recreates the true sample image from a Gaussian noise input. **DDPM are trained to estimate this reverse process** and require large datasets for reliable approximation, given the grand problem space. Mathematically, this can be interpreted as a large Markov Chain. Also, note that this algorithm is tractable when conditioned on noisy data, meaning minimal alterations of the noise can have significant effects on the resulting output.

**Denoising Diffusion Implicit Models (DDIM)** [28] increase the sampling speed of DDPMs, which require the simulation of a large Markov Chain. DDIMs construct a class of non-Markovian processes instead, to achieve the same objective with a **10-to-50-fold time reduction**. Furthermore, they achieve a **deterministic output given a randomised seed**, which enables greater consistency and creates the opportunity for semantically meaningful interpolate between two generated images.

**Classifier-free Diffusion Guidance (CFG)** [16] is an effective technique shown to increase the generative quality of DDPMs and DDIMs, which are class-conditioned. Previously used Image Classifiers are replaced by secondary unconditional model. CFG evaluates **both a conditioned diffusion model and an unconditioned diffusion model**, where the conditioning can be on any text prompt, class-label or embedding. Furthermore, the two models are trained on a single neural network sharing the same weights and parameters, however two separate models may also be used. The unconditioned model always receives a null token for its class identifier token, whereas the conditioned model is given and existing class token instead. The resulting scores from each model, are linearly combined in addition to a guidance weight parameter to adjust between the sample quality and diversity.

Working directly with pixels is computationally expensive, hence the rise of **Latent Diffusion Models (LDM) [19]**, which abstract images to a **high dimensional latent vector** and perform the diffusion process on the small vector instead, substantially increasing training and inference efficiency. LDMs have been widely implemented throughout various large-scale diffusion models, including but not limited to: Stable Diffusion [28], DALL·E 2 [26], and Imagen [31], which will be reviewed in further detail below.

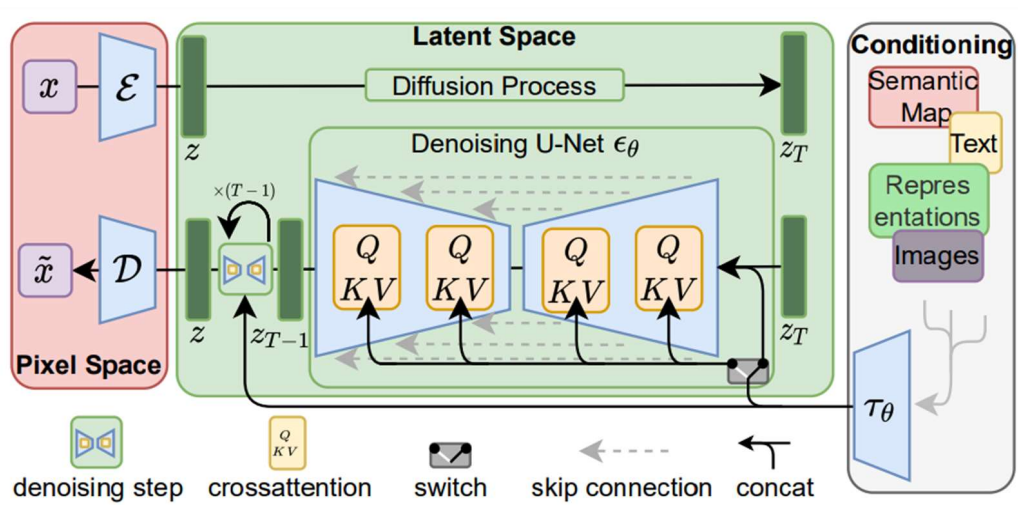

*Figure 3: Latent Diffusion Model Architecture [19]*

LDMs abstract pixel-level redundancies using an **autoencoder**, with higher level semantical concepts are then instead manipulated and generated by the diffusion model. Firstly, the raw image is processed by **encoder (E), producing a latent vector**. Secondly, LDMs can condition on a **wide range of condition modalities** (Prompt, Class Label, Semantic Map, Images). Each type of data also requires its own domain-specific encoder to map the condition into the same latent space before passing it onto the cross-attention mechanism [29]. Cross-attention layers are utilised to effectively combine the conditioned latent and image latent into a single encoding. Next, the concatenated latent is run through the **reverse diffusion process**, which utilises a time-conditioned **U-Net [30].** This is repeated for set number of iterations specified as a hyper-parameter. Finally, **decoder (D) reconstructs the image** form the latent vector to form a final output image.

Ho et al [31] increased sampling speed using a progressive distillation technique. This can be applied for both unconditional Image generation as well as class conditioned. A follow up paper by Meng et al [32] has reduced the required amount of denoising iterations using a two-stage **distillation process**, resulting in a **20-fold increase in speed.**

### 2.2.1 Qualitative Comparison of Latent Diffusion Models

The encouraging performance of LDM models, alongside growing appeal from the general public, has spurred the creation of a multitude of models, each targeting various market segments.

DALL-E 2 [33] is the second revision of OpenAI's Image Generation Model. 650 Million Images from across the CLIP and DALL E training sets, were used to train the model. This CLIP encoder [34] is used to encode the prompts. This is followed by the same diffusion process and decoder featured in the previous GLIDE model [35], utilising over 3.5 billion parameters. The decoder outputs a small intermediary 64x64 output image. This is then upsampled by two ADM-Net [36] upsamplers, first into a 256x256, followed by a final 1024x1024 image.

Imagen [35] is an analogous model created by Google Research. The model was trained on internal datasets, as well as the LAION-400M dataset [37], totalling over 860 million image-text pairs. A frozen T5-XXL [38] encoder was utilised over a CLIP encoder, as despite similar FID scores [39], human evaluators favoured images generated from T5 encodings. A standard U-Net Architecture was trained for the diffusion process. Similarly, a 64x64 image is generated by the decoder and upscaled twice [20] into a finalised 1024x1024 image. Performance analysis using the FID scores of COCO validation set [40], indicates superior performance over DALL-E 2.

ERNIRE-VILG 2.0 [41] is a large-scale **Chinese text-to-image diffusion model**. 170 million text-image pairs were used from internal Chinese datasets as well as English LAION pairs, which were translated into Mandarin. A unique architecture was implemented using a series of 10 expert denoising agents to incorporate domain-specific knowledge and decouple diffusion processes into separate stages. The model is considered to be **state-of-art** with a marginally lower FID score of 6.75 than Imagen on the COCO dataset.

**Stable Diffusion (SD) [19] is a highly prominent model** released by a collaboration of a Stability AI, CompVis LMU, and Runway with support from EleutherAI and LAION. Training of the most current version 1.5 [42], used subsets of the LAION-5B dataset [43], containing over **5.85 billion image-text pairs, of which 2.32 billion contain English language**. Initial pre-training was performed using images of size 256x256, and using 512x512 images for fine tuning. Drop out of the conditioning text was also used within the latest version. Similarly, for Imagen, a CLIP encoder was implemented. A purpose-built U-Net was trained from first principles, with over 860 million parameters. Although the original model is best suited to the generation of 512x512 images, given the training data, the decoder can be adapted to generate any image size, without the need for intermediary super-resolution processes. The **FID score exceeds 15, which is comparable to the performance of DALL-E** 2, however lesser than the State-of-the-art. It is important to note that SD is not proprietary as previous English Language based models, but **open to the public under the Responsible use of AI License (RAIL)** [44]. Due to this, significant research interest is placed on the model driving innovation. Approaches such as inpainting [45] and image to image modification extended the models capabilities and have been created as a result of **community support.**

### 2.2.2 Variations of Real Images

As previously stated, LDMs can condition on a wide range of modalities, including images. By leveraging this alternative conditioning mechanism, models can used in a **Img-2-Img methodology**, where real images are inputted into the model and a variation of them is outputted. This **removes the need for textual prompts**, which may be ambiguous about certain aspects of the image, reducing generation reliability.

To achieve this, **conditioning is performed upon the latent CLIP vector of a real image,** instead of a textual prompt. The previously utilised CLIP Text encoder is replaced by the CLIP Image encoder, and the resulting latent is set as the conditional latent.

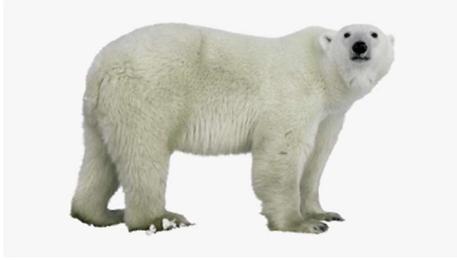 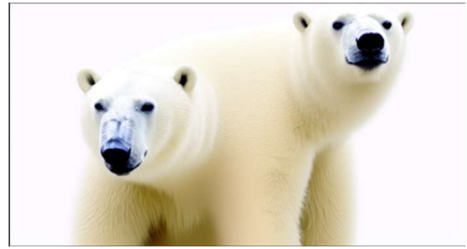

*A) Source Image          B) Variation of Source Image*

*Figure 4: Variation of Source Image, generated using Stable Diffusion (version 1.5)*

A brief investigation into the performance of Image Variation[1] using the Stable Diffusion Model was conducted. An example of a generated image can be seen above, using a Guidance scale of 7.5 and a predefined random seed of 1234. It is evident that semantic information around the style and content of the image is retained, however the image composition and small details are absent in the generated images, with anatomical issues being exaggerated. This would be problematic for the generation of illustrations. Therefore, from this limited empirical example, one can conclude that **reconstruction performance is an inadequate** substitute for the conventional prompt conditioning method.

# 3   Technical Approach

## 3.1   Dataset creation

To train the AI model for each of the methods, a dataset of books needed to be created. This had to include books that the model would be able to work with. Since the goal was to see if the AI model would be able to translate the images into different art styles, it was decided that books from countries with **distinct art styles should be selected. India, the UK, Japan and the Middle East,** the latter chosen to test if it would be possible to make the translated artwork culturally sensitive. The books were sourced from free, readily available websites, such as Open Library, Storyweaver, BookTrust, Freebooksforkids and Project Euclid, to name a few [46]–[50]. There were several criteria that had to be followed when choosing a book to get ideal results when working with the AI, and prompt generation. They are listed below, in Table 1:

---

[1] https://huggingface.co/spaces/lambdalabs/stable-diffusion-image-variations

|  | Should | Should not |
|---|---|---|
| Sentence Criteria | Describe the scene and make sense on its own. | Reference the previous page/happenings in the book

Be in a first-person narrative

Have questions, dialogue

Be too long, as the model has a word limit on the prompt

Have non-English names/words |
| Character Criteria | Human characters (easier to localise) | Anthropomorphic animal characters, as animal mannerisms may confuse the AI who assumes human characters |
| Book Criteria | Match the author's ethnic origin

Has already been translated correctly into English.

Written for modern audiences | Talk about the culture itself, or reference cultural happenings/beliefs (e.g., Christmas, Ramadan, religion) |

Table 1: A *table summarising the criteria in choosing books for the dataset.*

These limitations were quite challenging for UK books to be translated, as seen in a study done by the CLPE, 38% of books that were published in 2019 featured non-human characters, such as animals or inanimate objects at their main characters [21]. The only books that relatively matched these criteria were published more than 23 years ago. Hence, the book that was eventually chosen was written by a US author, but it was ensured that it would be easily comprehensible to UK audiences. It was also difficult and time consuming to find a book that completely adhered to the criteria, so some sacrifices were made. For the online questionnaire, 3 books were chosen from each country, and the rest of the dataset was used for overall AI model training and experimentation to see which yielded the best results. Each book was sorted into a separate folder and uploaded to GitHub. The text and corresponding image were extracted from each page of the book. The images were in a png format, and numbered, so it was compatible with Stable Diffusion. Where possible, it was ensured that the text was not visible in the image. The list of the full dataset of books is in Appendix E.

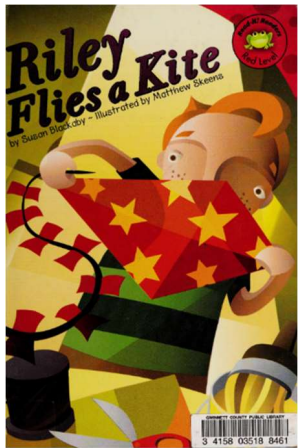 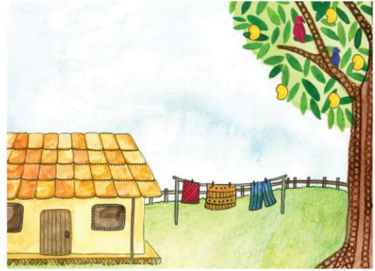 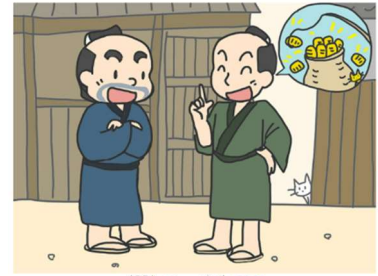

*UK: Riley Flies a Kite* [51]  *India: The Mango Tree* [52]  *Japan: Not You!* [23]

*Figure 5: The books chosen for AI model training, and image generation*

The book chosen to represent UK audiences was titled: Riley Flies a Kite, written by Susan Blackaby and illustrated by Matthew Skeens [51], and downloaded from an online open library. This story was about a young boy who is looking for a place to fly his kite. Here, the main characters, Riley and his father, and the locations (the football field, and the park) could be translated, which is provides a **better performance in the translation process, compared to the other UK books in the dataset**. However, in the beginning, the sentences reference the previous pages that are describing the assembly of the kite, which is not ideal.

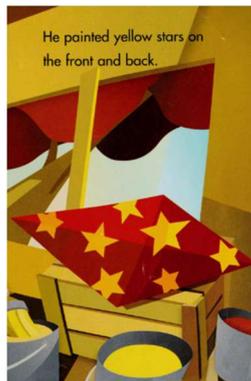 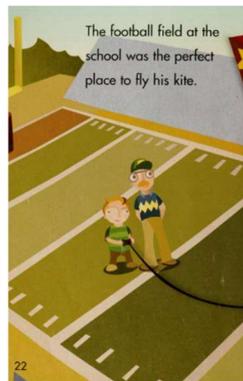 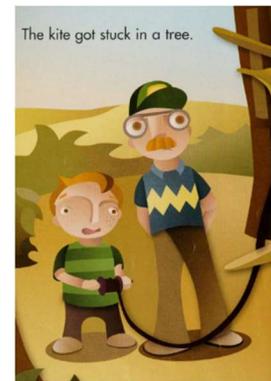

a) *Page 5 from Riley Flies a Kite*    b) *Page 22 from Riley Flies a Kite*    c) *Page 14 from Riley Flies a Kite*

*Figure 6: Excerpts from Riley Flies a Kite, pages 5,22 and 14 respectively. It can be seen that page 4 and 7 would give issues for the AI prompt (reference to previous page), but pages 22 and 14 are short and descriptive on their own, which are ideal* [51].

For Japanese audiences, the story was taken from a website, Hukumusume [23]. It was titled Not You! It is a 5-page short story from the Edo period in Japan, about a fisherman who wants to get gold coins from fishing, but instead catches a fish instead. It matches the criteria, as its sentences are relatively straightforward, except at the very end, where there is a little bit of dialogue.

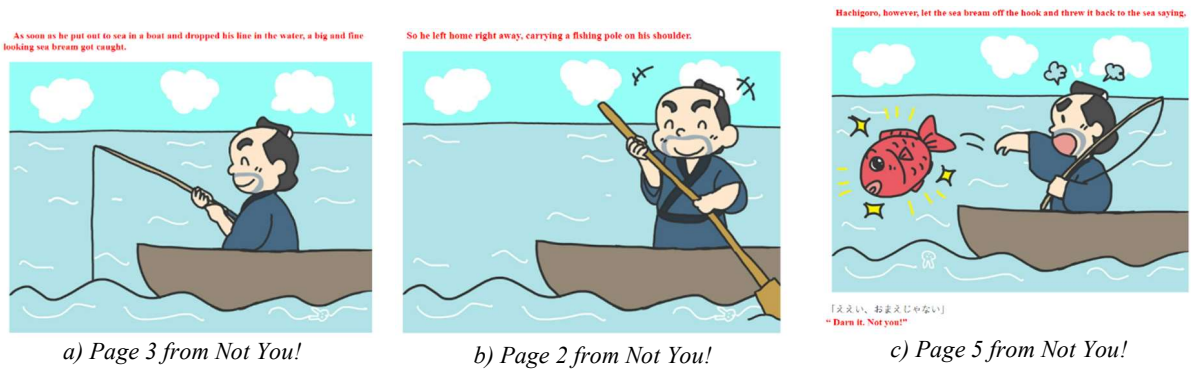

*a) Page 3 from Not You!*  *b) Page 2 from Not You!*  *c) Page 5 from Not You!*

*Figure 7: Excerpts from Not You!, pages respectively. It can be seen that pages 2 and 5 could give issues for the AI prompt (dialogue and reference to previous page), but page 3 is short and descriptive on its own, which is ideal [23].*

An Indian book was chosen from the Storyweaver website, titled The Mango Tree [52], which is a simple story about a boy visiting a mango tree at his grandmother's house. Even though a few aspects invalidate the criteria, it was too late to choose another book in the dataset, due to time constraints. It is in a first-person narrative, some sentences are clear, short, and concise, with enough detail.

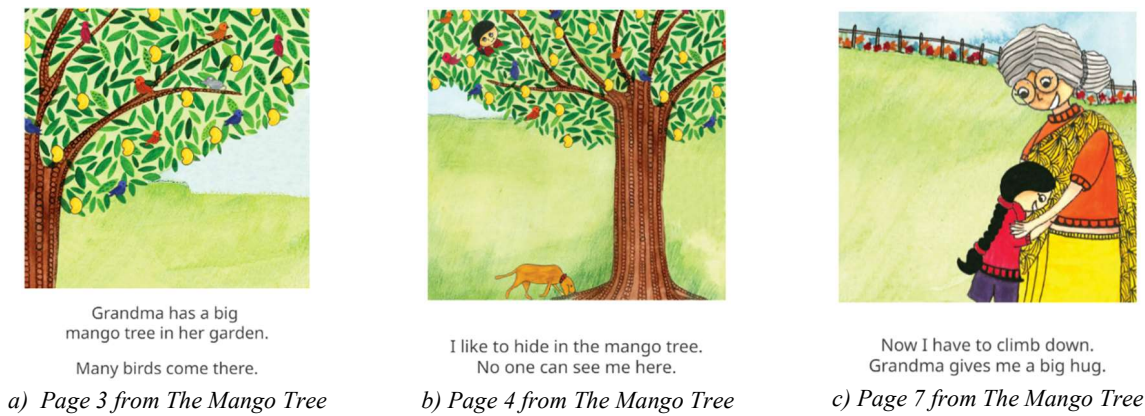

*a) Page 3 from The Mango Tree*  *b) Page 4 from The Mango Tree*  *c) Page 7 from The Mango Tree*

*Figure 8: Excerpts from The Mango Tree, pages 3,4 and 7 respectively. It can be seen that pages 4 and 7 would give issues for the AI prompt (first person narrative and reference to previous page), but page 3 is a good example.*

It was difficult to find books from Middle Eastern authors that fit the criteria as mentioned above, in the limited time available, and so the focus shifted to translating the UK into a Middle Eastern art-style.

### 3.2   Method 1: Prompt Augmentation using Keywords

There are many factors which can influence the image generated from a given text input. This method is the first, and most naïve method of the three discussed in this report. However, just because it is the simplest of the 3 methods, it does not necessarily mean that it will produce the worst results. It serves as a good base line to compare the other more involved methods, as it focuses on taking the **raw text** from a book for which an image needs to be generated, and **automatically editing it to produce a new prompt**, which is more likely to include the key objects mentioned in the text, as well as be **culturally sensitive**.

### 3.2.1 Prompt Engineering

Firstly, it is important to understand how Stable Diffusion generates images and what methods can be used to influence these results. This section will investigate various methods and as to how they can alter a given output. A paper by Witteveen and Andrews [53] **describe how prompts can be engineered for Stable Diffusion to produce the desired results**. They observed that different linguistic categories, also known as parts of speech, would affect the generated image in '*different, but consistent, ways.*' They noted that **nouns would dramatically change the output of the image** by introducing new content and that simple **adjectives would have a relatively small, yet significant**, effect. It was also observed that adding something to such as **"in the style of [artist]" would dramatically change the constitution of the image**. They performed additional experiments such as how effective was the repetition of words to change a prompt and found that *"having multiple occurrences of a word can change the image it often has no effect at all or has an effect that is not a desired semantic effect the word might be expected to contribute"*.

In addition to academic papers, there are also large databases available which illustrate different styles and techniques that can be used to achieve the desired results. For example, this database created by Maks-s [54] contains examples of how different modifiers will affect the outcome of the model. One example exploring classifier free guidance (CFG), which is a parameter that controls how much the image generation process follows the text prompt, can be seen in Figure 9a. As is evident from this example, **increasing the CFG too many leads to the model overfitting** and creating images which are not recognisable as the original prompt. Another parameter which can be changed to produce different outputs is the seed used to generate the image. The seed is what is used to initialize pseudo randomizers for the diffusion model. Inherently, it does not describe any information about what should or should not be in the final image and allows for reproducible image generation since stable diffusion is deterministic (i.e.. will produce the same image given the same prompt and seed). However, a post on Reddit [55], demonstrates that the seed chosen for a given prompt can have a large effect on the colour and composition of an image. Some samples are shown in Figure 9b.

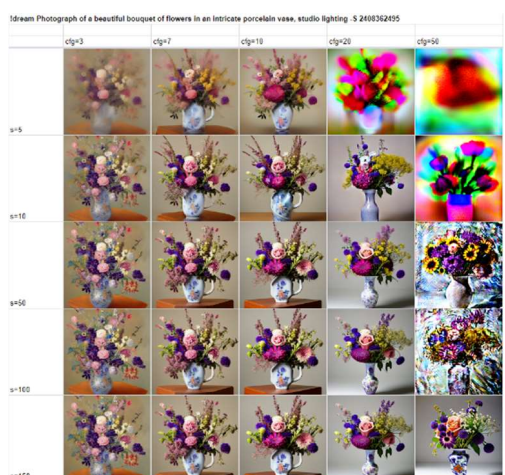
*Figure 9a: Effect of CFG on output [56]*

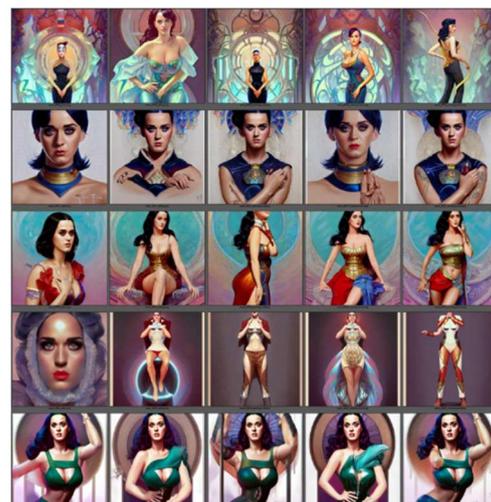
*Figure 9b: Effect of seed selection on output [55]*

From these examples, it can be seen that by keeping the seed the same and slightly editing the prompt produces a image of similar style, background and framing. This is called seed editing [57] and can be useful for creating multiple images which seem coherent when viewed together. However this property does not hold when the various prompts are vastly different from each other.

**The punctuation which is used in a prompt can also have dramatic effects on the final output.** Adding exclamation marks to words **will increase the weight** of that particular word within the prompt as seen here [58]. In contrast, adding brackets will reduce the weight of the word or phrase within the bracket as demonstrated here [59]. It is important to be aware of this as books will naturally include punctuation which could influence the image generation unintentionally. The use of commas is also important when creating effective prompts as shown in this article [60] the prompts which generate the best **results use commas** to separate distinct qualities of the desired output such as a noun **phrase, style or expression.**

### 3.2.2 Prompt Processing

From the research presented above, it is clear that there are various techniques which can be applied to a standard sentence or page from a book in order to create a prompt which will be more likely to produce the desired image. This approach looks at a way to automate this process and an overview is outlined below in Figure 10.

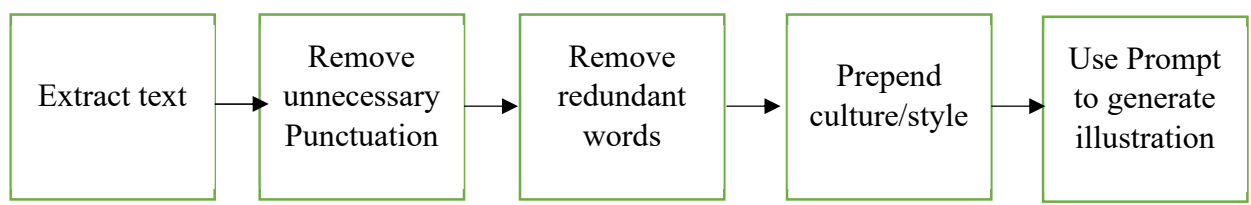

*Figure 10: Prompt Processing pipeline*

The first stage is to remove any punctuation which could have unintended consequences on the output. For example, question marks, exclamation marks and parentheses should be removed, since these affect the weighting of the previous word, which is usually not what is desired from the prompt. Punctuation such as commas, and full stops are generally used in the same context in natural language and prompt design, i.e. both are used to separate different ideas or phrases within a sentence, so these do not need to be removed. Next, some **co-ordinating conjunctions** such as 'and', 'but' and 'yet' are usually used to link two phrases or clauses together but do not provide any additional information about the scene or story being described by the book. As mentioned in the prompt exploration section, it is frequent practice in Stable Diffusion to use commas to separate clauses or phrases so these co-ordinating conjunctions can be replaced by commas using regex.

As mentioned previously, different word types have different magnitudes of effects on the output image. Thus, it is useful to have the words linked to their associated word types also known as Parts of Speech (POS). While there are different formats of classifying and tagging POS, one of the most widely used is the Penn Treebank tags [61] which contains 36 unique tags. By implementing a POS tagger, it is possible to compare the effects on using different participles of verbs however from initial testing there did not appear to be a significant difference between different verb types. More complex analysis of POS could possibly be used to further improve the performance of this method however this was deemed out of scope for the project.

Stable Diffusion also has **a maximum token limit of 75 tokens** so it is important to remove words which will not contribute to improving the prompt, without removing words which will.

Since language is complex, words can have larger or smaller effects depending on the context in which they are used. As such, only words which have consistently insignificant effects can be removed using a simple regex. These include words such as "the" or "a". On the other hand, possessive pronouns such as "his" or "her" are usually superfluous when included in a sentence such as 'The boy took his football to the park" since the ideal of a male character is already encoded in the phrase "the boy" and as such Stable Diffusion would likely include a male character even if the word "his" was removed. However, consider another example where if the first page of a book contains the text "The boys mum bought him a new hat," and page 2 contains the text, "His hat was too big". Since each page is linked with a single prompt and there is no context added of previous pages, removing "his" from page 2 would lead to the prompt, "hat was too big" which would most likely lead to the generation of an image of a large hat, but not necessarily one that belongs to a person. The expected output from the original text would likely be one of a male character wearing a hat which is too big.

### 3.2.3 Choosing Seed

As mentioned previously, the choice of seed can have a large effect on the composition, features and style of generated images, and thus is important. One possible naïve approach would be to generate multiple full books worth of images, and use human intervention to decide which is the best. However, this is quite an inefficient and subjective strategy which would still require human input which opposes the objectives of this project. An alternative, more objective approach, is to take a text prompt and image from a book which has an illustration that is culturally sensitive and use this text prompt to generate a large number of sample images. Then, the cosine similarity of the clip encoding of each of the generated images could be calculated against the target image, choosing the seed which resulted in the highest value.

To test the effectiveness of this idea, **1,000 images were generated from the prompt** "*Indian children's book style, Grandma has big mango tree in her garden. Many birds come there*". Below the objective image (Figure 11a) can be compared to the **best generated image** (Figure 11b) which had a **cosine similarity of 0.91 between the CLIP encodings**. The generated image with the **worst performance** is also pictured in Figure 11c, which had a cosine similarity **between embeddings of 0.62.** As is demonstrated by these results, it is clear that the CLIP embeddings encapsulate the style and contents of the image and as such, can be used as an option to select the best seed.

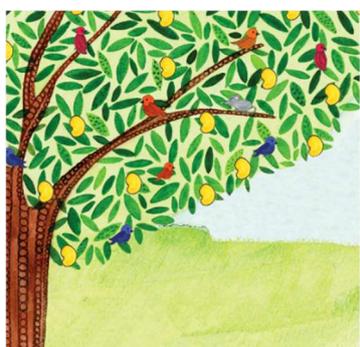
*Figure 11a) Original Image, taken from the first page of The Mango Tree.*

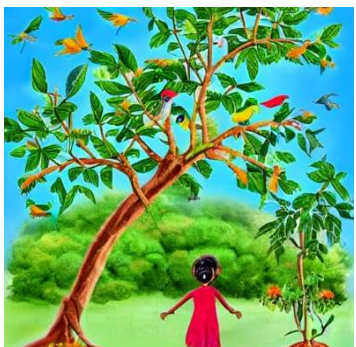
*Figure 11b) Best generated image according to cosine sim*

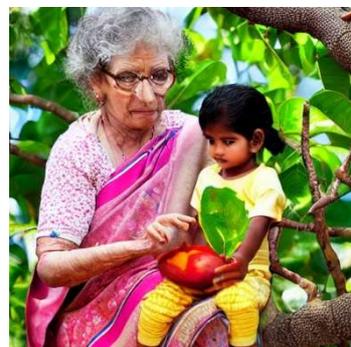
*Figure 11c) Worst generated image according to cosine sim*

## 3.3 Method 2: CLIP Embedding Mask

Method 2 explores using a CLIP Embedding Mask to encode the cultural information of the chosen geo-location. However, one **issue** with this approach, as well as **with most prompt editing in general, is that the success of the manipulation is dependent on the text selection which often requires domain specific knowledge in order to be most effective**. A paper by Kocasari et al.[62] explored how this issue can be overcome by *"automatically determining the most successful and relevant textbased edits using a pre-trained StyleGAN model"* which uses a beam search approach[63] to identify which keywords will lead to the most success. These keywords are selected based on their relevance and editability. Since these keywords are chosen based on CLIP embeddings by the StyleGAN, it removes the need for the person editing the prompt to have domain specific knowledge. Their results showed clear improvement when compared to other latent methods, both supervised and unsupervised.

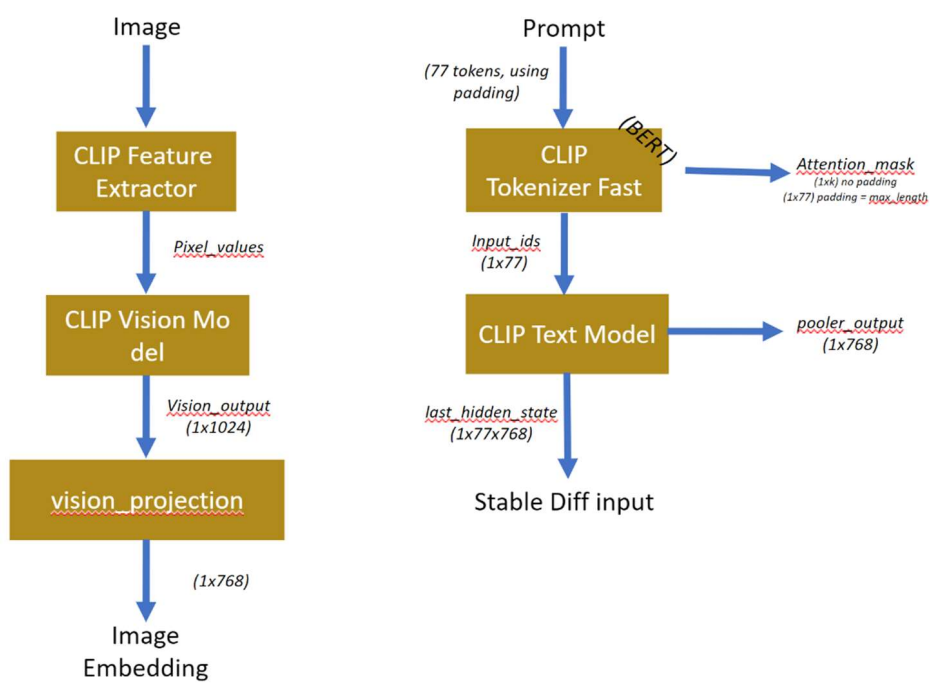

*Figure 12: Pipeline Process for the embedding of Images and Prompts in CLIP version "clip-vit-large-patch14"*

### 3.3.1 Multi-Lingual Comprehension

Since a lot of cultures or countries have their own language or dialect, it is reasonable to infer that the use of a certain language can encode certain cultural ideas based when generating stable diffusion images from multi-lingual prompts. If it is possible to get an accurate image from a prompt in the language of the chosen culture, it is likely to be more culturally sensitive than images generated from the same prompt but in English. However, the stable diffusion model **was trained with English captions** which means it will not perform as well as in other languages. Figure 13 shows the exploration of this idea. Various images were generated using the same prompt but translated into different languages. As can be seen, Spanish, German and French performed well; clearly showing a bear in a sitting position with Spanish and English clearly showing a polar bear as opposed to a brown bear. In contrast, Dutch, Hindi, Japanese, Arabic and Polish all failed to generate a bear at all or even anyone sitting.

| English | German | French |
| --- | --- | --- |

| English | German | French |
| --- | --- | --- |

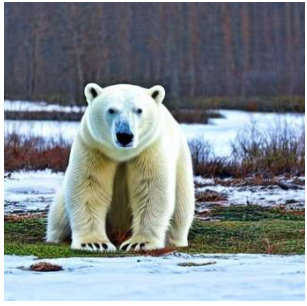 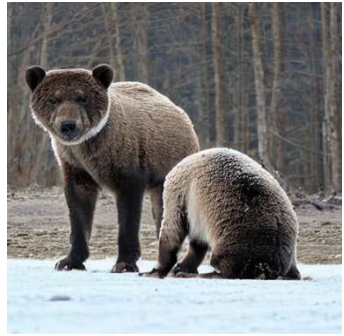 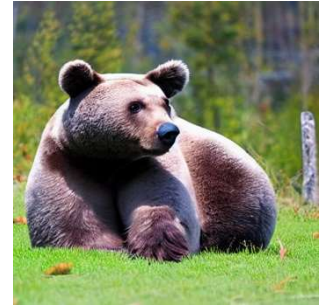

| "A photo of a Polar Bear sitting" | "Ein Bild eines sitzenden Eisbären" | "Une photo d'un ours polaire assis" |
| --- | --- | --- |

| Spanish | Dutch | Hindi |
| --- | --- | --- |

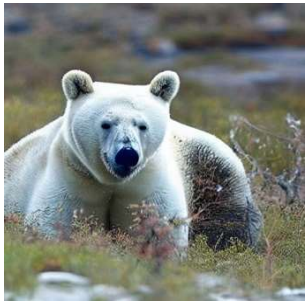 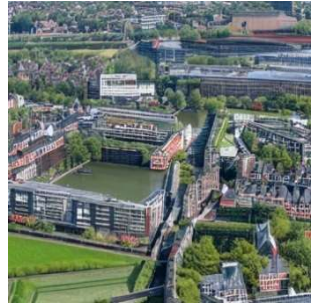 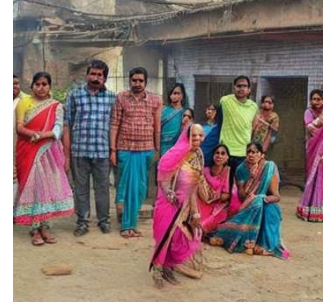

| "Una foto de un oso polar sentado" | "Een foto van een zittende ijsbeer" | "बैठे हुए ध्रुवीय भालू की एक तस्वीर" |
| --- | --- | --- |

| Japanese | Arabic | Polish |
| --- | --- | --- |

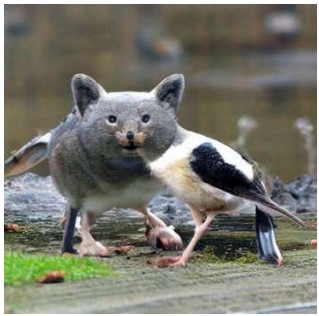 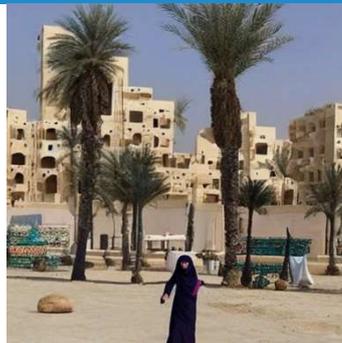 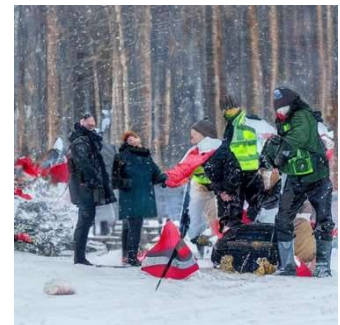

| "ホッキョクグマが座っている写真" | "صورة لدب قطبي جالس" | "Zdjęcie siedzącego niedźwiedzia polarnego" |
| --- | --- | --- |

Figure 13: Results of Multi-Lingual Comprehension

A more comprehensive investigation of the multilingual capabilities of Stable Diffusion explores the prompt "Two cats playing ping-pong on orange background" [64] . This showed similar findings with German, French and Spanish performing well but most other languages struggled to generate a cat or a ping pong ball or anything orange. However, it is interesting to observe that the art style varies wildly depending on which language is used to despite no art style being specified in the prompt. This suggests that the language being used does embed some information about the culture however since Stable Diffusion was not trained on different languages it will perform poorly trying to generate relevant images.

An explanation for this is could be that the Stable Diffusion model was trained on a subset of the Lioan-5B dataset [43]. As such any language which have **limited representation in this dataset will perform poorly in stable diffusion**. French, German, and Spanish were all in the top 5 most represented languages in this dataset.

### 3.3.2 Mask Selection

The choice of keywords selected as well as the multiplier applied to the mask will affect which areas and to what extent the original image is generated, as seen in the example below. Figure 14d shows how using the Indian mask removes the chair and the man is now sitting on the floor. Sitting on the floor is more common in Indian culture than in other places around the world. In addition, the man's facial features have also changed from the original, he now has a darker skin tone which is more common to people of Indian backgrounds. In Figure 14, it is clear that the facial features are changed based on the mask applied while still retaining the original features of the image such as the colouring and pose of the man. Figure 14a shows the original image for reference.

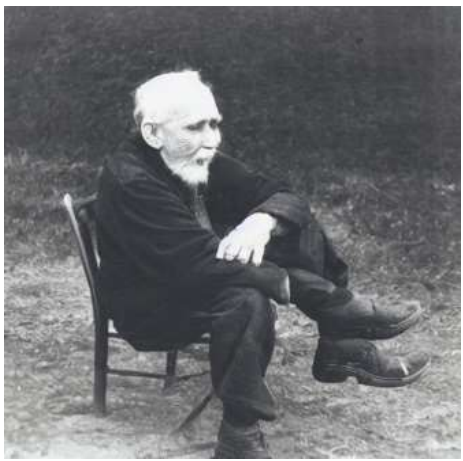

a)   Prompt:  "A photo of an old man sitting"

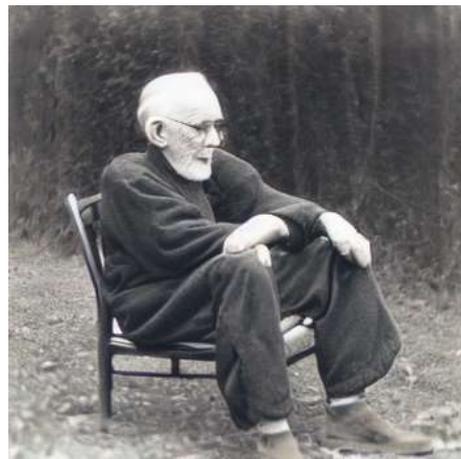

b)   Mask: "口米Ｏ嚮厘" (Japanese Cartoon) Multiplier: 0.2

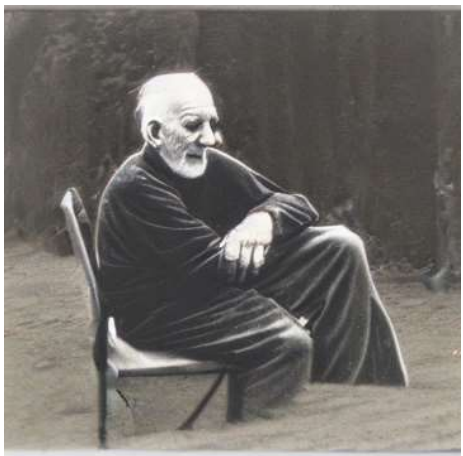

c)   Mask: "كارتون عربي" (Arabic cartoon) Multiplier: 0.3

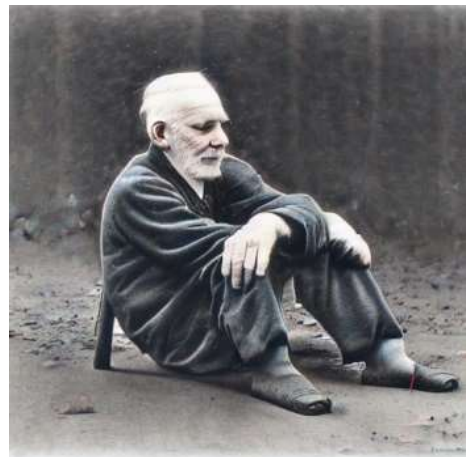

d)   Mask: "भारतीय कार्टून style " = (Indian Cartoon) 2Multiplier:  3

Figure 14: Iterations of various cultural masks applied to SD generated image using the prompt "A photo of an old man sitting" and seed 1234.

## 3.4 Method 3: Cross Attention Control with Editorial Prompts

SD and many other models explored in Section 2.2, may be **challenging to control** when image generation is conditioned upon a singular prompt, 77 tokens long. This behaviour can result in inconsistent images being generated, as well as requiring unintuitive prompt adaptations though prompt engineering to be made, as demonstrated with Seed-Editing in Method 1 (Section 3.2.3).

Although a seed remains consistent when altering prompts, even by a single word, unintentional alterations to the **composition, perspective and styling** of the image may be introduced. For example, Figure 13 translates the prompt into different languages, which causes the polar bear to become a brown bear given German and French prompts. This may come as a result of **learned common associations,** with most bears being brown in Western Europe.

Previous approaches by Avrahami et al [65], restricted the area in which the changes were applied, minimising the risk of undesired re-styling or perspective shifts. However, this inhibits global changes which are necessary to achieve successful image generation of contrasting cultures.

**Cross Attention Control** [66] investigated by Hertz et al, enables much finer transformation of the prompt, in order to map the prompt to a desired culture, by modifying the internal attention maps of the diffusion model during inference. A **secondary editorial prompt** is introduced in order to adjust the attention mechanism prior to the image decoder generating an image from the resulting latent vector. Interestingly, spatial layouts and geometry from the original generation is preserved to a high degree. This forms the basis of novel **Prompt-to-Prompt Editing [66]** approach.

The method utilised can be used for **Style Injection and Global Editing** where the initial prompt is appended with a description of a different style or contextual information, such as cultural background. This is essential for cultural image generation. Target Replication is also made possible by replacing nouns and named entities from the initial prompts with the ones desired, which replaces the original subject with a new one using the same pose.

Inversive Methods explored in Section 2.2.2, can also be leveraged by this approach, enabling the editing of real images, without the need for an initial prompt. However, empirical evidence shows that the reconstruction performance of mapping real images into latent vectors is insufficient, and therefore the decision was made to avoid the use of real images.

As described in Section 2.2, **Cross Attention layers** [152] combine the conditioned latent, representing the initial prompt, with the image latent into a single encoding. A **spatial attention map** within each layer dictates the model's abstractions for a singular token (shown below). These Maps can be generated for both the initial and editorial prompts, enabling various replacements, additions and transformations to combine the two prompts adjusting the semantic information. Furthermore, the separation of tokens enables additional guidance weights to be given to each token, increasing or decreasing the importance of a word in the prompt, which will increment its manifestation within the generated image. This is dictated by the "prompt_edit_token_weights", within the utilised implementation[2], which stores an array of integer pairs, storing the index of the token and its corresponding importance.

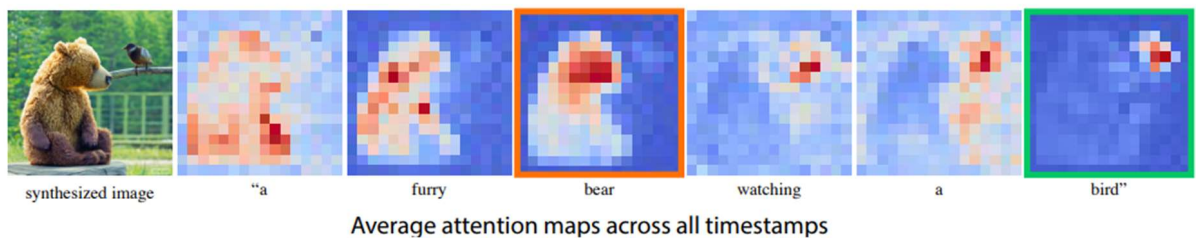

*Figure 15: Visualisation of the average Attention Map for each token within a conditioned text prompt* [66]

Various parameters in the python implementation[2] such as "prompt_edit_spatial_start" and "prompt_edit_spatial_end" dictate the models creativity with respect to the initial prompt individually for smaller and larger details respectively. This is adds further customisation on top of the Classifier Free Guidance parameter used within LDMs. It is important to note "prompt_edit_spatial_end" must be greater than "prompt_edit_spatial_start". Similarly, "prompt_edit_tokens_start" and "prompt_edit_tokens_end" also adjust the model's creativity for small and large details, however are intended for used with inverse methods using latents from real images. The effects of these parameters will be further explored within Section 4.3.

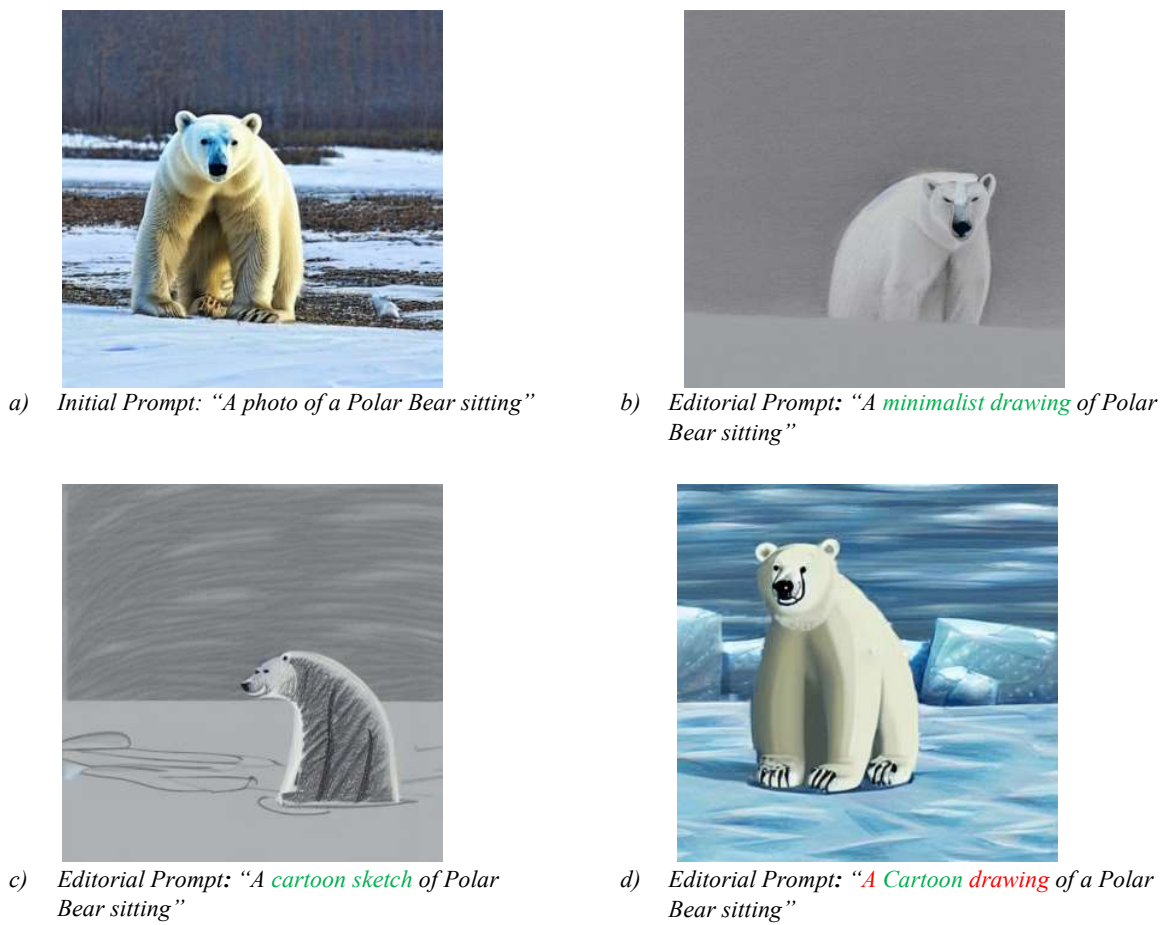

a) *Initial Prompt: "A photo of a Polar Bear sitting"*

b) *Editorial Prompt: "A minimalist drawing of Polar Bear sitting"*

c) *Editorial Prompt: "A cartoon sketch of Polar Bear sitting"*

d) *Editorial Prompt: "A Cartoon drawing of a Polar Bear sitting"*

*Figure 16: prompt_edit_spatial_start=0.7, prompt_edit_spatial_end=1.0,*

---
[2] https://github.com/bloc97/CrossAttentionControl

# 4 Analysis of Results

This section will examine the strengths and weaknesses of each method when generating images for the various children's books listed in Appendix E.

In general, the methods in this report were very good at generating images from **simple and descriptive inputs**, such as those from "In My Garden" [20]. However, there was a struggle with **negative inputs**, such as "The kite did not fly" (page 5 of "Riley Flies a Kite" [51]). As shown in Figure 17, the **negative token is usually ignored,** and the model instead draws the kite in the air.

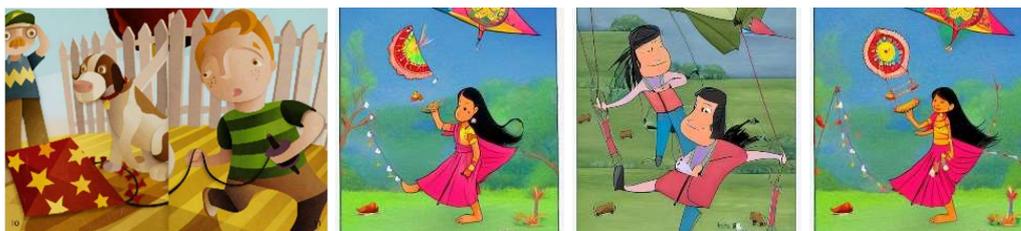

*Figure 17: generated results for page 5 of "Riley Flies a Kite". The accompanying text reads: "The wind did not blow. The kite would not fly.". From left to right: the original image captured from [51], image generated by method 1, image generated by method 2, image generated by method 3*

Additionally, because every page of the books was input into the Stable Diffusion model **separately**, all three methods struggled with consistently drawing recurring characters between pages. This is especially obvious when **pronouns are used to describe an animal** rather than a human. For example, on page 9 of "In My Garden", the 'she' pronoun refers to a bee introduced in the previous page. However, since this context has been lost, each method has generated images containing a human girl, as shown in Figure 18.

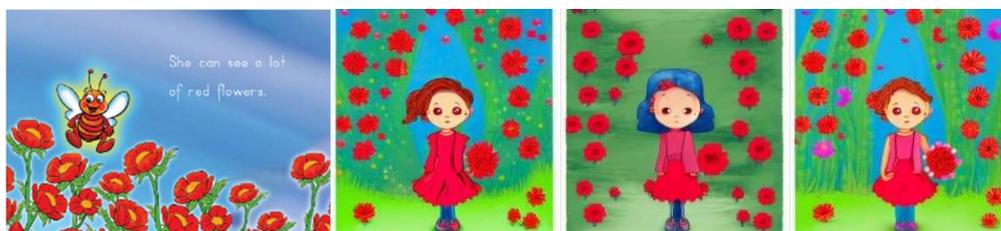

*Figure 18: generated results for page 9 of "In My Garden". The accompanying text reads: "She can see a lot of red flowers". From left to right: the original image captured from [20], image generated by method 1, image generated by method 2, image generated by method 3.*

Another problem displayed in Figure 18 is that the SD model **distorts human faces**, especially the **eyes**. This creates slightly uncanny characters that can be **off-putting to readers**. On other occasions, **too many** eyes are drawn, or **none** are drawn at all, which is sometimes more acceptable in pictures with a simplistic art style.

Another trend among all three methods is that an image can occasionally be generated **too realistically**, despite the desired children's book art style being explicitly specified. This can be very inappropriate for children's RfP books, especially when compounded with the distortions explained previously. Nonetheless, even with realistic images, all three models usually change the characters' features (e.g. their clothes, skin tone and hair colour) to **fit with the target culture**.

## 4.1 Method 1 Observations

This section will showcase the results produced by method 1 and **illustrate some of the characteristics and common features which the method produces**. In addition, it will explore in which areas the method performs well and in which areas it performs poorly and offer possible justifications for this. All images were **generated using the same seed, 1234**, which was not pre-selected using the cosine similarity method detailed in section 3.2, since this method would likely choose a seed which was better for the sample it was tested against, but not necessarily across the pages, books or cultures which tested.

As demonstrated in Figure 19, this method, particularly when attempting to generate images geo-localised to India, would often produce more realistic or life-like images despite this not being specified in the prompt. In addition, throughout the results it is clear to see that the model **struggles to accurately depict faces**, as shown in Figure 19a, the eyes in particular cause the face to look strange and inhuman. However, it does illustrate all the key ideas and themes from the story, there is clearly a child hiding in a mango tree and thematically it is very similar to the original shown in Figure 19b. The child also has a skin tone which would be familiar to children who read the story and her clothing is also typical of the country.

An interesting point to note is that the model can usually generate accurate looking faces of famous people, likely due to having more access to data, and perhaps a workaround to the realistic face issue could include replacing instances of generic people with famous people. However, this would cause issues with using their likeness without permission, and in addition, most story books are illustrated in a non-realistic style as shown in Figure 19b so this is likely a complicated fix that does not address the root problem, which is the incorrect style.

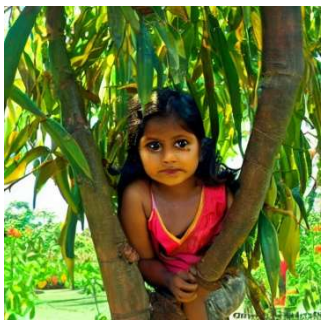
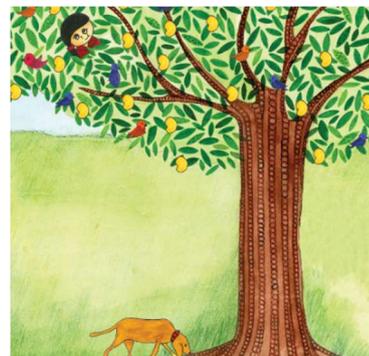

*Figure 19a) Prompt - "Indian childrens book style, I like to hide in mango tree. No one can see me here." Seed - 1234*

*Figure 19b) Original image taken from page 1 of The Mango Tree [52]*

On the other hand, the style shown in the example in Figure 20a is more representative of most of the images produced by the model for this example book as can be seen in Appendix B. Once again, this figure illustrates that the model struggles to draw faces even in non-realistic styles however it is less noticeable when viewed in this style. We can also see once again the **skin tone is more similar to that found in India** and the clothing is of a similar style to that found in the reference image in Figure 20b. However, it is difficult when looking at Figure 20a to discern which of the characters is supposed to be the grandmother, whereas Figure 20b clearly illustrates the grandmother character by her grey hair.

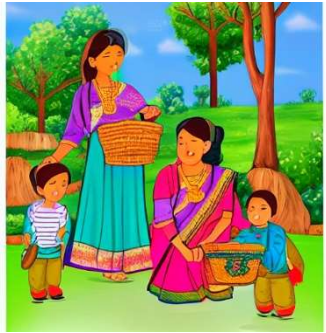
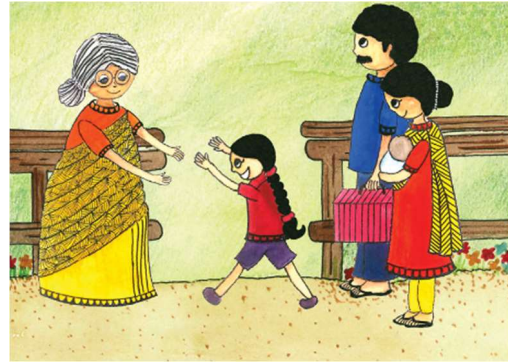

*Figure 20a) Prompt - "Indian childrens book style, We like to visit Grandma's village." Seed - 1234*

*Figure 20b) Original image taken from page 3 of The Mango Tree [52]*

The example in Figure 21 is geo-located to the UK however, testing showed that simply using 'childrens book' instead of 'UK childrens book' still resulted **in a good performance since Stable Diffusion natively uses the English language**. By not including the keyword UK, it allows for more tokens to be used in the prompt content. Figure 21a) also shows how Stable Diffusion struggles to generate the birds faces, and while from a distance it is easy to recognise birds, upon closer inspection some of the birds show irregularities, specifically the one in the centre left of the image.

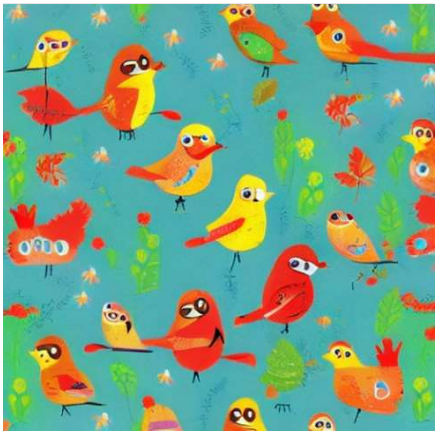
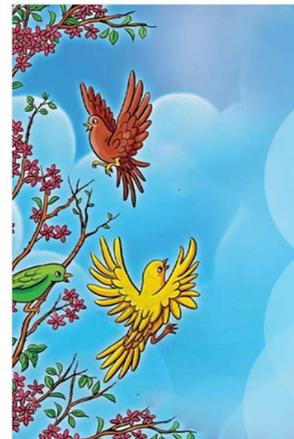

*Figure 21a) Prompt - "childrens book style, I can see birds." Seed - 1234*

*Figure 21b) Original image taken from page 6 of In My Garden[67]*

As can be seen in Figure 22, the art style is clearly different from those generated for the previous books and the character also looks to be more like something that would be found in a Japanese story book, especially when compared to the characters shown in Figures 19 and 20. However, while the style and characters are geo-located correctly, the illustration does not necessarily capture all of the details of the text. The characters hands are incorrect as there appears to be multiple hands overlaying one another, and there is no illustration of the gold coins which are mentioned or anything to do with the fishing. This is likely a result of **words at the end of prompts being weighted less heavily than words at the start of the prompt**. However, it is important to note that it is not always the case that the illustration in a book encapsulates everything which was mentioned on that page, since it is meant to be an accessory to the text, and not a replacement.

Additionally, in the lower right-hand corner of Figure 22a) there is a banner which resembles Japanese text. The model would occasionally produce images which contained text or characters which resembled text for a variety of cultures. Using phrases such as manga or anime instead of Japanese story book, did result in more characters in the aforementioned art style. However, it also included more instances of text or characters being generated, which ruined the overall quality of the images since these were usually nonsensical, **or not proper Japanese characters.**

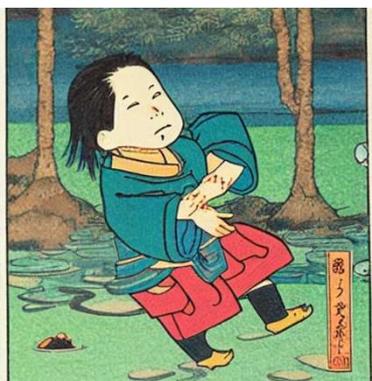

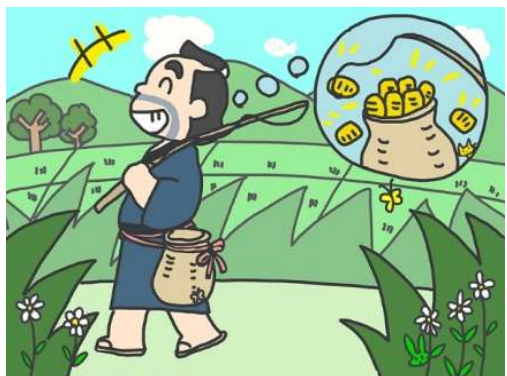

*Figure 22a) Prompt - "japanese children book style, Hachigoro heard that someone had caught 50 gold coins, fishing." Seed - 1234*

*Figure 22b) Original image taken from page 1 of Not You[68]*

The image in Figure 23 shows a problem which not only affects this method, but also the other 2 which are discussed later in this report, which **is character consistency**. While the original images clearly depict the same character throughout the story, this is not the case for the model generated images since each image has no context about the images generated previously.

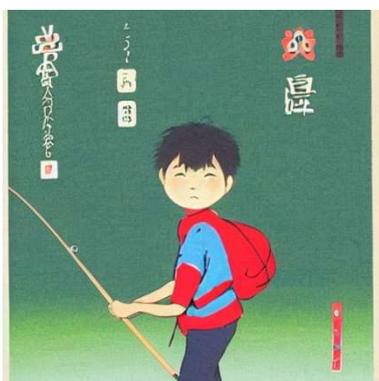

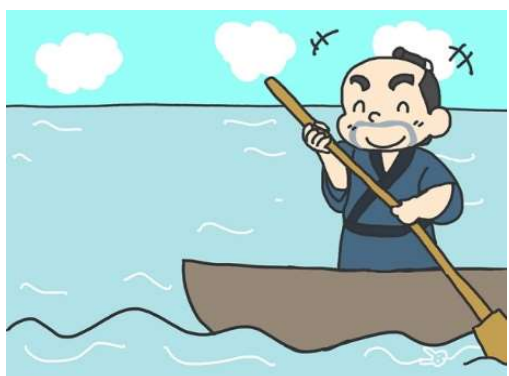

*Figure 23a) Prompt - "japanese children book style, So he left home right away, carrying fishing pole on his shoulder." Seed - 1234*

*Figure 23b) Original image taken from page 2 of Not You[68]*

Figure 24 once again shows the issue of the model attempting to generate realistic looking images instead of the children's book style. Like the example in Figure 19, the model especially **struggles to generate realistic looking eyes**, in this instance, not generating them at all. However, when comparing to the sample image in Figure 24b, it is clear that the key themes and content from the text is represented in the generated image as it includes a large fish which is the same colour in both examples, as well as someone who is clearly a fisherman, and a boat.

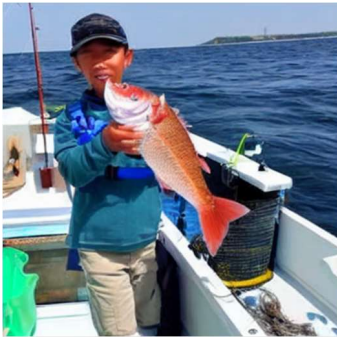

*Figure 24a) Prompt - "japanese children book style, As soon as he put out to sea in boat, dropped his line in water, big, fine-looking sea bream got caught." Seed - 1234*

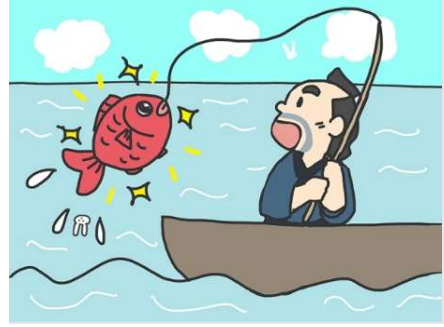

*Figure 24b) Original image taken from page 4 of Not You[68]*

This next section showcases the results when one book, Riley Flies a Kite [51], which is suitable for UK audiences, is converted into 3 different cultures: Japanese, Middle Eastern and Indian. This resembles how the technology would be used in the real world.

Figure 25 shows how using the same seed and very similar prompts can produce very similar images with slightly different content. **For example, the skin tone of the children matches what would be familiar from books in the respective regions with Japanese being the lightest skin tone, then Middle Eastern, then Indian**. Also, the kite, while a little visually confusing, since it appears to be made of 2 separate parts (or possibly two different kites), is consistent through all 3 images with slight stylistic changes.

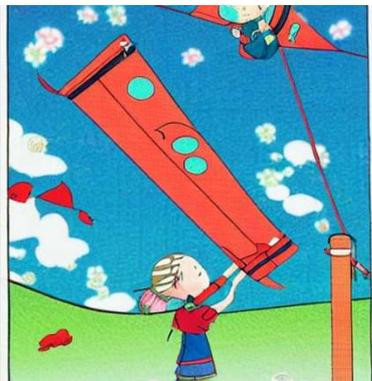

*Figure 25a) Prompt - "japanese childrens book style, Where could Riley fly this kite?" Seed - 1234*

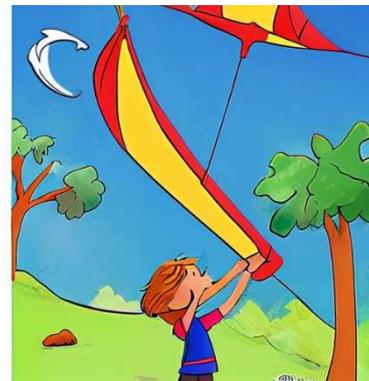

*Figure 25b) Prompt - "middle east childrens book style, Where could Riley fly this kite?" Seed -1234*

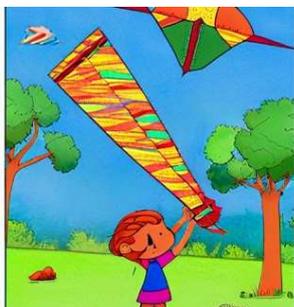

*Figure 25c) Prompt - "indian childrens book style, Where could Riley fly this kite?" Seed - 1234*

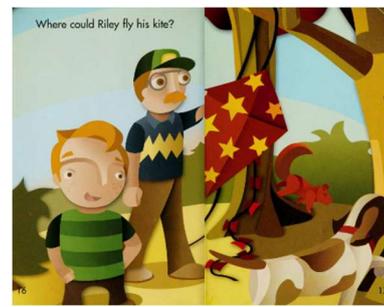

*Figure 25d) Original image from page 8 of Riley Flies a Kite [51]*

Looking at another example, Figure 26b shows the recurring issue with prompt being geo-localised to India attempting to produce realistic images, which is not the case for the Middle Eastern and Japanese examples, shown in Figure 26a and 26c respectively. Both these examples are similar to one another but once again, have the common issue when illustrating eyes. Interestingly, none of the generated prompts interpret the park as an area with swings and play areas for children but rather as an open field. The original image shown in Figure 26d shows a park with a swing set, as well as including a car as the chosen mode of transport and the family dog in the back seat. Clearly, the model will never generate these if they are not explicitly mentioned in the prompt.

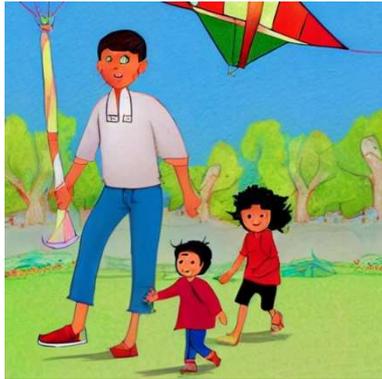
*Figure 26a) Prompt - "middle east childrens book style, Riley, his dad took kite to park." Seed - 1234*

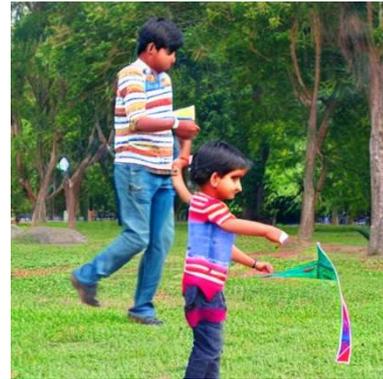
*Figure 26b) Prompt - "indian childrens book style, Riley, his dad took kite to park." Seed - 1234*

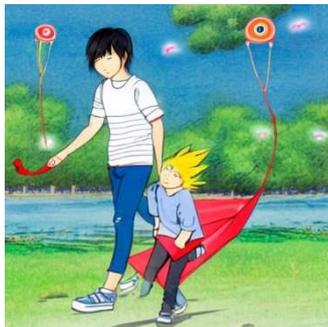
*Figure 26c) Prompt - "japanese childrens book style, Riley, his dad took kite to park." Seed - 1234*

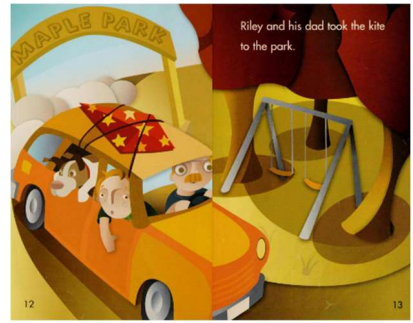
*Figure 26d) Original image from page 6 of Riley Flies a Kite [51]*

## 4.2  Method 2 Observations

As demonstrated in Figure 27, this method also struggles with the same problems as method 1. In particular, when generating images for **Indian audiences**, the model will create a more **realistic picture**. It also **incorrectly generates faces**, giving the girl an **extra eye** on her forehead. However, like method 1, the girl has also been given a **darker skin tone** common in India.

While **the key objects of the girl and the tree are shown**, the scene is conveyed worse in this method than in method 1, since the girl is only standing next to the tree rather than hiding in it. The generated image is also much **less colourful** than that of method 1, resulting in a **less preferrable image for a children's book.**

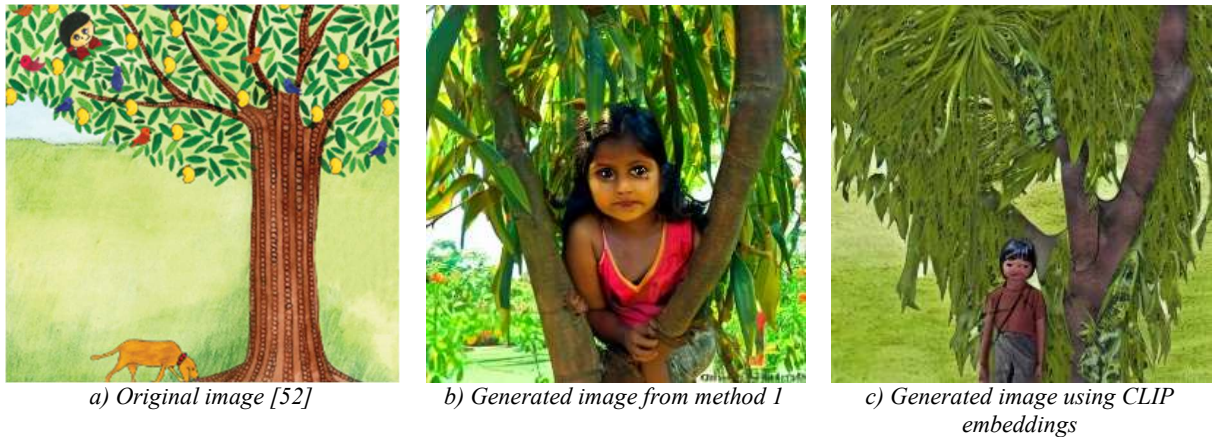

*a) Original image [52]*     *b) Generated image from method 1*     *c) Generated image using CLIP embeddings*

Figure 27: The Mango Tree page 3. The corresponding text reads "I like to hide in the mango tree. No one can see me here."

Figure 28 shows an **improvement** in both the **art style** and conveying the **same message** as the corresponding sentences ("Grandma has a big mango tree in her garden. Many birds come there."). The grandmother is clearly depicted in a **sari** and **her hair is grey**, allowing her age to be easily conveyed. However, her grey hair is not shown in other pictures of the same book, since this **method struggles with consistency**.

The image generated with this method (Figure 28c) is also **not as effective** as that of method 1 (Figure 28b). In Figure 28b, the tree in the background has been **drawn with mangoes**, whereas in Figure 28c, the tree in the background is **fruitless**. If a child did not understand the word 'mango' in the book, they would not be able to learn that it is a fruit through seeing the image generated by method 2. Figure 28b is also **brighter** and has **more detail** in the tree's leaves and the grandmother's sari, so it can be concluded that method 1 would be more appropriate for a children's book.

In both generated images, **the grandmother's eyes have not been drawn at all**. However, because of the **simplistic art style**, this **does not ruin the images** as much as others, in which the model has generated extra or distorted eyes.

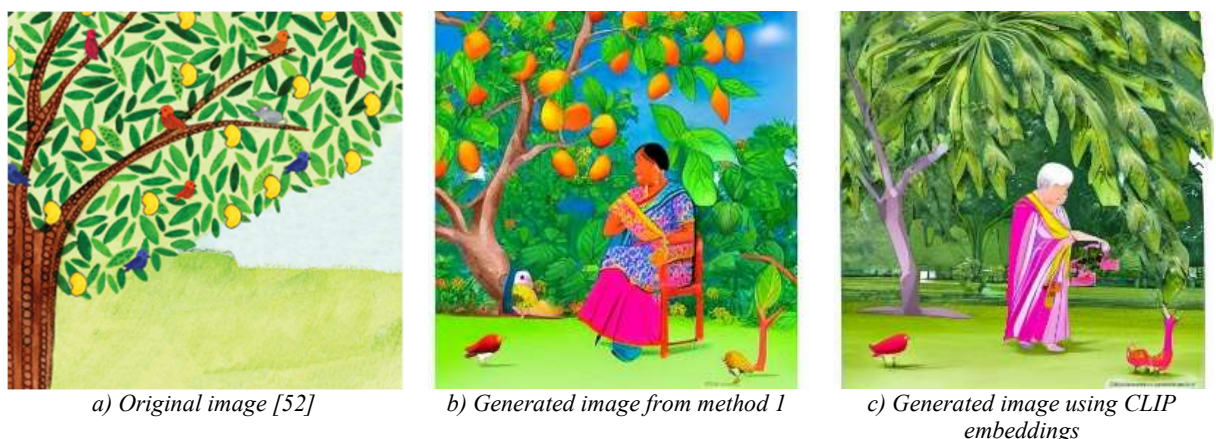

*a) Original image [52]*     *b) Generated image from method 1*     *c) Generated image using CLIP embeddings*

Figure 28: The Mango Tree page 2. The corresponding text reads "Grandma has a big mango tree in her garden. Many birds come there."

As shown in Figure 29c, when creating images geo-localised for Japanese audiences, this method emulates the target art style quite well, by using **bold black outlines** around the character and giving him a **traditional Japanese outfit**. However, the image is **not as**

**colourful** as the picture generated using method 1 (Figure 29b) and the **background does not look finished**. Like method 1, the image generated using this method does not contain as much context as the original image (Figure 29a), which shows the main character, Hachigoro, carrying a **fishing pole** thinking about the **gold** someone had caught while fishing. There are also characters that look like **Japanese kanji** in the bottom right corner, which could translate to something **inappropriate** or might **not be actual characters** at all.

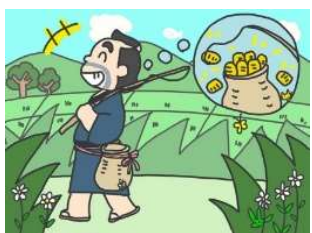 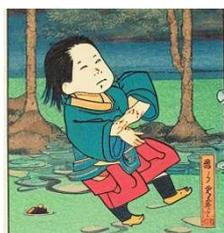 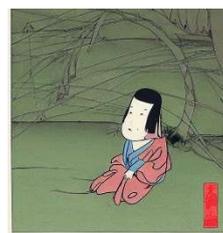

*a) Original image [23]*  *b) Generated image from method 1*  *c) Generated image using CLIP embeddings*

*Figure 29: Not You! page 1. The accompanying text reads "Hachigoro heard that someone had caught 50 gold coins while fishing."*

For page 4 of "Not You!", the image generated from method 1 (Figure 30b) is too realistic for a children's book. On the other hand, the image in Figure 30c has **obvious but detailed linework**, which is very reminiscent of well-known **Japanese Ukiyo-e paintings** [69]. However, the general idea of the accompanying sentence is lost since only the fish is shown in the drawing, and both the boat and fisherman are completely omitted. The bright red colour of the sea bream in both the original image (Figure 30a) and in the image generated using method 1 (Figure 30b) is also not present in Figure 30c.

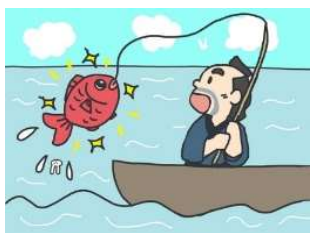 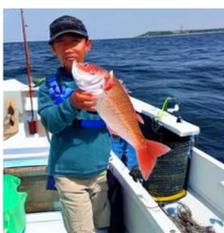 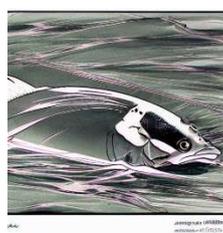

*a) Original image [23]*  *b) Generated image from method 1*  *c) Generated image using CLIP embeddings*

*Figure 30: Not You! page 4. The accompanying text reads "Hachigoro, however, let the sea bream off the hook and threw it back to the sea saying, 'Darn it. Not you'"*

Figure 31 demonstrates one of the main problems for **all three methods**: when the input sentence references something from a **previous page**, the generated images are unable to draw a suitable image since they do not have this context. In previous pages of the book, it states that Riley is **making a paper kite** but the model, which is only given the text for one page at a time, has **no context** for the kite so instead generates images of people wearing red. However, it is obvious that the generated images are still geo-localised for their respective cultures as the **skin tones, clothes** and **hairstyles** are varied to fit their target cultures.

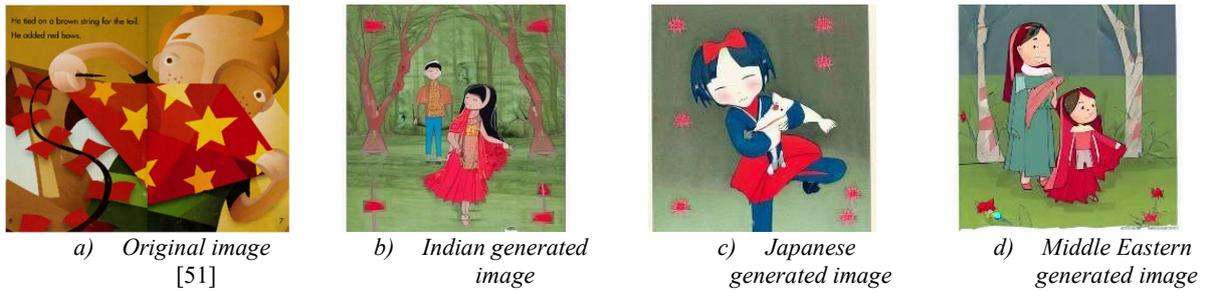

*a) Original image [51]*    *b) Indian generated image*    *c) Japanese generated image*    *d) Middle Eastern generated image*

Figure 31: Riley Flies a Kite page 3. The accompanying text reads "He added red bows."

Figure 32 shows the generated images of page 11 of "Riley Flies a Kite", which were geo-localised for the three different cultures. Each picture clearly depicts a child flying a kite, changing their skin tones to fit the target culture, with the boy in the Indian picture (Figure 32b) having the darkest skin. However, the child in the Middle Eastern picture (Figure 32d) has a **lighter skin tone** than expected and the hand holding onto the kite is **very distorted**. Although Figure 32c follows the Japanese art style quite well, the girl depicted appears to have **3 hands** holding the kite. Additionally, all three of the generated images have completely omitted the football field depicted in the original (Figure 32a). This might be because the SD model gives higher importance to tokens at the beginning of the prompt so the second sentence was considered less relevant.

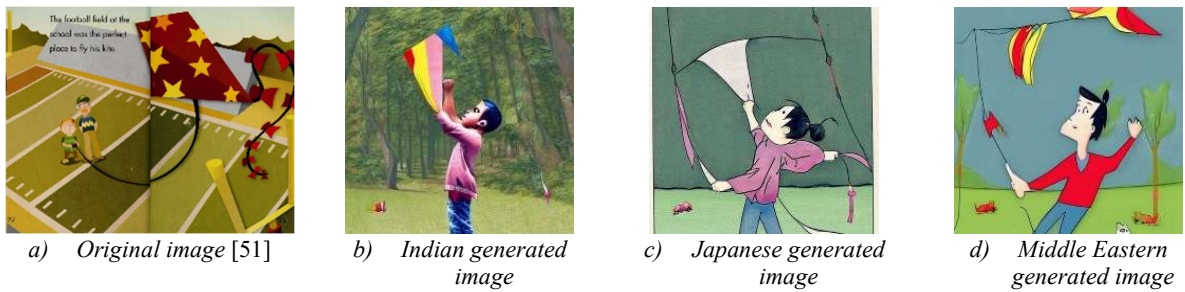

*a) Original image [51]*    *b) Indian generated image*    *c) Japanese generated image*    *d) Middle Eastern generated image*

Figure 32: Riley Flies a Kite page 11. The accompanying text reads "He knew a great place to fly his kite. The football field at the school was the perfect place to fly his kite."

## 4.3 Method 3 Observations

The generated images for method 3 and 1 were compared, as both methods heavily rely on the use of the prompts into the AI model. Here, the effect of the parameters such as "prompt_edit_spatial_end", "prompt_edit_token_weights", and "prompt_edit_spatial_start" are discussed, and how they can be tuned to generate better results.

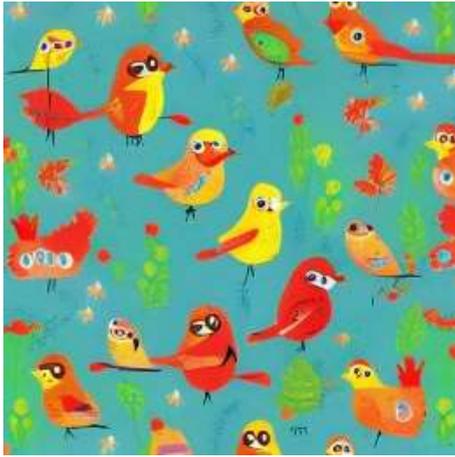 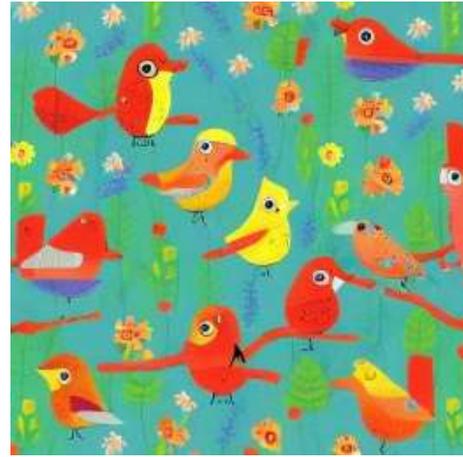

a) *Method 1 with Initial Prompt: "childrens book style, I can see birds."*   b) *Method 3 with Editorial Prompt: "childrens book style, I can see English birds."*

Figure 33: Images Generated with Method 1 and 3 for British interpretation of In My Garden, Page 4

By looking at Figure 33a) and b), it can be seen that method 3 creates more refined, legible images. This was a common occurrence in the generated images, which reinforces the claims made by Hertz et al [66]. Between these two images, when the "English" keyword is added, the images in b) are more contextually defined as being birds, by having more bird features, such as wings and beaks, as well as two feet. Furthermore, flowers were added in the background, instead of being misshapen patches of colour, which can tie in with the English countryside being more correlated with having more flora.

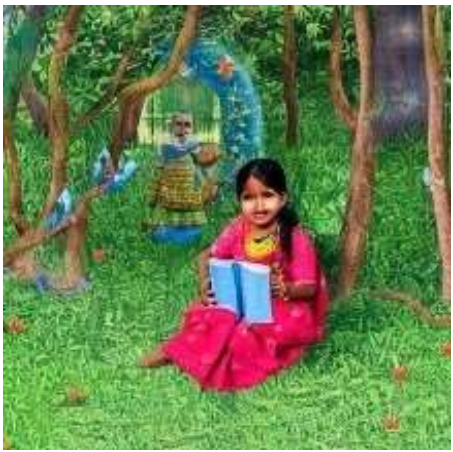 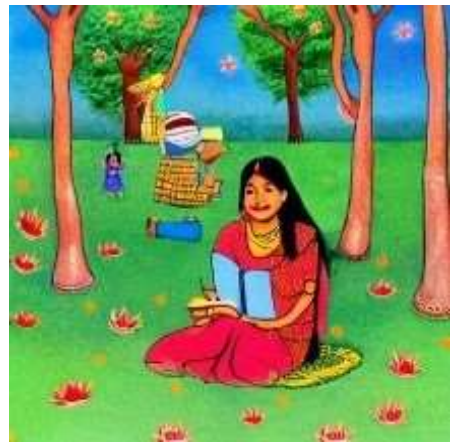

a) *Method 1 with Initial Prompt: "Indian childrens book style, He needed place without trees, he needed breeze."*   b) *Method 3 with Editorial Prompt: "Indian childrens book style, He needed Indian place without Indian trees, he needed Indian breeze."*

Figure 34: Images Generated with Method 1 and 3 for Indian interpretation of Riley Flies a Kite Page 9

In the UK book Riley Flies a Kite, the prompt for the sentence "He needed a place without trees," causes the image generated to focus on the word "trees," and instead generates an abundance of them. This is very contrary to what the original sentence is conveying, which is rectified in Method 3's editorial prompt, where the Indian keyword is inserted multiple times to each important word, such as: place, trees and breeze. The resultant image then had a reduction in foliage, and became more stylized and representant of a children's book. As mentioned in Section 3.4, the "prompt_edit_spatial_end" parameter dictates the model's creativity for larger details, and this is a good example of how it creates a difference. Even though this was the maximum amount that the parameter could be set at, it would be better if there was a way to further increase this value to yield better results.

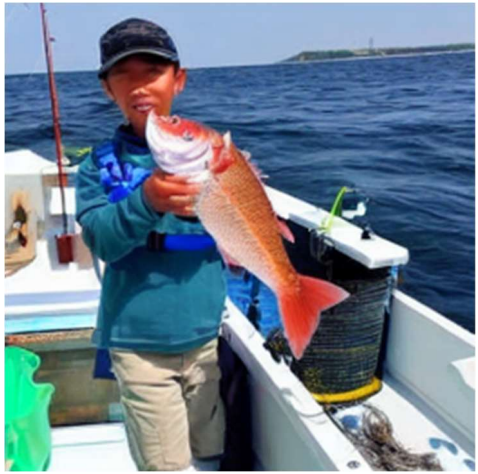 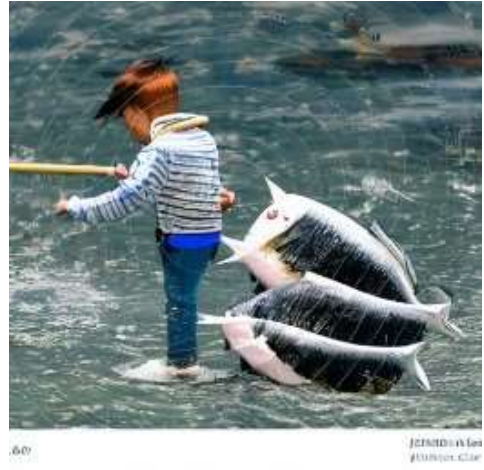

*a) Method 1 with Initial Prompt: "japanese children book style, As soon as he put out to sea in boat, dropped his line in water, big, fine-looking sea bream got caught."*

*b) Method 3 with Editorial Prompt: "japanese children book style, As soon as he put out to sea in Japanese boat, dropped his Japanese line in Japanese water, big, fine-looking Japanese sea Japanese bream got caught."*

*Figure 35: Images Generated with Method 1 and 3 for Japanese interpretation of Not You! Page 4*

However, large changes in the total global image generate a worse performance. Figure 35a) is a clear image that matches the prompt well, with good composition that degrades once the editorial prompt is applied, with the Japanese keyword being used. The resultant image is more unrealistic, and so the "prompt_edit_spatial_end" parameter should be decremented, preferably below 0, to improve cases such as this. This could also be caused by the length of the prompt, and the multiple repetitions of the keyword may have confused the model, all in all generating a case scenario where the prompt processing may fail.

Additionally, Method 3 generates an image that has the absence of a boat, which is contradicts the prompt sentence. Here, utilising weights of the importance of each token represented by the "prompt_edit_token_weights" array, would be an essential improvement, as increasing the importance of the token "boat" would mean it had a greater chance of manifesting in the image. Unfortunately, due to the primitive prompt processing, weights for key tokens in the prompt were not able to be identified.

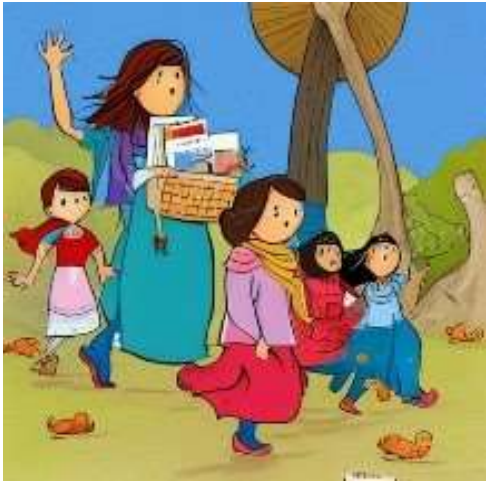 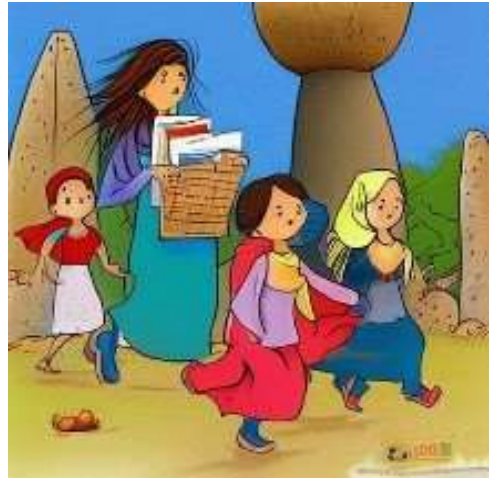

a) Method 1 with Initial Prompt: "middle eastern childrens book style, The wind did not blow."

b) Method 3 with Editorial Prompt: "middle eastern childrens book style, The middle eastern wind did not blow."

Figure 36: Images Generated with Method 1 and 3 for Middle Eastern interpretation of Riley Flies a Kite Page 4

Figure 36b) shows a successful culturally sensitive image generated, which method 1 failed to do. Here, method 3 adapted the image from method 1, changing the 2 female characters to be in a hijab, as well as making a slight change to their skin colour. As mentioned in Section 3.4, the "prompt_edit_spatial_start" parameter dictates the model's creativity for smaller details, and in this case, the implemented value for this parameter was set to the minimum value 0.0. Due to what has occurred in the generated image, this parameter could perhaps be increased to give the model a greater chance of covering all the female characters with hijabs.

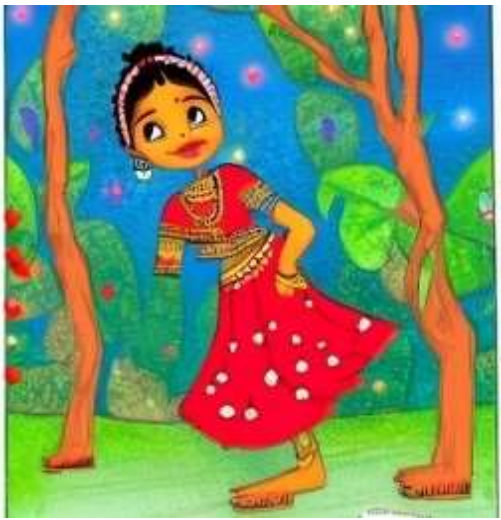 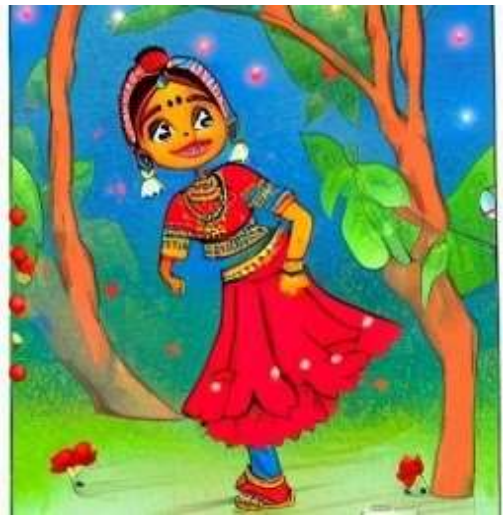

a) Method 1 with Initial Prompt: "Indian childrens book style, He added red bows."

b) Method 3 with Editorial Prompt: "Indian childrens book style, He added red Indian bows."

Figure 37: Images Generated with Method 1 and 3 for Indian Interpretation of Riley Flies a Kite Page 3

There may also be cases where there are too much unnecessary details added to the generated images. In the case of Figure 37b), more details are added to her dress, and facial features. This realism causes the image to lose the children's book style that is preferred. In a scenario like this, these fine details can be reduced by decreasing the "prompt_edit_spatial_start" parameter. However, this is not possible, as it is already set at minimum value. Instead, improvements should be made to the prompt processing, to minimise the risk of repeating this mistake.

# 5 Evaluation

Because the quality of a picture is generally subjective, this project is primarily **evaluated qualitatively**. Due to the feedback received during the seminar in November, it was decided that the evaluation process should involve external appraisal, more specifically from those that were **more likely to recognise and be familiar with the art styles** from the target nationalities. Hence, an online questionnaire was created in Microsoft Forms.

It was identified that there exists an increased risk of a **biased evaluation** from a **limited sample size** survey. Therefore, an **unsupervised quantitative evaluation** using **Frechet Inception Distance** (FID) [39] was conducted in parallel with the questionnaire. This calculates the similarity between the original and generated images from the same input text.

## 5.1 Success Criteria

The success of this project is determined by how well the generated pictures assist with making a given story more **understandable** to children of different cultures. Each generated image will be assessed by how well it conveys the **same scene** as the corresponding text and whether it contains any **distortions** or **inappropriate** sections. The **geo-location feature** will also be judged by **comparing the images to art styles of the target culture** and checking if someone from the target culture could understand the given story **without needing to read the text**.

## 5.2 Qualitative Evaluation: Online Questionnaire

Since this questionnaire would include personal data being collected and/or processed, an ethics application form was submitted with all the necessary forms, participant and consent information. The study included participants, students from the University of Southampton, that are in the Japanese, Indian and Middle Eastern Societies, and have these countries as their national or ethnic origin.

After the consent page, the participants are asked a series of personal questions, such as their gender, age, and their level of cultural knowledge for the chosen country, where the latter was written using [20] as a basis. A 5-point Likert scale [19] was used to rank the images, as dichotomous questions would not give a useful measurement of how best suited the images are to the prompt, as these are not objective measurements.

There are 2 parts to the questionnaire. Firstly, participants ranked the generated images (depending on their ethnic/national origin) from the chosen UK Book, Riley Flies a Kite, where its images have been translated into either an Indian, Japanese and Middle Eastern art style. All in all, they had to rank a total of 4 photos: the first is the original photo which is kept as a control variable, and the other 3 are the best photos generated using each of the 3 methods. They then ranked each generated photo as to how close it is to the art style that they are familiar with.

The second part consisted of testing the AI generated images with the original text. Here, the participants needed to rank the AI generated images based on how well they matched the original text from the book. Participants were given a book based on their ethnic/national origin, where its images have been translated into that culture's art style, using 3 different methods. All this was done using the Likert scale provided. An excerpt of two of the questions from Part 1 and 2 of the questionnaires for the Indian art-style images is pictured below, in Figure 38. An example of the full questionnaire is provided in the Appendix.

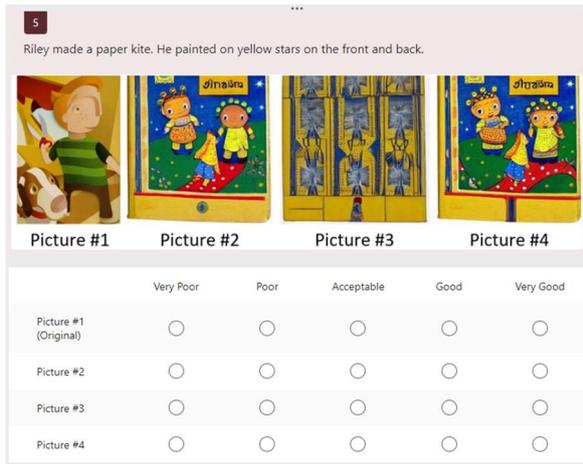

*Figure 38a): An example for how Part 1 of the questionnaire is, the images from the UK book have been translated into an Indian art style and now must be ranked.*

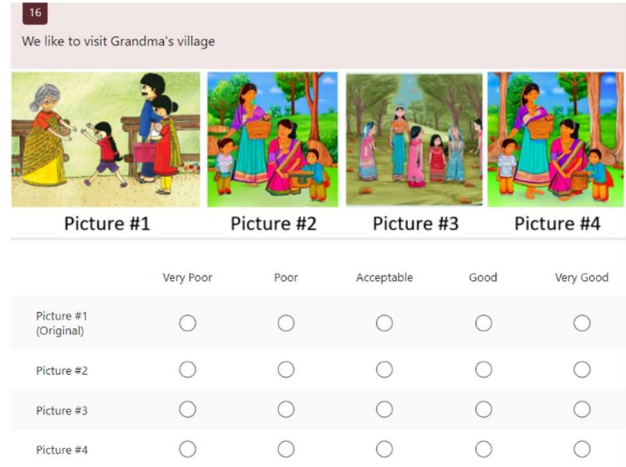

*Figure 38b): An example for how Part 2 of the questionnaire is, the images from the Indian book have been generated by the AI into an Indian art style, and the best out of these must now be ranked.*

### 5.2.1 Questionnaire Results

In order to analyse the results of each page, each rating on the Likert scale was given a numerical value (Very Poor=-2, Poor=-1, Acceptable=0, Good=1, Very Good=2). The responses were then used to calculate the **average rating** for each method. For each question in the survey, picture #1 corresponded to the original image and pictures #2, #3 and #4 related to method 1, 2 and 3 respectively. The **mean** ratings for each **individual book** were then calculated for each method.

Because the questionnaires were only approved for distribution slightly before the start of the Christmas break, **most students were unavailable**, including those in the appropriate culture societies. This meant that only the questionnaire sent to the **Japanese Society** had enough responses for it to be used to evaluate the project's ability to geo-localise images.

In the first section of the questionnaire, the participants were asked to rate the images corresponding to pages of the UK book "Riley Flies a Kite" [51] when localised for Japanese audiences. The results are summarised in Table 2. All methods, including the original illustration, were given an average rating close to 0, corresponding to a rating of "**Acceptable**". The original book was generally rated the highest with a mean value of 0.325. Out of the methods described in this report, **method 3 was the best** with a rating of 0.234 and **method 2 was the worst** with a rating of -0.091.

| Page | Original | Method 1 | Method 2 | Method 3 |
|---|---|---|---|---|
| 1 | -0.286 | 1.286 | -0.143 | 0.571 |
| 2 | 0.000 | -0.143 | 0.286 | 0.714 |
| 3 | 0.429 | -0.857 | -0.286 | -1.000 |
| 4 | 0.143 | -0.429 | 0.714 | -0.429 |
| 5 | 1.000 | 0.000 | -0.857 | -0.286 |
| 6 | 0.571 | 0.571 | 0.000 | 0.857 |
| 7 | 0.429 | -0.429 | -0.143 | 0.286 |
| 8 | 0.143 | 0.429 | -0.286 | 0.429 |
| 9 | 0.429 | 0.714 | -0.429 | 0.571 |
| 10 | 0.286 | -0.143 | 0.429 | 0.143 |
| 11 | 0.429 | 0.571 | -0.286 | 0.714 |
| **Average** | **0.325** | **0.143** | **-0.091** | **0.234** |

*Table 2: Results for Riley Flies a Kite from Japanese questionnaire*

Figure 39 shows that for page 6 of Riley Flies a Kite the generated pictures were generally rated as highly as the original. This might be due to the fact that the accompanying text describes quite a **simple scene**, which can easily be drawn by the Stable Diffusion model. Therefore, all three generated images clearly depict a child and an adult with a kite in the park. Picture #2 (generated by method 1) was rated slightly higher than the others, including the original image, which is possibly due to the **brighter colours** and an art style similar to **manga and anime**, which would be very appropriate for a child's Read for Pleasure book. Additionally, Picture #4 seems **too realistic** and the kite in Picture #3 is **not as easily identifiable**.

Most the characters in the generated pictures are also depicted with **paler skin** and **darker hair**, which are very common attributes of Japanese people. Because 97.9% of the Japanese population are Japanese (CIA World Factbook [70]), this would allow most people in Japan to associate with the characters of the story more easily but might also alienate those not in the majority if every generated character looks the same. Notably, method 1 depicts Riley with **blonde hair**, which could allow for a sense of **diversity** in the images.

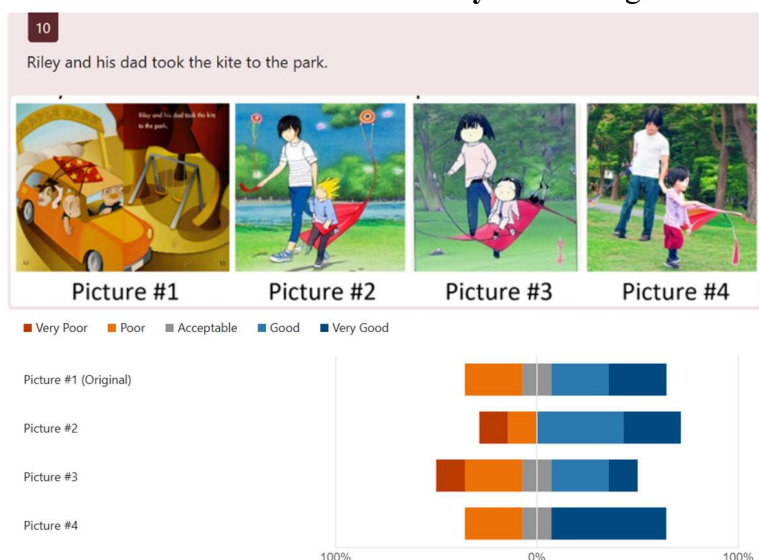

*Figure 39: Evaluation results of page 6 of Riley Flies a Kite when translated for Japanese audiences.*

As demonstrated in Figure 40, all three models struggle when a pages' corresponding text refers to previous pages. On page 3 of Riley Flies a Kite, the text reads "He added red bows". Because the original illustration had the context that Riley was making a paper kite from the previous pages, it was able to make a more relevant image. This is supported by the results in Figure 40 as the majority of participants (57.2%) rated it good or very good. On the other hand, all of the generated images were generally rated poor or very poor (61.4%, 57.2% and 85.7% for methods 1, 2 and 3 respectively). This is because, although the images seemed closer to the target Japanese art style, they did not contain any reference to the kite and instead drew red bows on a child's head. Both pictures #2 and #4 (generated by methods 1 and 3 respectively) were rated much worse than picture #3 since the **girl's eyes have been distorted**, which could **scare** and **upset** the reader if it was used in a children's book.

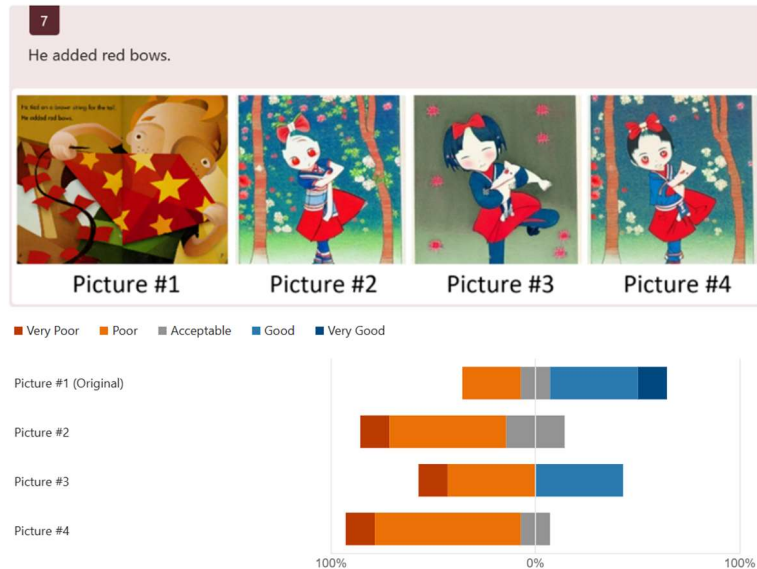

*Figure 40: Evaluation results of page 3 of Riley Flies a Kite when translated for Japanese audiences*

Figure 41 shows that the images generated by method 1 and 3 (pictures #2 and #4 respectively) have been rated higher than the original image. This supports the fact that the methods can effectively create appropriate images with **art styles familiar to Japanese audiences**. Each of the generated images depict a child with **pale skin** and **dark hair** flying a kite. However, picture #3 was rated lower than the other images, including the original, as the girl depicted has an **extra arm** and the whole image is **not as colourful**. Pictures #2 and #4 are very similar, except the latter had more ratings that were good or very good (57.2% compared to 42.9%). This could be since the child has a **shorter haircut**, better conveying they are a boy, as suggested by the 'he' pronouns.

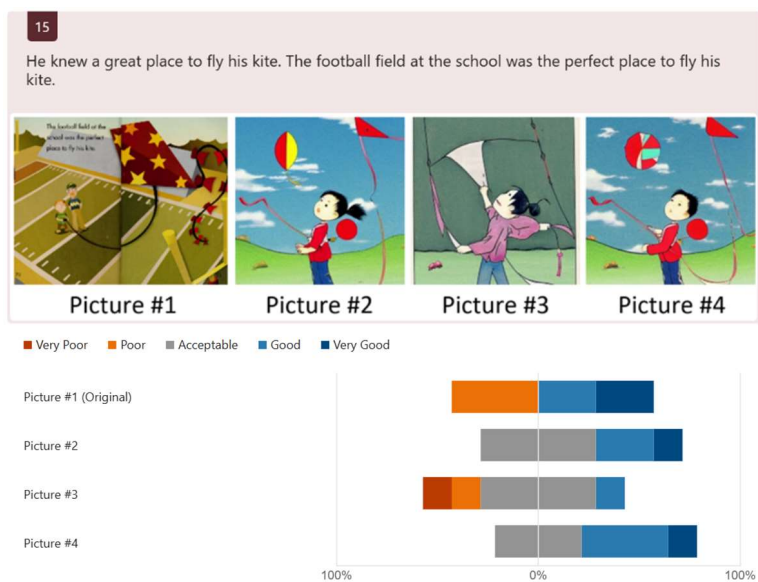

*Figure 41: Evaluation results of page 11 of Riley Flies a Kite when translated for Japanese audiences*

The second section of the questionnaire asked the participants to evaluate the images generated for a Japanese book, "Not You!" [23]. The mean ratings of each page are given in Table 3. Similar to the results related to "Riley Flies a Kite", the generated images were generally given an "Acceptable" rating, and method 2 was regarded the worst out of the three with an average rating of -0.107. Unlike the results for "Riley Flies a Kite", **method 1 was valued higher than**

**method 3** on every page except page 1. The original illustrations for the book were also given mostly "Good" or "Very Good" ratings, as shown by the mean value of 1.571. This is much better than the pictures from "Riley Flies a Kite" as the images were **already created for Japanese audiences**. They were also able to **contain more relevant features** since the human illustrator could decide which parts of the corresponding sentence were important.

| Page | Original | Method 1 | Method 2 | Method 3 |
|---|---|---|---|---|
| 1 | 1.857 | 0.429 | -0.143 | 0.571 |
| 2 | 1.286 | 1.000 | 0.143 | 0.571 |
| 3 | 1.571 | -0.286 | 0.143 | -0.857 |
| 4 | 1.571 | 0.286 | -0.571 | 0.000 |
| **Average** | **1.571** | **0.357** | **-0.107** | **0.071** |

*Table 3: Results for Not You! from Japanese questionnaire*

For page 2 of "Not You!", the generated pictures depict a boy while the original image displays Hachigoro as an adult. Nevertheless, this **does not seem to have affected the results** since method 1 (Picture #2) was rated quite highly alongside the original image (Picture #1), as shown in Figure 42. The generated image clearly shows a **boy carrying a fishing pole** and the **backpack** also suggests he has left home, as described in the initial text. Since all the relevant information has been conveyed, the **character's age**, which is not mentioned in the story, has **not resulted in a significant effect** in the picture's appraisal.

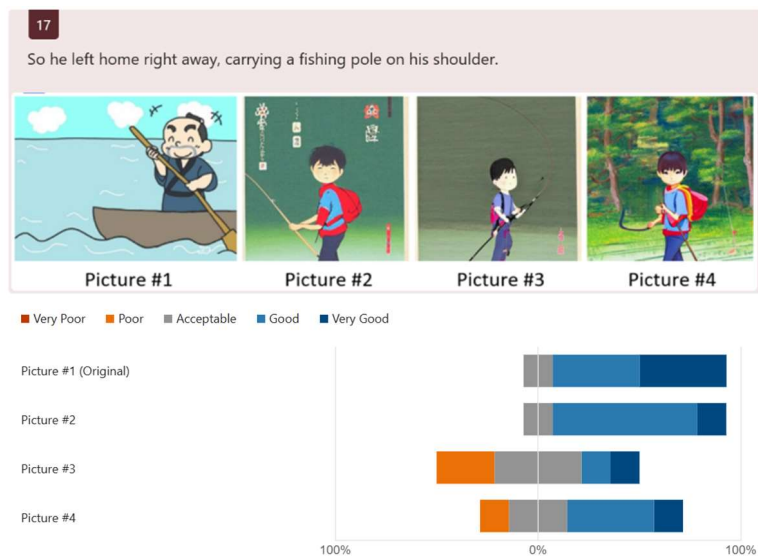

*Figure 42: Evaluation results of page 2 of Not You! when translated for Japanese audiences*

In the final section, the participants were asked about the necessity of having artwork that represented their own culture in children's books. As shown in Figure 43, the responses ranged from neutral to very important, with **no negative responses** and the majority considering it to be somewhat important. Because of this, the customer would be pleased as there is a **definite interest for geo-localising artwork** and therefore a reason to continue the work described in this report.

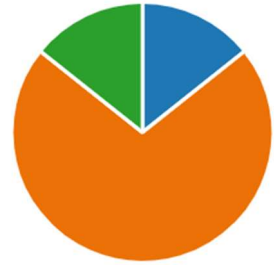

20. How important is it to you to have artwork from your own culture in children's books?

- Very important: 1
- Somewhat important: 5
- Neutral: 1
- Somewhat not important: 0
- Not important at all: 0

*Figure 43: Questionnaire Results exploring the importance of geo-localised artwork*

## 5.3 Quantitative Evaluation: Frechet Inception Distance

Quantitative evaluation was conducted in order to numerically assess performance without dependence on the subjective views of members of the public. The approach follows the industry standard **Frechet Inception Distance (FID)** [39] metric, which calculates the **distance between two representational feature vectors**. Both real and generated images must be mapped to the same vector space by a single model, to enable this comparison. Next, the average distance between all images in two sets of images is calculated. A minimal FID score is optimal, as it shows the two sets are similar. Although CLIP image encoder may be utilised for this purpose, it is already in use as a prerequisite to the LDM model, which may introduce a bias given the significant dependence on it. As a result, the **Inception V3 model [71]** will be implemented, given its favourable performance, in addition to minimal cross-over with CLIP. Furthermore, the **increased vector size of 2048** enables retainment of higher levels of detail and promotes minimal abstraction, which is favourable in an evaluative model. On the other hand, images require transformation into 299x299 in order to be inputted into the model, reducing the details given to the model.

Method 1 simply augmented the prompt with keywords prior to image generation. Additionally, other stylised models explored in section 6.1.1, such as Disney and Japanese, were also compared using the same prompts. Method 2 applied a mask to CLIP embedding, whereas Method 3 utilised cross-attention mechanisms to refine the original prompt with another. The FID score was calculated against each of the generated images and the original.

| Method | FID Score (against original) | Number of Images |
|---|---|---|
| Method 1 | 277.24 | 35 |
| Method 1 (with stylised models) | 251.22 | 90 |
| Method 2 | 273.90 | 35 |
| Method 3 | 284.68 | 35 |
| All Methods | 249.13 | 160 |

*Table 4: Tabular representation of FID score for each method surveyed, alongside number of images used.*

**Method 2 was closest to original with an FID score of 273.90. Although Method 1 features an FID score of 277.24, with the addition of the 3 stylised models explored in Section 7.1.1 an even lower FID score of 251.22 was produced. Method 3 varied the most from the original image, having the highest FID score of 284.68.** As the FID score for all images is lower than any of the individual methods and therefore cannot be representative of the average

generated image. This indicates that an insufficient amount of samples were supplied for an accurate evaluation, given that **FID is better suited for large sample sizes.**

## 5.4 Summary

Overall, all three methods are quite good at generating images for pages accompanied by **simple but descriptive input text**. However, for inputs that include **negatives** (e.g. 'not', 'without'), all three methods tend to ignore the negative token and **generate a picture showing the opposite**. Sentences that refer to **information given on a previous page** also result in **less accurate pictures** due to the loss of context.

Occasionally, the SD model will also generate images containing **distortions**, particularly a characters' **face** and **eyes**. This problem is more noticeable if the image has been generated **too realistically** for a children's book, which occurred more often when creating books for **Indian audiences**.

Nevertheless, all three models can successfully geo-localise artwork, by generating images that contain **features relating to the target culture**. This is especially demonstrated using **hairstyles, skin tones and outfits** common in that area of the world. Moreover, these features are still depicted in the more realistic images.

Unfortunately, due to time constraints, the **questionnaires** were only able to receive a **small number** of responses. Ideally, more people would have completed each survey, which would have resulted in a more **accurate representation of the public** and would have also allowed us to suitably assess the methods across all three cultures. Nevertheless, the limited responses still indicate that the **traditionally illustrated images are generally better** than all three methods, especially for books that originate from the target culture. Method 2 was rated lower than the other two on average, which may be due to the generated images normally being **less colourful**, so less suitable for a children's book. Both methods 1 and 3 were rated similarly, with **method 3** being rated slightly higher for the **UK book** and **method 1** having a slightly higher rating for the **Japanese book**. The questionnaire results also show a **distinct interest** in AI-generated artwork being used in geo-localising children's books.

The **quantitative results** calculated by the FID metric [39] suggests that, without the addition of stylised models, method 2 is the best method and method 3 is the least effective. However, since FID works better with larger sample sizes, **more accurate results would be developed with a larger dataset** of generated images by translating more books or using different seeds to produce the same book.

Overall, **method 1 is the best method**, since it was rated quite highly in both the qualitative and quantitative results, unlike the other two methods, which were the least effective for one of the two evaluation metrics.

# 6 Conclusion

In our initial Project Specification (Appendix archive), our aims were to generate images that have various styles of artwork to be more **understandable** and **sensitive** to different cultures around the world. These images were required to **keep the semantic information** supplied by the text, and preferably have a **similar composition** as the original image. It has been demonstrated that **the methods discussed in this report have obtained these goals** for the selected cultures: the UK, India, Japan and the Middle East. All three methods usually created images in the **style of artwork originating from that culture**, also changing the **characters' skin tone, hairstyle** and **attire** most common in that country. Additionally, many of the generated images **conveyed the most important information** described in the input text, working best with **simple, descriptive sentences**.

The Frechet Inception Distance used for quantitative evaluation also measured the **similarity between the generated and original images**, but this criterion was deemed **less important** than the other goals. This was because the composition of an image could differ from that of the original without any loss of semantic information.

Using a combination of qualitative and quantitative evaluation methods explained in Section 5, it was found that the most effective method for geo-localising a given book was **method 1** (Prompt Augmentation using Keywords). While all three methods were generally not as effective as a traditionally illustrated book, they are **much faster** to make and **less skill** is required. Therefore, this would **reduce the total cost for our client** and still result in an acceptable end product.

## 6.1 Challenges

Over the course of this project, **many challenges had to be overcome.** From the start, due to the novelty of the idea, its **feasibility had to be investigated**. This involved researching various image generation models and exploring their capabilities and possible drawbacks. An effective way to **evaluate** the project also had to be devised to appropriately assess its success. By using both **qualitative** (in the form of a questionnaire) and **quantitative** (using FID) methods of evaluation, a **more reliable examination** of our results were obtained from **unbiased, external sources**.

Once Stable Diffusion was selected as the basis of this project, an exploration was required to find the **optimal way to phrase the model's input** that causes the most desirable image to be generated. The original text from a book then needed to be **manipulated into this form** to increase the chances of an appropriate output. This also involved trying to find the **most important parts of a given sentence** or paragraph, which was implemented with the use of a POS tagger.

Many challenges were also faced because of the **group aspect** of this project. Further detailed in Section 8, this included **teamwork, communication, time management** and **organisation**. Time management was especially important due to the short timespan given.

# 7 Discussion

## 7.1 Future Advancements

Promising future advancements may be made to further increase image quality by incorporating Large Language Models and personalised stylistic models. The presented approach can also be adapted to Video and 3D sculpture generation for novel illustrations in digital webbooks.

### 7.1.1 Stylised Models

Stability AI, one of the creators of Stable Diffusion, is already **perusing localised and personalised models in emerging markets, such as India** by partnering with Local Companies[3]. The quality and accuracy of generated Illustrations may be further increased by leveraging this approach. A range of methodologies for stylising LDMs exist with varying degrees of customisation and computational requirements.

Approaches such as Textual Inversion[4] allow for the capture of singular novel concepts, given only a limited number of exemplar images. The concept is mapped to a new unseen token or word. Other recent methods for fine tuning Text-to-Image Diffusion Models, such as have Dream Booth [72] and Ever Dream[5] have enabled the synthesis of given subjects into novel renditions, backgrounds and contexts. Despite promising performance in subject specific generations these models have limited semantical context, and therefore would **not be appropriate for the generation of images in broad semantical areas, such as cultures**.

#### 7.1.1.1 Fine Tuning

Large Scale Fine Tuning[6] enables supervised training of LDMs with tailored User Defined Datasets, containing images labelled with textual descriptions. However, many iterations and extensive datasets are required to achieve suitable performance. Overfitting is also a very prevalent issue, and therefore counter-measures such as drop out layers should be implemented. All this requires large computational costs. A minimum of **24GB of VRAM** is needed to fine tune the model, however multiple GPUs totalling 30GB is recommended for faster training with larger batch sizes.

A Prominent example of Large Scale Fine Tuning is Japanese Stable Diffusion[7]. The model is based on Stable Diffusion version 1.4 and fine-tuned with over 100 million images with Japanese descriptions from a subset of the LAION-5B Dataset [43] were used to train the model. Due to stark differences between the English and Japanese, a specific Japanese tokenizer was necessary to be created. An example of generated images can be seen in Figure 44. A brief empirical observation of the generated images shown the performance of **Japanese**

---

[3] https://www.businessworld.in/article/Eros-Investments-Collaborates-With-Stability-Ai-For-Stable-Diffusion-/25-08-2022-443759
[4] https://huggingface.co/docs/diffusers/training/text_inversion
[5] https://github.com/victorchall/EveryDream-trainer
[6] https://huggingface.co/docs/diffusers/training/text2image
[7] https://huggingface.co/rinna/japanese-stable-diffusion

**SD using Japanese prompts to exhibit degraded performance when compared to SD version 1.5 used within Method 1.**

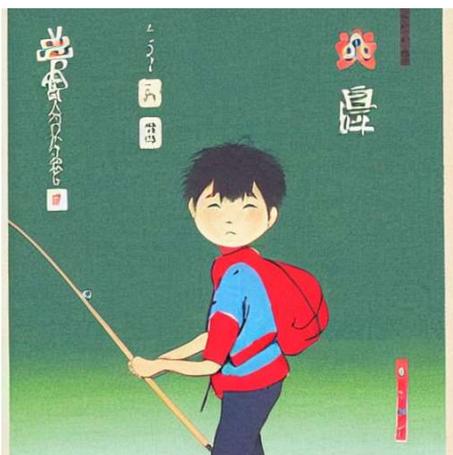 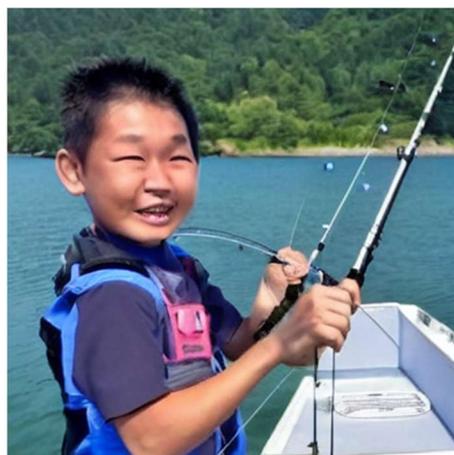

A) SD version 1.5 (method 1) with Prompt: "japanese children book style, As soon as he put out to sea in boat, dropped his line in water, big, fine-looking sea bream got caught."

B) Japanese SD with Prompt: "児童書風 舟で海に出て、釣り糸を水に落とした途端、大きくて立派な鯛が釣れた。"

Figure 44: Images Generated with SD and Japanese SD for Japanese interpretation of Not you Page 2

Other models such SD Pokémon Diffusers Model[8] were investigated. However, the limited variations in the dataset combined with smaller size meant images were highly homogeneous and lacked the ability to convey the appropriate semantics from the prompt. On the other hand, another model Nitro Diffusion[9], can synthesise images in multiple different style, including Disney Animations. **Given the varied contextual backgrounds in the training set, in addition to significant overlap between Disney productions and children books illustrations, enabled the generation of reliable and consistently styled imagery.**

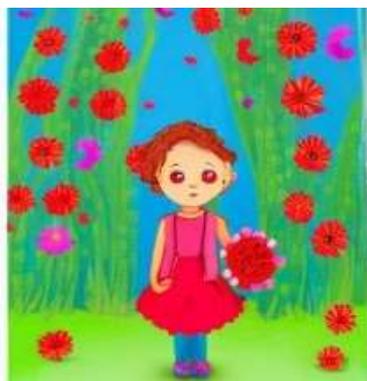 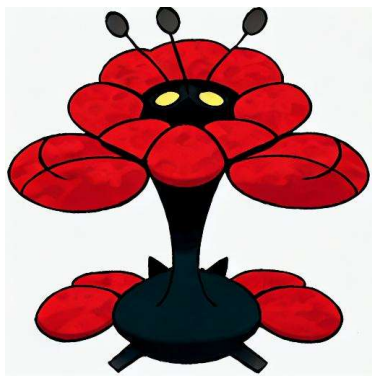 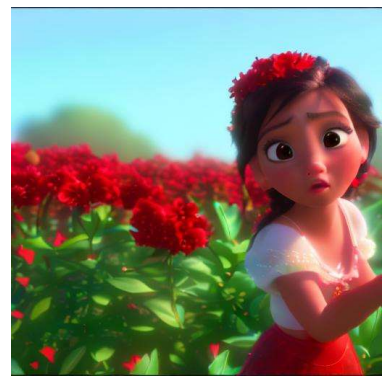

a) SD version 1.5 (Method 1) with Prompt: "childrens book style, She can see lot of red flowers."

b) SD Pokemon Diffusers[*4] with Prompt: "She can see a lot of red flowers"

c) Nitro Diffusion[*5] with Prompt: "disney style She can see lot of red flowers."

Figure 45: Images Generated with various SD models for In My Garden Page 9

Although most explored models gave the limited performance gained, Nitro Diffusion[9] generated comparable semantic information to Method 1, whilst retaining a consistent style. **A recreation of the Nitro Diffusion[*5] model for Childrens books could provide substantial** quality advances in future iterations, however large increases in budgetary demands would need to be made.

---

[8] https://huggingface.co/lambdalabs/sd-pokemon-diffusers
[9] https://huggingface.co/nitrosocke/Nitro-Diffusion

### 7.1.2 Large Language Model Assistance

**Current implementation of prompt generations is deprived of detail.** This a consequence of simple sentences featured in children's book which rely on visual imagery to engage the child instead. However, this Image-to-Image generation is not leveraged in our image translation, as it has been shown to degrade performance (Section 2.2.2). Incorporating the use of long-form non-fiction books, would expand the level of detail in textual descriptions. However, the use of such books would require text summarization to condense protracted whole page descriptions into a single prompt 77 tokens long. **By removing ambiguity in prompts, the quality and reliability of generated imagery can be greatly increased.**

Given this context, Large Language models [73] are the most prominent choice for the **augmentation of simplistic prompt from children's books, as well as the summarisation of long-form non-fiction books.** ChatGPT [74] is a State of Art Chat Bot, created by OpenAI. By incorporating a range of reinforcement learning techniques, a conversational dialogue may be formed, which is able to retain large amounts of contextual information. **This enables the model to leverage an overview a substantial amount of contextual semantical information from the whole book prior to generating a prompt for the SD model.**

| Original text: | He painted yellow stars on the front and back. |
|---|---|
| ChatGPT input: | There is a book called Riley Flies a Kite. It reads the following:<br>"""<br>He painted yellow stars on the front and back.<br>He tied on a brown string for the tail.<br>He added red bows.<br>The wind did not blow.<br>The kite would not fly.<br>Riley and his dad took the kite to the park.<br>The kite got stuck in a tree.<br>Where could Riley fly this kite?<br>He needed a place without trees, and he needed a breeze.<br>Riley had an idea.<br>He knew a great place to fly his kite.<br>"""<br>Write a Stable Diffusion prompt describing: He painted yellow stars on the front and back |
| ChatGPT output: | Riley showed off his creativity by carefully painting yellow stars on both the front and back of his kite, adding a touch of brightness and fun to his flying creation. |

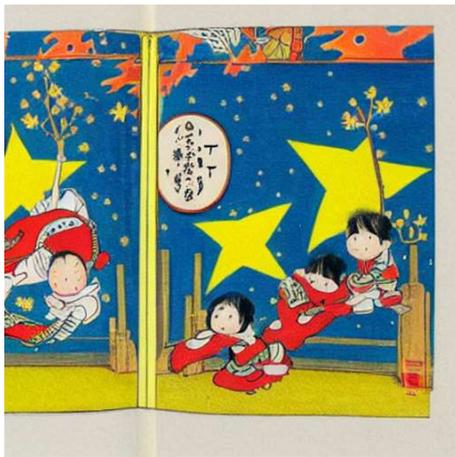
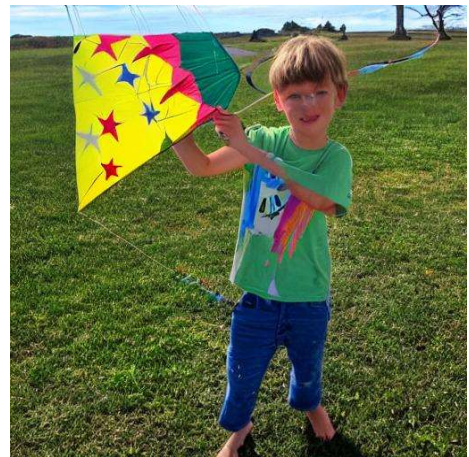

*a) Method 1 using Standard Prompt*  *b) Method 1 using ChatGPT Prompt*

*Figure 46: Images Generated with Method 1 using Standard and ChatGPT prompts of Riley Flies a Kite, Page 1*

Cohere Prompt Engineering[10] offers a similar approach online, however it's performance is on par with ChatGPT.

### 7.1.3  Expansion of Mediums for Digital Illustrations

The medium of web-based text media allows illustrations to leverage alternative mediums such as **Video and 3D sculptures.** Press websites have utilised these mediums, however, their use remains restricted within children's webbooks, due to the high production costs involved. Given current advances in LDM models being adapted to synthesis these new mediums, production costs will significantly reduce in the future and enable **more engaging illustrations to be formed**.

**Video Diffusion Models [75]** extend the Diffusion Model Architecture by concatenating U-Net layers into 3D U-Net layers, which process a fixed length series of frames, as opposed to a single image. Imagen Video is prominent model in this space, using a combination of 7 sub-models, which perform a variety of tasks such as: **text-conditional video generation, spatial super-resolution, and temporal super-resolution.** This promotes high quality and high-fidelity video generation. Other approaches include Phenaki [76], and state-of-art Make a Video[77], which can generate arbitrarily long video clips by conditioning on a **series of prompts as opposed to a singular prompt**. These scenes generated individually from each prompt are then transitioned between and coherently joined to form longer video content.

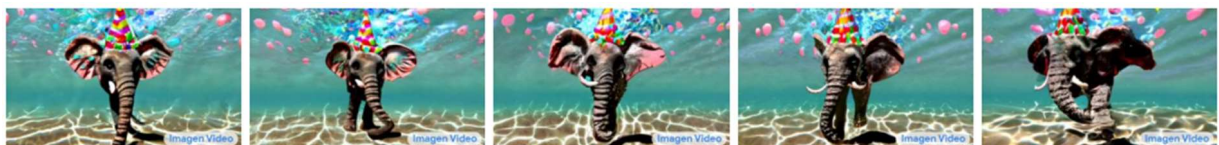

*Figure 47: Video Frames generated from a textual prompt using Imagen Video Model* [78]

Novel approaches, such as DreamFusion[79] and 3DiM[80] explore the synthesis of 3D model. **Neural Radiance Fields (NeRF)** enable the **representation of 3 dimensional objects within Deep Neural Networks** and their high dimensional abstractions.

---

[10] https://docs.cohere.ai/docs/prompt-engineering

The 3 Methods explored for the cultural alignment of generated images, could be adapted to the Video and 3D Diffusion models mentioned above in the future, in order to **create more engaging and interactive digital illustrations for children.**

## 7.2 Bias and Limitations

### 7.2.1 Limitations

Stable Diffusion version 1.5 was used in the majority of this project. However, in using this model came a few limitations. As seen in the pictures generated in Methods 1, 2 and 3, the AI occasionally will produce a photorealistic image, however, this is imperfect as in the case of human characters, some of their facial features are distorted, sometimes with the wrong number of appendages. Text, especially non-English text, will be rendered illegible, the latter facing issues due to western bias. Additionally, certain prompts may cause performance issues, such as ones that involve composition of different characters or shapes together. On the other hand, if the prompt is too short, and provides no helpful description, the model underperforms.

Context is important for some sentences in storybooks, as some parts in the story will reference previous pages or text, which could cause issues with the model, or some terminology used that might be unknown to it. This stems from the fact that the models Stable Diffusion, as well as CLIP, are trained only on data available on the internet, and hence, are not really familiar with traditional literature. Thus, future models could benefit by training with information related to different storybook style in different countries with their cultural themes, which all in all could create better generated images.

Looking at the software side, it was found that issues arose due to the autoencoding part of the model was lossy, causing images to not be exactly close to the input prompt. The model was trained on a large-scale dataset, LAION-5B, which contains adult material and thus, in the long term for this project, not fit for stand-alone product use without additional safety mechanisms and considerations [42]. Moreover, no additional measures were used to deduplicate the dataset, and as a result in the training data, some degree of memorization can be observed for images that are duplicated. It is possible that the training data can be searched at [81]and used to assist in the future detection of memorized images.

### 7.2.2 Bias

Aside from the limitations faced, the model was also subject to bias, in this case, social bias. As a result, the images generated can be unintentionally discriminatory, or stereotype cultures. In the images generated, aspects from western cultures are more favoured, as it is usually the default background that the model goes to [42]. Furthermore, non-English prompts have worse performance as evidenced in Section 3.3.1, this is disadvantageous to non-Western prompts. This is something that could be improved on for the future of Stable Diffusion, for it to be able to be used more widely in other countries.

## 7.3 Legal and Ethical Considerations

There are few risks this project may encounter, such as the end product having potential sources of product liability litigation, such as stereotyping / reduced diversification, and there may be potential for misuse/abuse for this product.

### 7.3.1 Possible Ethical Violations and Risks

As listed in the official website for Stable Diffusion, there are possible ways for this model to be misused. Creation of images that have cultural, religious or important environmental aspects that are wrongly represented can cause them to have harmful effects or become purposefully demeaning/dehumanizing. Promoting these images may also lead to or add on to discrimination and harmful stereotypes, causing disinformation [42]. Additionally, consent will also be an issue, as the model can create images with people that inadvertently impersonate others. This method could also be used with malicious intent, to create pornographic or egregiously violent images without consent of these same people. An upcoming issue that AI generated images are also facing is taking/learning illustrations from other artists, or even copyrighted images to generate artwork. These images have found to generated without consent, and in some cases, violate terms of use in copyright [82].

Furthermore, the software itself may pose certain ethical problems, such as loopholes in licensing. The Stable Diffusion model is an open-source code, and is licensed under OpenRAIL-M, which stands for Responsible AI Licensing for models. This prevents companies from allowing their models being used for malicious use. However, there is a clause that states that the licensor can update the model, and it is up to the user to ensure that they have the latest version [83]. This open possibility of acquisition of Stability AI, may degrade performance of any consequent versions, whilst and restricting usage of previous better performant models, and thus allow for it to be taken advantage of.

# 8 Project Management

## 8.1 Agile Project Management

Due to the nature of this project, this project needed to undergo Agile Project Management (APM). This is a method that allows the whole process to be flexible and dynamic, where the end product is subject to change, and not constrained by any definite goals in the initial management plan [84]. Compared to using a Waterfall Project Management plan, in which a series of phases are constructed, and each phase of the plan cannot proceed without approval, this plan is beneficial to software-based projects. Here, in a certain time period (also known as a sprint), a set amount of work is done, and after each sprint is completed, it will be revised, and feedback will be given. In this way, the feedback will determine what needs to be done next in the project, and where improvements and changes can be made. This overall continuous improvement prevents any large-scale failures and has a higher success rate for the project. However, there must be active and clear communication between all team members to ensure their efforts will go towards the same goal, as well as quick decision making [85].

A plan following an agile project management style can be broken down into 5 main phases, which are summarised in Figure 48.

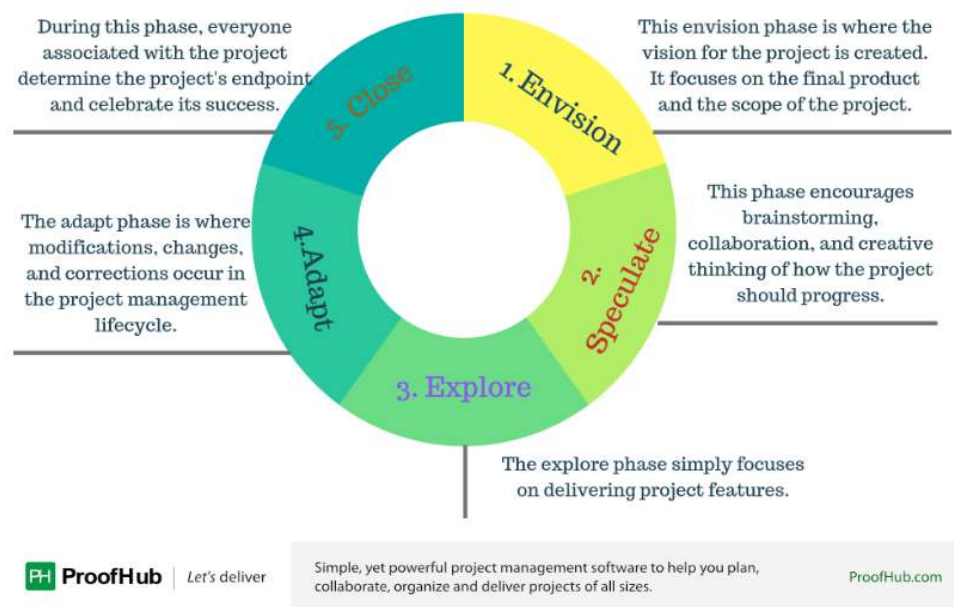

*Figure 48: 5 phases that happen when following an APM [86]*

To see our goals in the long term, a Gantt chart was used, and updated after the seminar in November. It is included in Appendix F. However, as this project utilises APM, a software, Jira [87] was used instead, as it is more interactive and adaptable than a Gantt Chart. It is a work management tool which provides various ways to organise agile management, and in the case of this project, the Kanban boards, backlogs and roadmaps were used [88]. Kanban boards are made up of 4 columns: to do, in progress, in review and done, with the tasks passing through these categories only once they have been completed. Kanban boards allow a transparency for the members, allowing them to view what work must be done, and allocate it to themselves [89]. A roadmap visualises the goals along a set timeline, which can include any deliverables, deadlines, or objectives [90]. In Jira, the dates can be easily set, making it flexible and convenient to use. The October 2022 to January 2023 roadmap created for this project can be found in Appendix G. A backlog is a compilation of the task into a list, and each task is done in each sprint. This backlog depends on the structure of the roadmap, and thus, tasks can be removed or added. Overall, these subsets of tools helped to allocate tasks easily, making it easier to view long-term goals and keeping on track with deadlines.

## 8.2 Meetings

To ensure everyone on the team had an idea of what was going on, and the tasks that needed to be done to complete the project, it was decided in the first two weeks that biweekly meetings would be held, and that all should be in attendance. The meetings were held on Monday at 2:30pm, and with the supervisor on Fridays at 4:30pm. This was organised by creating a calendar on Outlook, and initially the meetings were held online via a created group in Microsoft Teams, but towards the end of the semester, face to face meetings were held. For any further discussions, WhatsApp was used, as a simple way to easily contact different members, while ensuring everyone is on the same page.

For each of the deadlines, such as the October and November seminar, meetings were held more regularly, and the presentations were rehearsed prior to the seminar. In each meeting, minutes were taken, and the next set of tasks to undertake between meetings were determined, as well as any accomplishments that had been made. Any problems or obstacles were also addressed during this time. In this way, it was ensured that progress could be steadily made. The minutes and other files, such as files for submissions and research notes were stored in a shared Microsoft Sharepoint, and any programming related files were stored on GitHub. Examples of the minutes can be found in the Appendix archive.

## 8.3 Division of Responsibilities and Authorship

The team was made up of 4 members, 2 of which were working towards a degree in Computer Science, and the other half, Electrical and Electronic engineering. Since this project leaned towards AI and was something that the latter was not very familiar with, the non-technical tasks were delegated based on other strengths, such as management and evaluation skills. It was also ensured that each member was continuously busy, either aiding or working independently on a part of the project.

The independent roles and responsibilities undertaken by each member are: Project Manager [91], Technical Manager [92], Software Developer[93] and Evaluator[94]. Table 5 summarises the responsibilities that each member possesses, for each role that they take. This was done in the beginning of the project, with the help of the supervisor and opinions and consent of all team members.

|  | Responsibilities |
| --- | --- |
| Project Manager | <ul><li>Monitors and manages the overall project and progress</li><li>Keeps track of deadlines and any deliverables</li><li>Delegates tasks</li><li>Plans out the next step of action</li></ul> |

| | | |
|---|---|---|
| | | • Manages any project risks that are prevalent, and has contingency plans<br>• Interfaces with the supervisor to ensure that the project is going in the correct direction |
| Technical Manager | | • Usually has a good level of expertise and knowledge on the project topic.<br>• Oversee all technical matters<br>• Troubleshoot problems, and do quality assurance<br>• Able to take up managerial roles |
| Software Developer | | • Have good knowledge of different programming languages<br>• Write clean, concise code, according to the specifications needed<br>• Create effective algorithms<br>• Have good problem-solving skills<br>• Troubleshoot and verify existing programs/softwares |
| Evaluation Manager | | • Routinely analyse and review any data collected, and synthesise any evaluation findings<br>• Improve and oversee any databases used<br>• Implement any evaluation processes, and specify what is needed in the evaluation part of the report<br>• Consult and contact appropriate people during the revision process |

Table 5: Summary of the responsibilities for the roles undertaken by each member of the project.

The report was sectioned off and written based on the knowledgeability and the overall ability of each member in the group. The authorship is summarised in Table 6 below.

| Group Members | Roles | Written Section | Word count |
|---|---|---|---|
| Aneesha Amodini | Project Manager | Abstract[co],1,4.3[co],5.2, 7.2,7.3, 8 | 4,910 |
| Victor Suarez | Technical Manager | 2.1, 3.2, 3.3[co], 9 | 4,932 |
| Jakub Dylag | Technical Manager & Software Developer & Researcher | Abstract[co], 2.2, 3.3[co], 3.4, 4.3[co], 5.3, 7.1, Appendix H | 5,313 |
| James Wald | Evaluation Manager | 4, 4.2, 5, 5.1, 5.2.1, 5.4, 6, Appendices A-D | 3,934 |

Table 6: Summary of word count and written section done by each group member for this project.
co = co-written by both members

## 8.4 Setbacks

There was a setback faced in the qualitative evaluation section of the report. Due to miscommunication issues, the questionnaire took longer to complete than expected. Additionally, it was realised that due to the nature of the questionnaire, an ethics form had to be submitted as sensitive data such as ethnicity, age, and gender would be collected. Awaiting approval led to a delay in the distribution of the questionnaire. Since the questionnaire was approved at the beginning of the Christmas break, the committee members of the societies became harder to contact. As a result, only the Japanese questionnaire had any responses before the deadline. However, a contingency plan was created, using quantitative evaluation instead, relying on calculations instead, yielding the desired results.

# Appendices

# Appendix A: UK Generated Images
## In My Garden (Original images from [20])

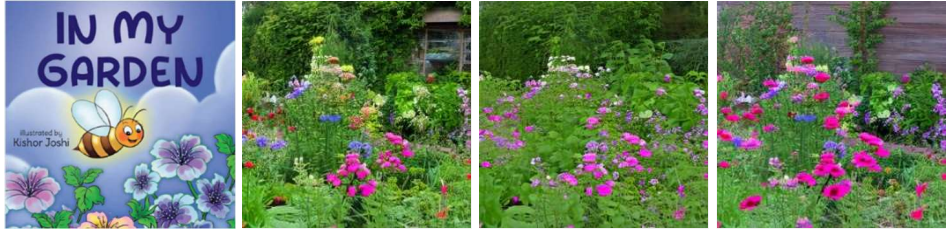

Page 1: "I see a lot of butterflies."

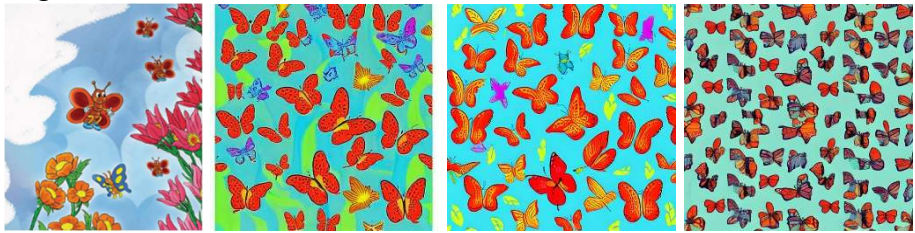

    Original        Method 1        Method 2        Method 3

Page 2: "I see a lot of bees."

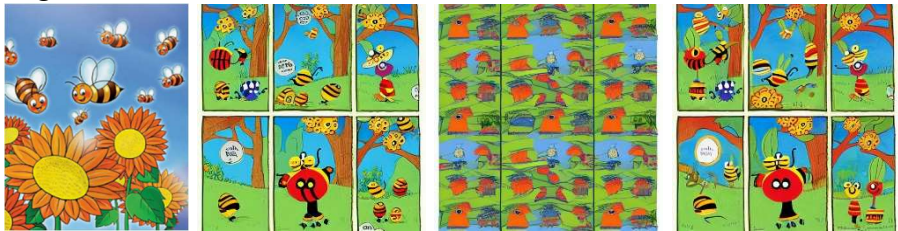

    Original        Method 1        Method 2        Method 3

Page 3: "I see big bees in my garden."

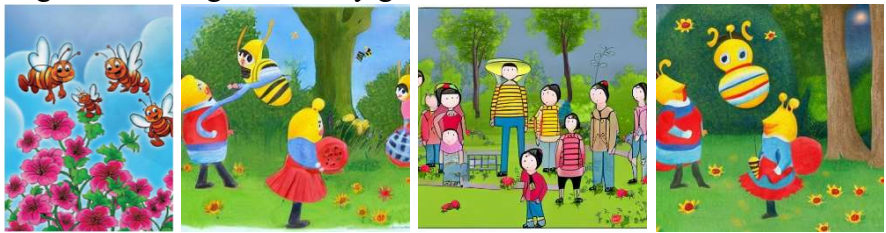

    Original        Method 1        Method 2        Method 3

Page 4: "I can see birds."

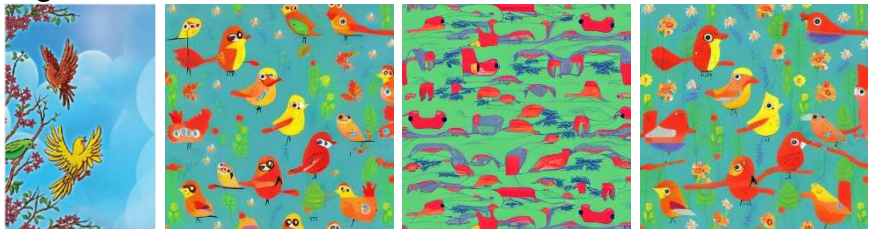

    Original        Method 1        Method 2        Method 3

Page 5:"I see big flowers. I see small flowers."

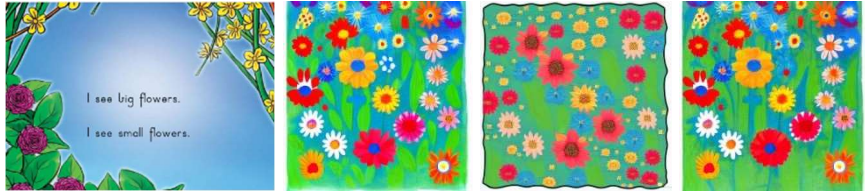

Original      Method 1      Method 2      Method 3

Page 6:"I can see red flowers in my garden."

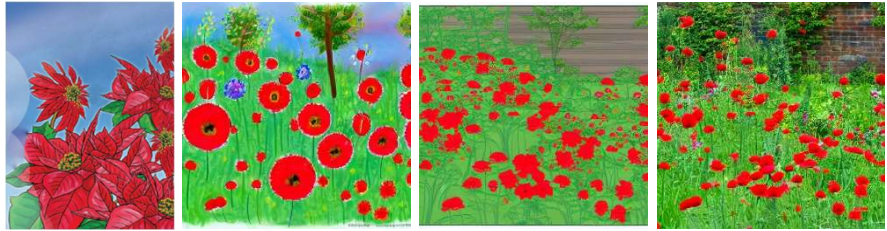

Original      Method 1      Method 2      Method 3

Page 7:"I can see white flowers."

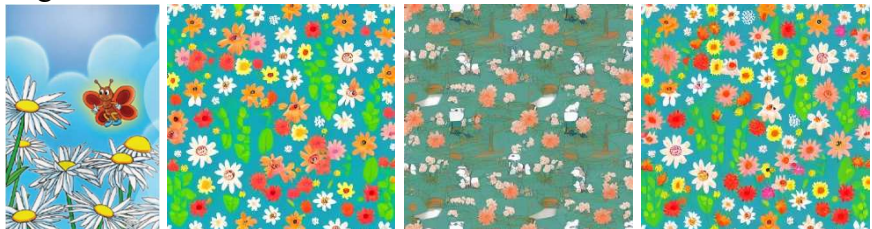

Original      Method 1      Method 2      Method 3

Page 8:"A bee can see the flowers."

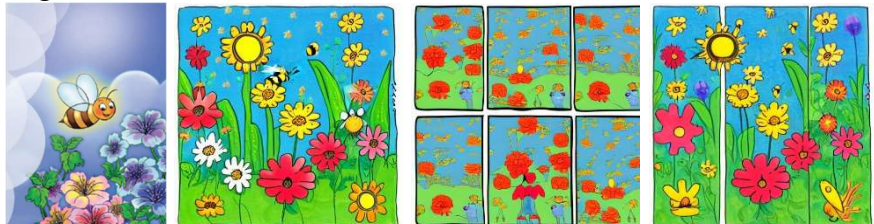

Original      Method 1      Method 2      Method 3

Page 9:"She can see a lot of red flowers."

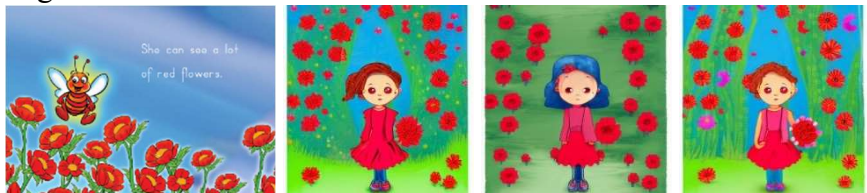

Original      Method 1      Method 2      Method 3

Page 10: "I love my garden."

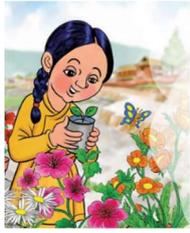 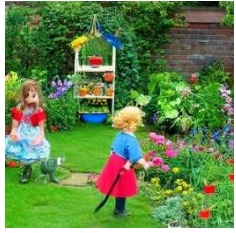 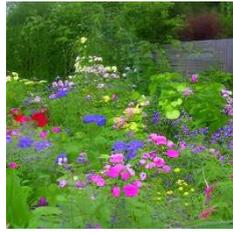 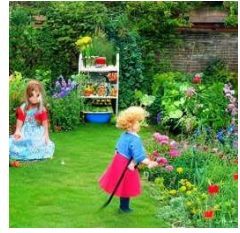

Original    Method 1    Method 2    Method 3

# Appendix B: India Generated Images
## The Mango Tree (Original images from [32])

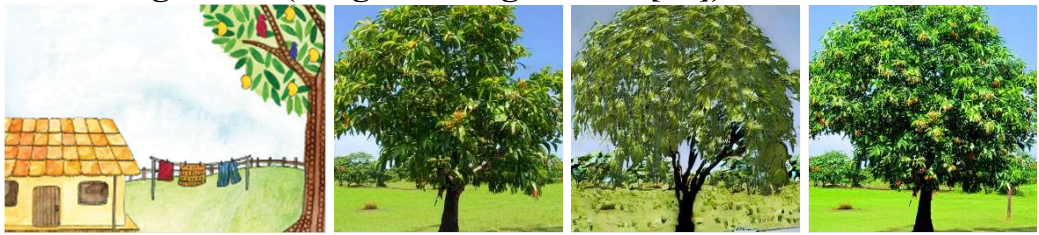

Page 1: We like to visit Grandma's village

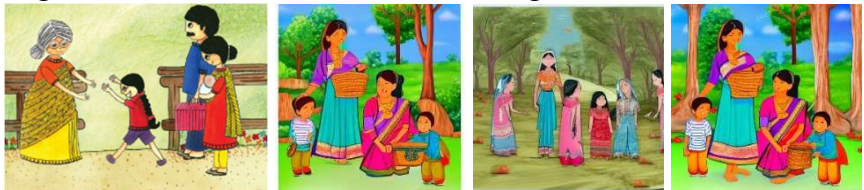

    Original        Method 1        Method 2        Method 3

Page 2: Grandma has a big mango tree in her garden. Many birds come there.

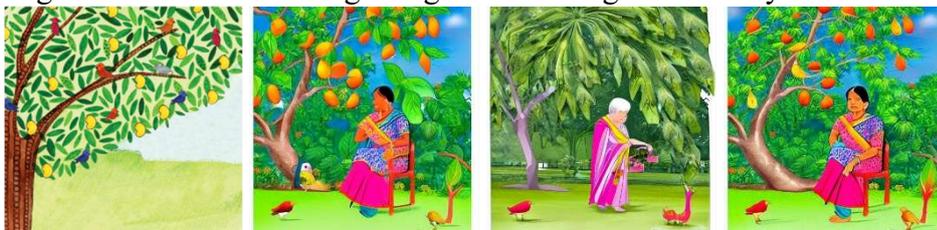

    Original        Method 1        Method 2        Method 3

Page 3: I like to hide in the mango tree. No one can see me here.

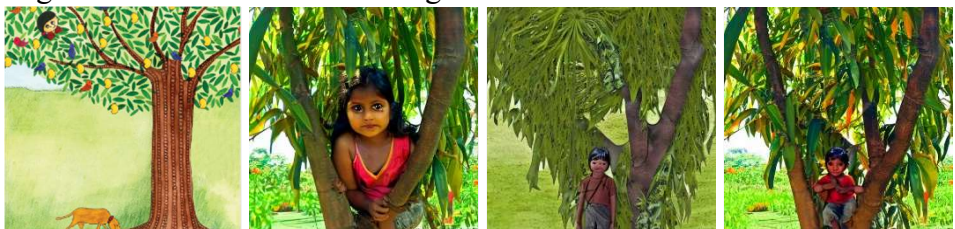

    Original        Method 1        Method 2        Method 3

Page 4: Father looks for me near the well. Mother looks for me near the cowshed. But no one can find me.

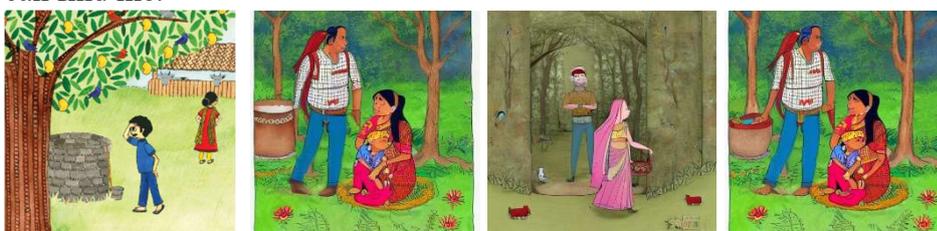

    Original        Method 1        Method 2        Method 3

Page 5: Oh no! At last, Grandma's dog finds me.

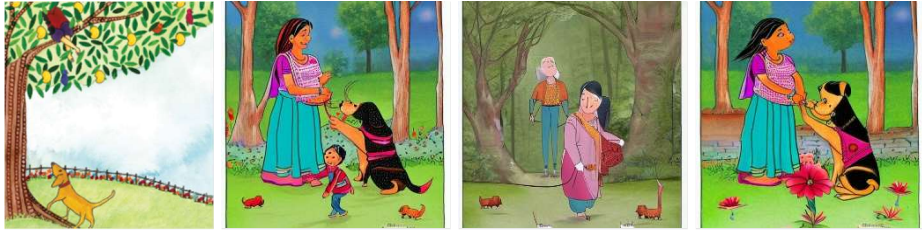

Original　　　　　Method 1　　　　　Method 2　　　　　Method 3

Page 6: Now I have to climb down. Grandma gives me a big hug.

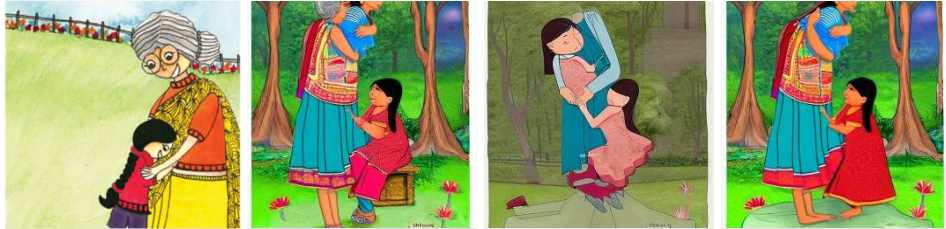

Original　　　　　Method 1　　　　　Method 2　　　　　Method 3

## Riley Flies a Kite (Original images from [51])

Page 1: Riley made a paper kite.

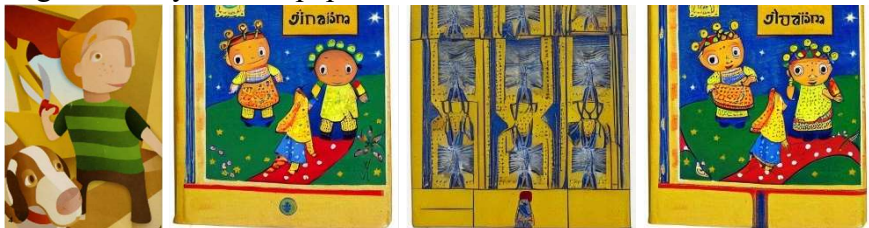

Original　　　　　Method 1　　　　　Method 2　　　　　Method 3

Page 2: He painted yellow stars on the front and back.

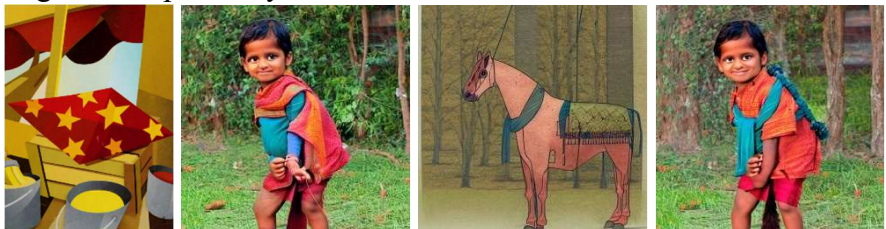

Original　　　　　Method 1　　　　　Method 2　　　　　Method 3

Page 3: He tied on a brown string for the tail. He added red bows.

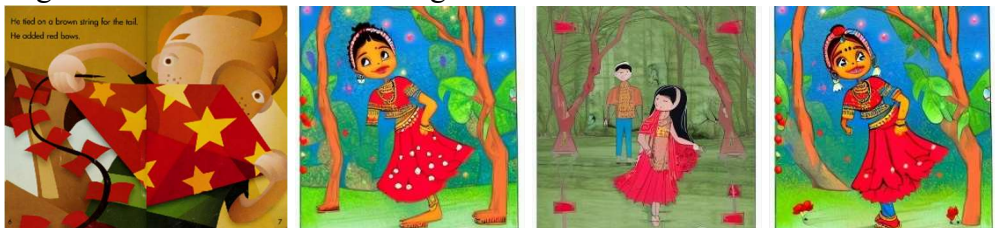

Original　　　　　Method 1　　　　　Method 2　　　　　Method 3

Page 4: Riley and his dad went out to the backyard.

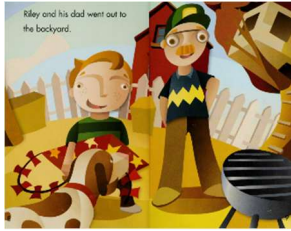 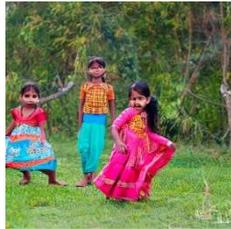 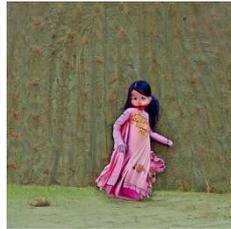 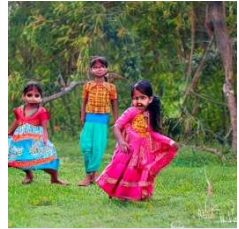

    Original            Method 1            Method 2            Method 3

Page 5: The wind did not blow. The kite would not fly.

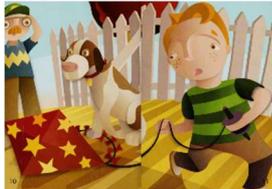 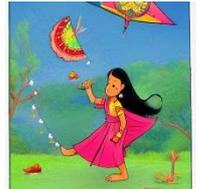 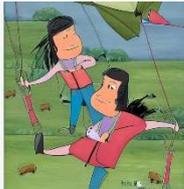 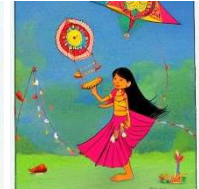

    Original            Method 1            Method 2            Method 3

Page 6: Riley and his dad took the kite to the park.

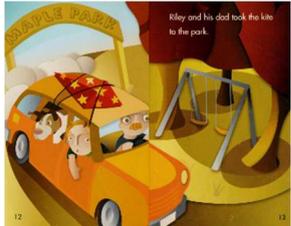 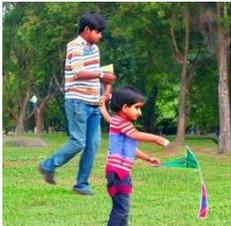 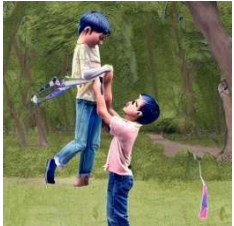 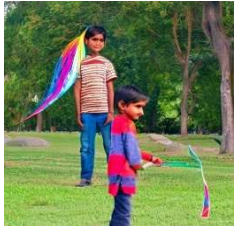

    Original            Method 1            Method 2            Method 3

Page 7: The kite got stuck in a tree.

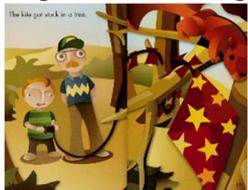 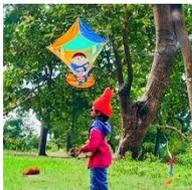 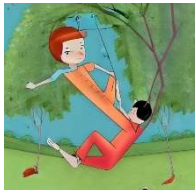 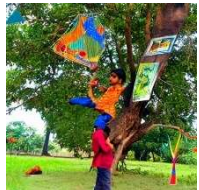

    Original            Method 1            Method 2            Method 3

Page 8: Where could Riley fly this kite?

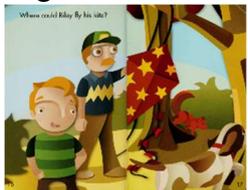 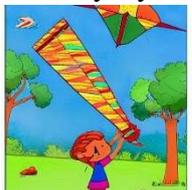 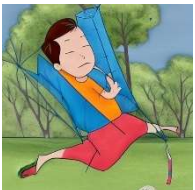 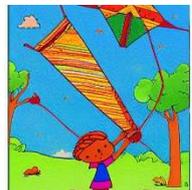

    Original            Method 1            Method 2            Method 3

Page 9: He needed a place without trees, and he needed a breeze.

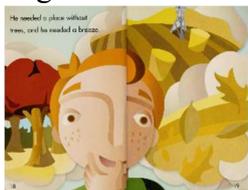 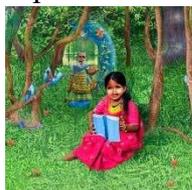 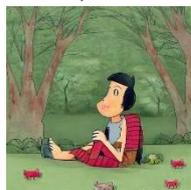 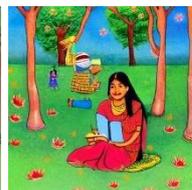

    Original            Method 1            Method 2            Method 3

Page 10: Riley had an idea. He knew a great place to fly his kite.

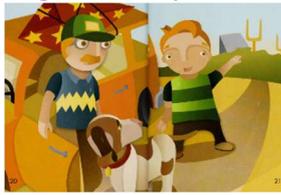 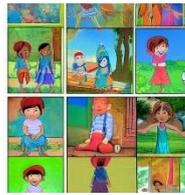 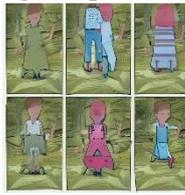 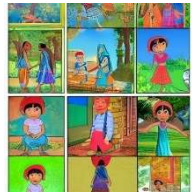

  Original           Method 1           Method 2           Method 3

Page 11: The football field at the school was the perfect place to fly his kite.

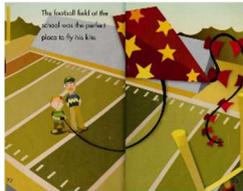 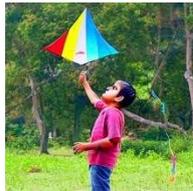 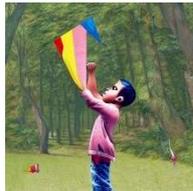 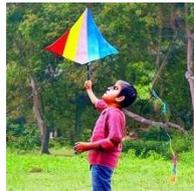

  Original           Method 1           Method 2           Method 3

## Appendix C: Japan Generated Images
### Not You! (Original images from [23])

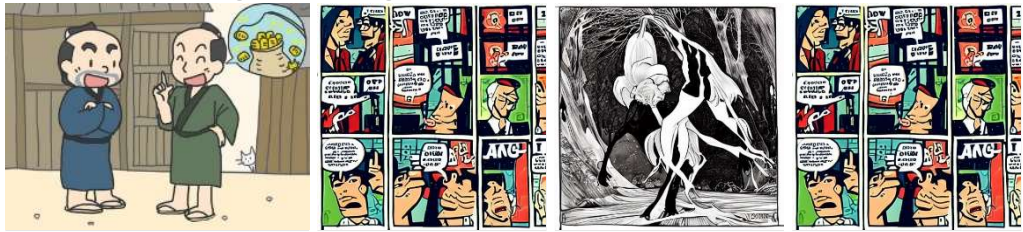

Page 1:"Hachigoro heard that someone had caught 50 gold coins while fishing."

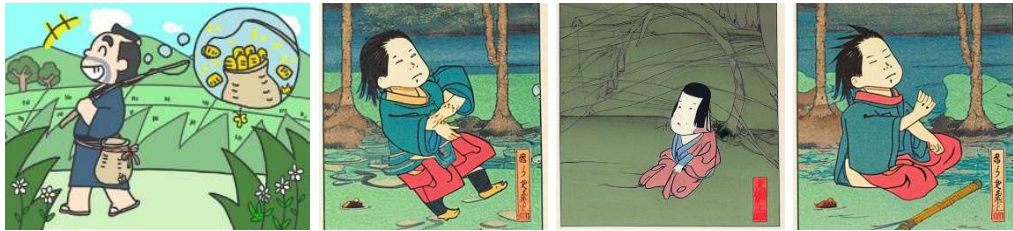

    Original            Method 1           Method 2           Method 3

Page 2:"So he left home right away, carrying a fishing pole on his shoulder."

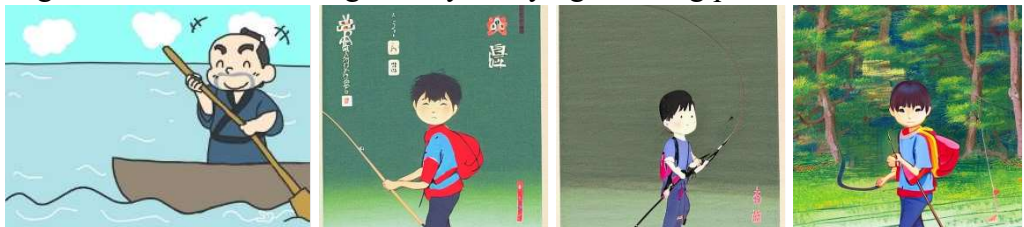

    Original            Method 1           Method 2           Method 3

Page 3:""

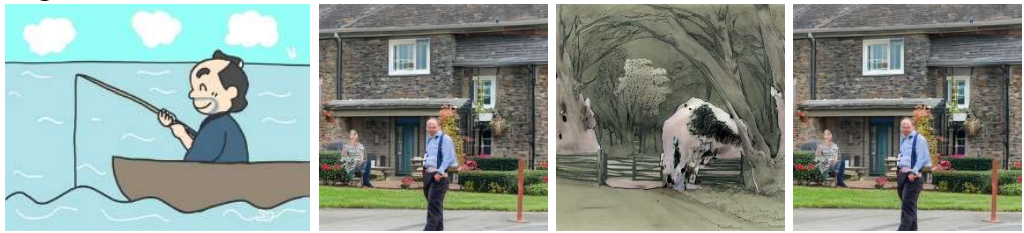

   Picture #1          Picture #2          Picture #3          Picture #4

Page 4:"As soon as he put out to sea in a boat and dropped his line in the water, a big and fine-looking sea bream got caught."

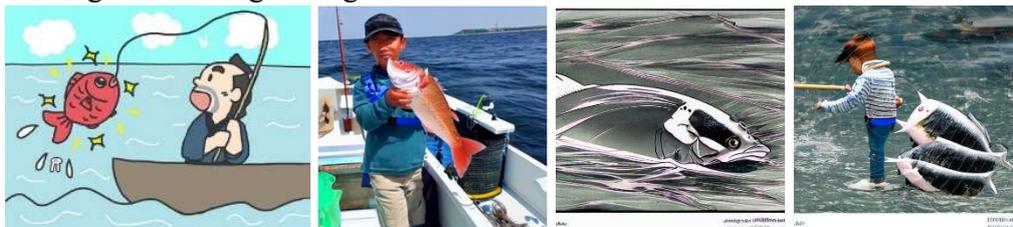

    Original            Method 1           Method 2           Method 3

Page 5:"Hachigoro, however, let the sea bream off the hook and threw it back to the sea saying, 'Darn it. Not you'"

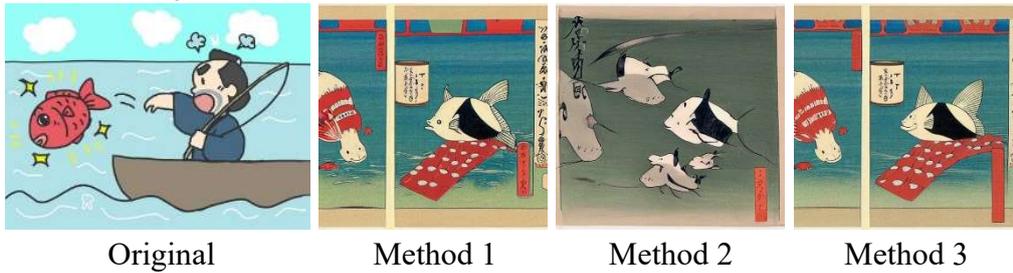

Original      Method 1      Method 2      Method 3

## Riley Flies a Kite (Original images from [51])

Page 1: Riley made a paper kite. He painted yellow stars on the front and back.

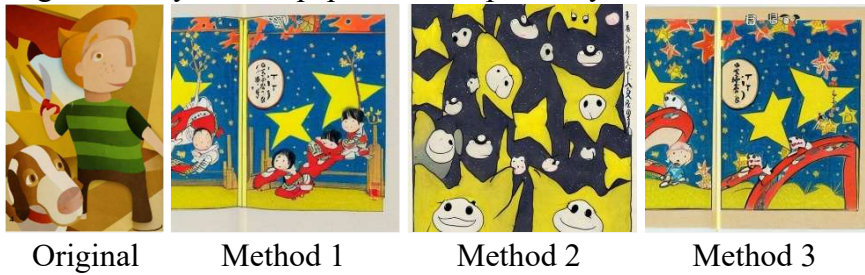

Original      Method 1      Method 2      Method 3

Page 2: He tied on a brown string for the tail.

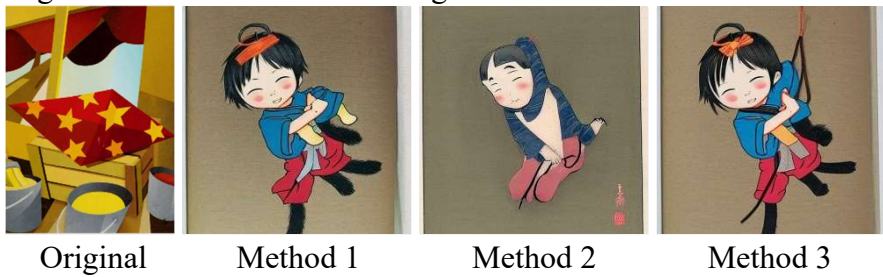

Original      Method 1      Method 2      Method 3

Page 3: He added red bows.

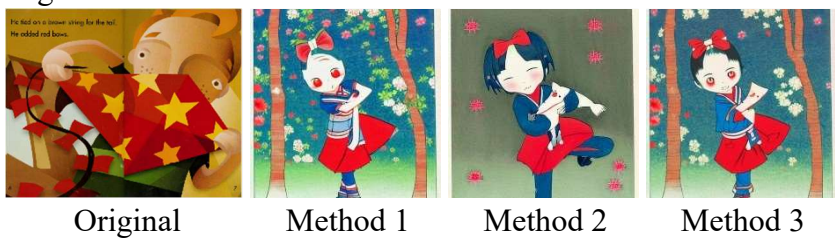

Original      Method 1      Method 2      Method 3

Page 4: Riley and his dad went out to the backyard. The wind did not blow.

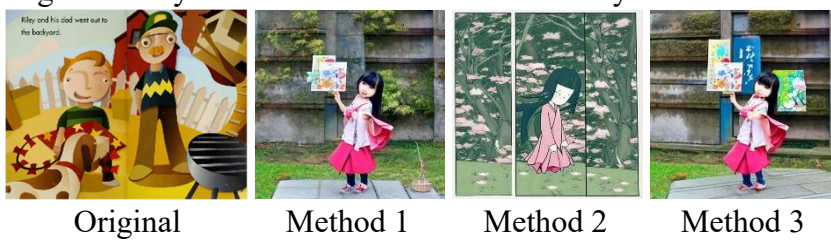

Original      Method 1      Method 2      Method 3

Page 5:   The kite would not fly.

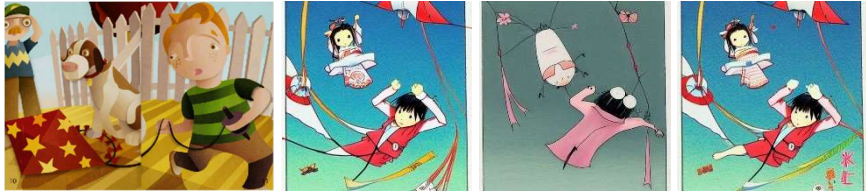

      Original          Method 1        Method 2        Method 3

Page 6:   Riley and his dad took the kite to the park.

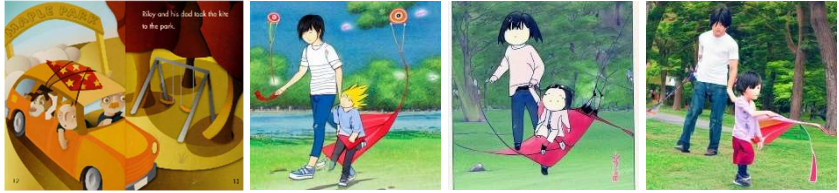

      Original          Method 1         Method 2        Method 3

Page 7:   The kite got stuck in a tree.

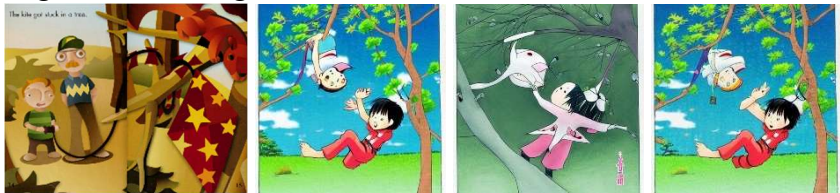

      Original          Method 1         Method 2        Method 3

Page 8:   Where could Riley fly this kite?

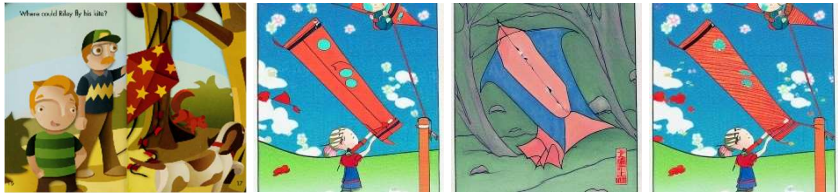

      Original          Method 1         Method 2        Method 3

Page 9:   He needed a place without trees, and he needed a breeze.

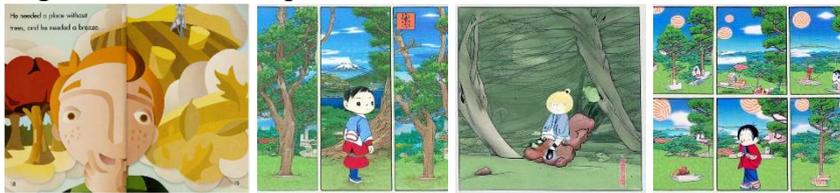

      Original          Method 1         Method 2        Method 3

Page 10: Riley had an idea.

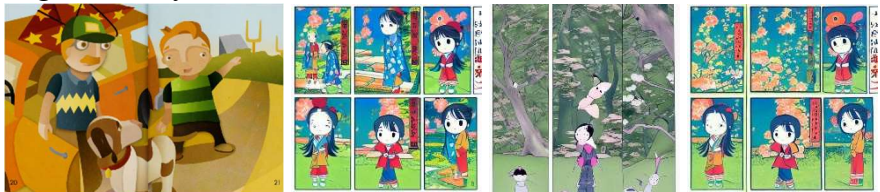

      Original          Method 1         Method 2        Method 3

Page 11: He knew a great place to fly his kite. The football field at the school was the perfect place to fly his kite.

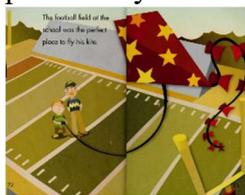 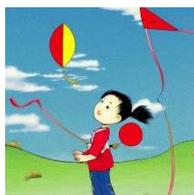 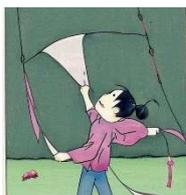 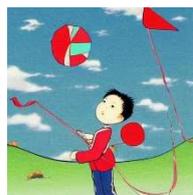

Original    Method 1    Method 2    Method 3

# Appendix D: Middle East Generated Images

## Riley Flies a Kite (Original images from [51])

Page 1: Riley made a paper kite. He painted yellow stars on the front and back.

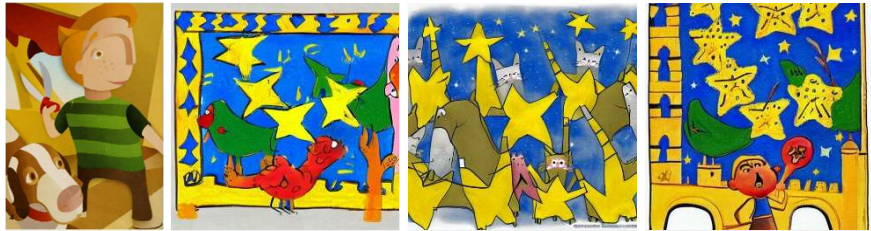

    Original        Method 1        Method 2        Method 3

Page 2: He tied on a brown string for the tail.

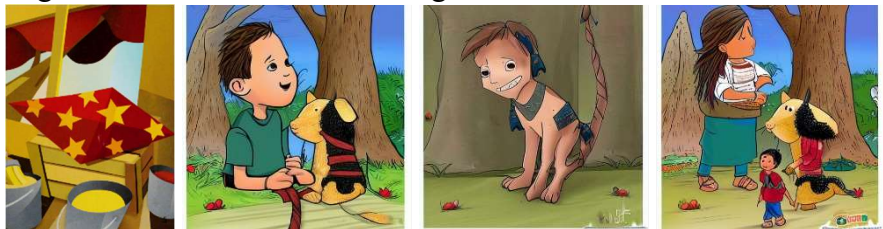

    Original        Method 1        Method 2        Method 3

Page 3: He added red bows.

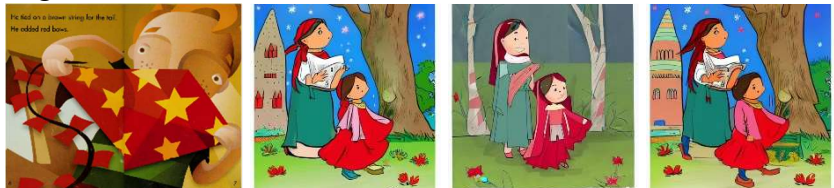

    Original        Method 1        Method 2        Method 3

Page 4: Riley and his dad went out to the backyard. The wind did not blow.

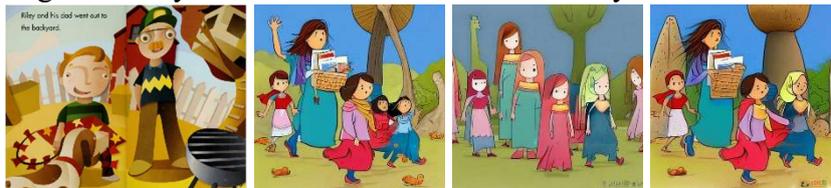

    Original        Method 1        Method 2        Method 3

Page 5: The kite would not fly.

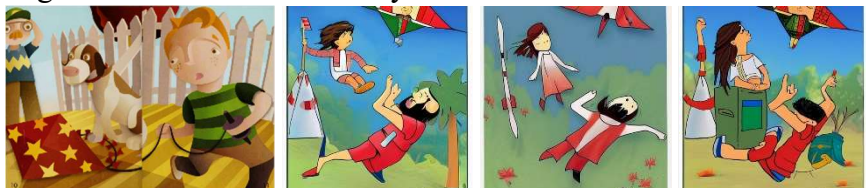

    Original        Method 1        Method 2        Method 3

Page 6: Riley and his dad took the kite to the park.

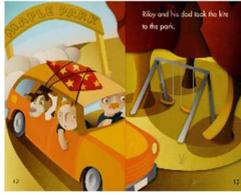 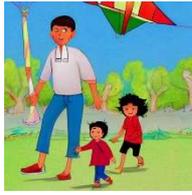 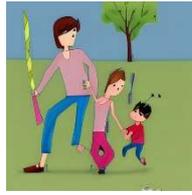 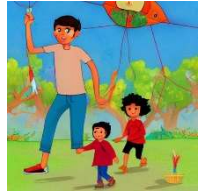

    Original        Method 1        Method 2        Method 3

Page 7: The kite got stuck in a tree.

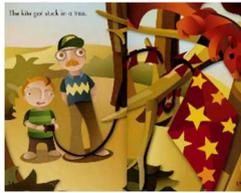 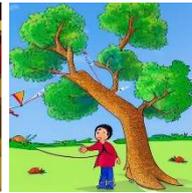 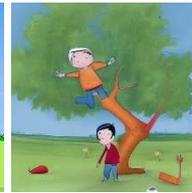 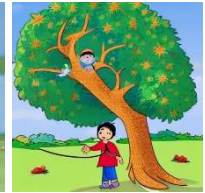

    Original        Method 1        Method 2        Method 3

Page 8: Where could Riley fly this kite?

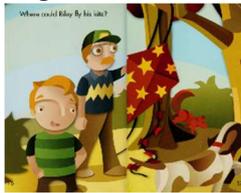 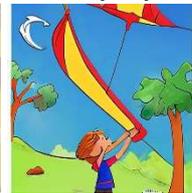 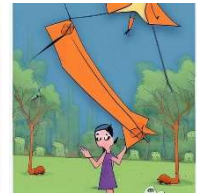 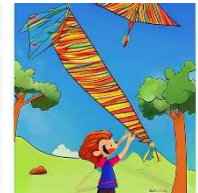

    Original        Method 1        Method 2        Method 3

Page 9: He needed a place without trees, and he needed a breeze.

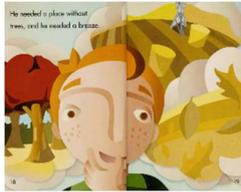 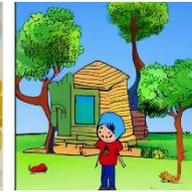 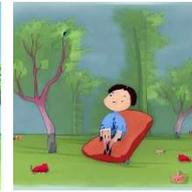 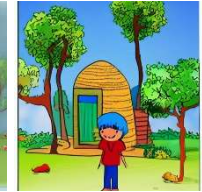

    Original        Method 1        Method 2        Method 3

Page 10: Riley had an idea.

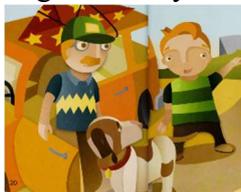 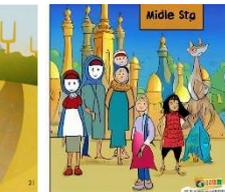 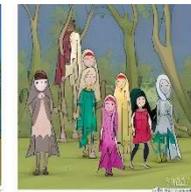 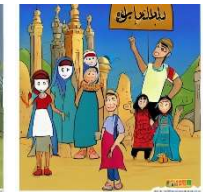

    Original        Method 1        Method 2        Method 3

Page 11: He knew a great place to fly his kite. The football field at the school was the perfect place to fly his kite.

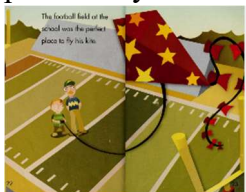 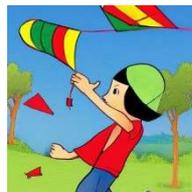 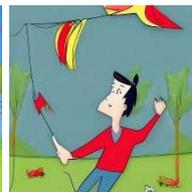 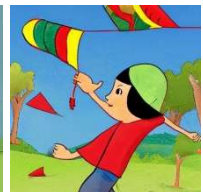

    Original        Method 1        Method 2        Method 3

# Appendix E: Full list of Books in Dataset

| English  | Riley Flies a Kite [51]<br>A Dog on a Log [95]<br>I Can Paint! [96] |
|----------|---------------------------------------------------------------------|
| Indian   | A Street, or a Zoo [97]<br>The Kitten [98]<br>The Mango Tree [32]   |
| Japanese | Not You! [23]<br>The Snow Woman [99]                                |

# Appendix F: Gantt Chart

Figures 49, 50 and 51 show the original Gantt Chart (made in October), the revised Gantt Chart (made in November) and Final Gantt Chart (made in January).

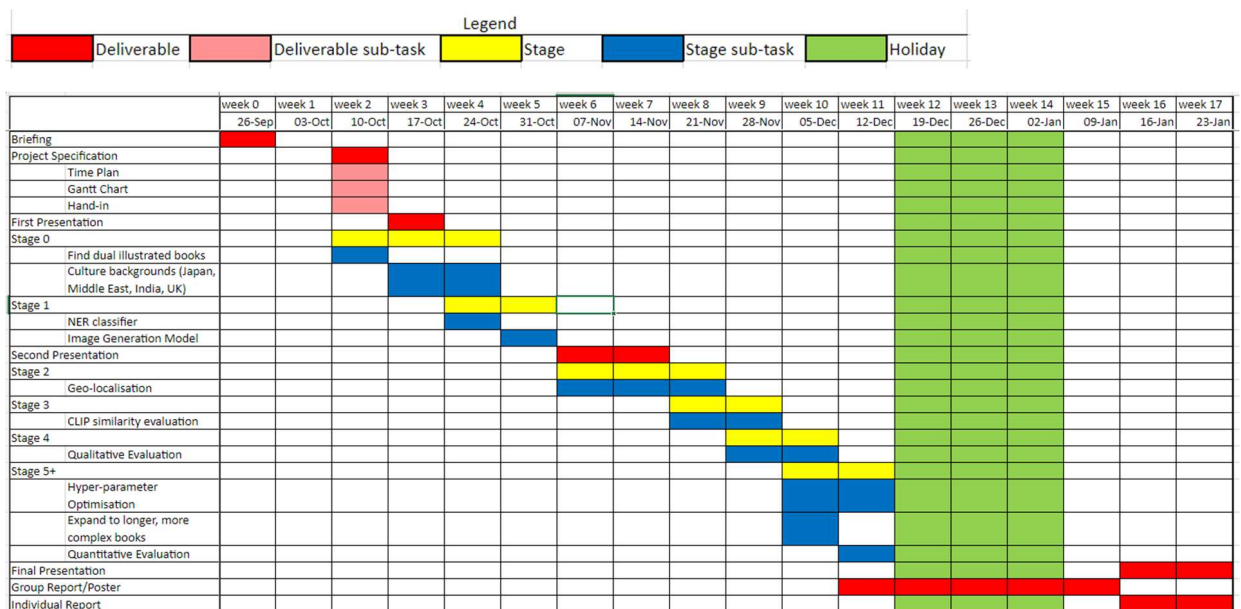

*Figure 49: Initial Gantt Chart*

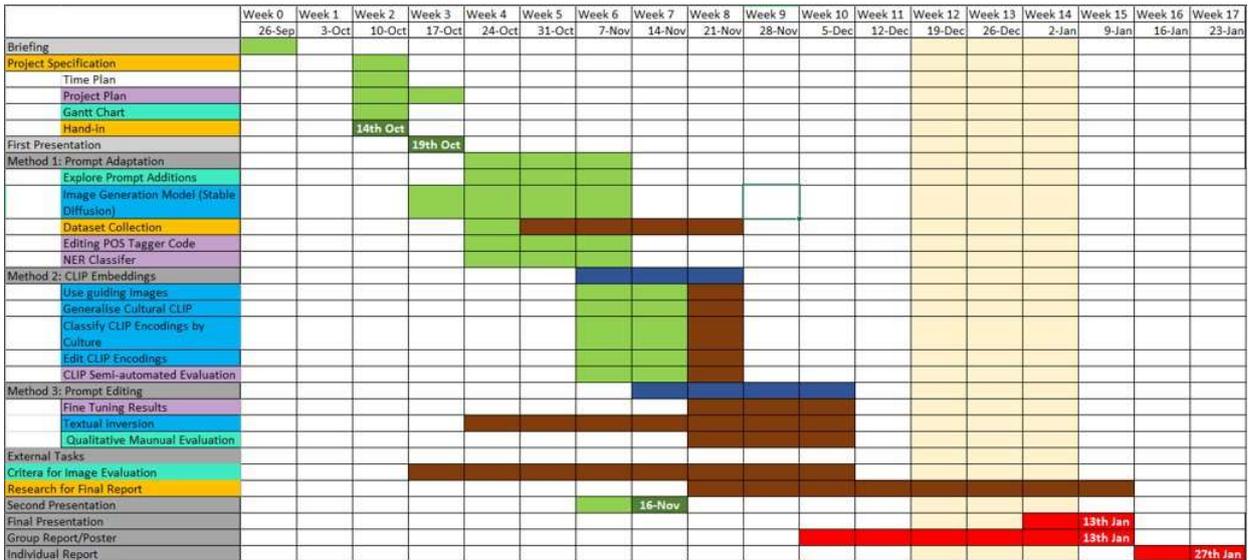
*Figure 50: Revised Gantt Chart, with highlighted member tasks*

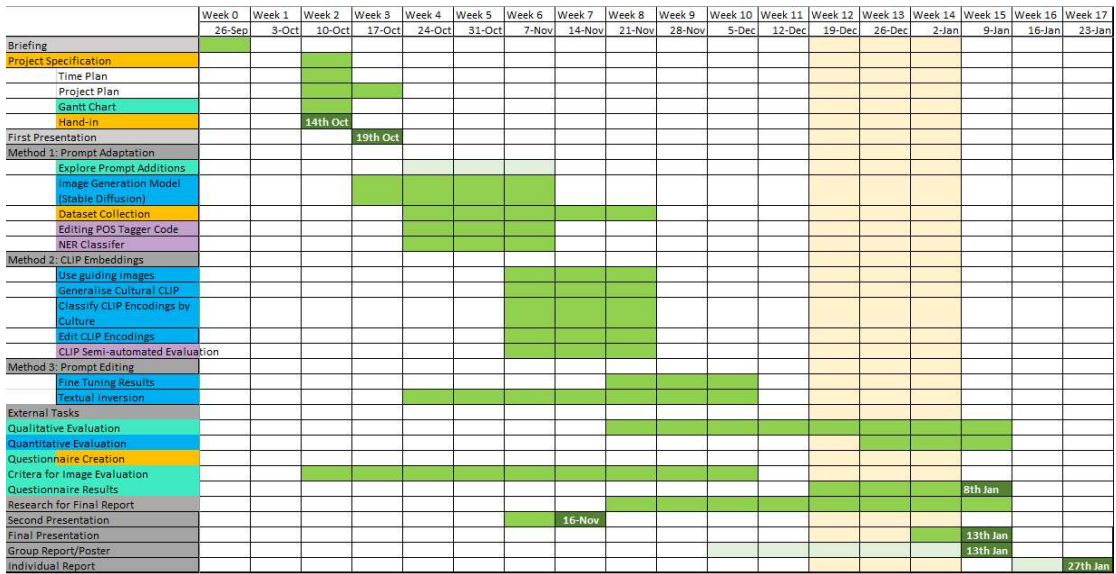
*Figure 51: Final Gantt Chart, with highlighted member tasks*

# Appendix G: Jira Roadmap

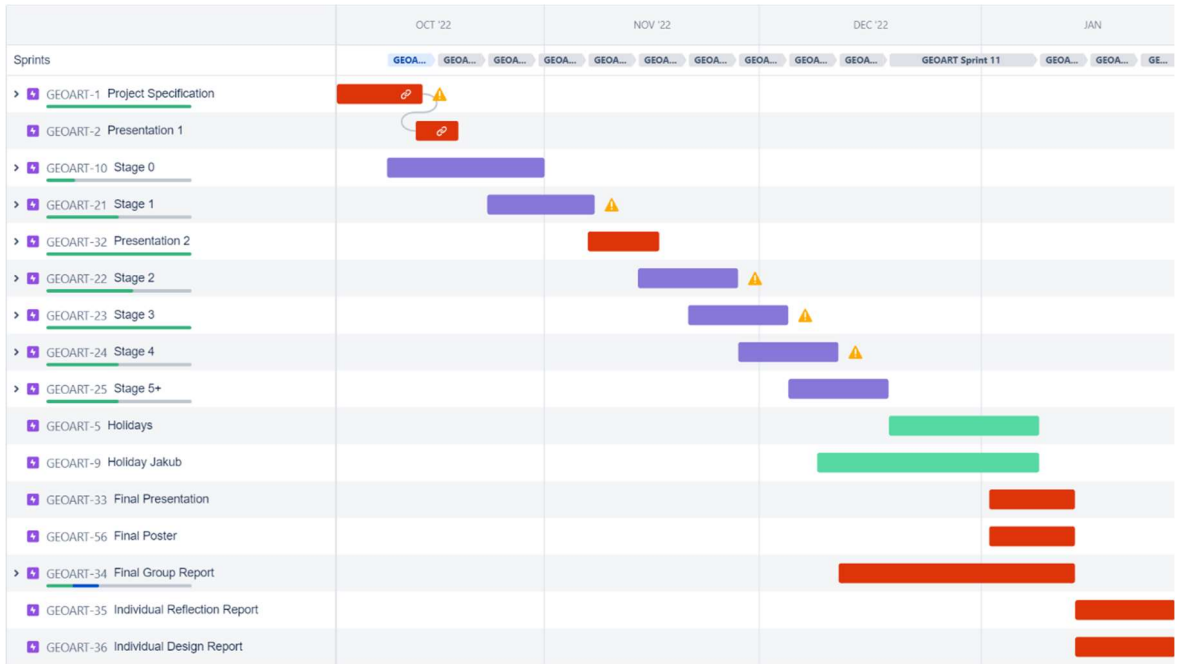

Figure 52: Jira Roadmap for Project – October to January

# Appendix H: Stylised Models Generated Images

## Japanese Stable Diffusion

Images Generated using the Japanese Stable Diffusion Model for the book Not You [23]

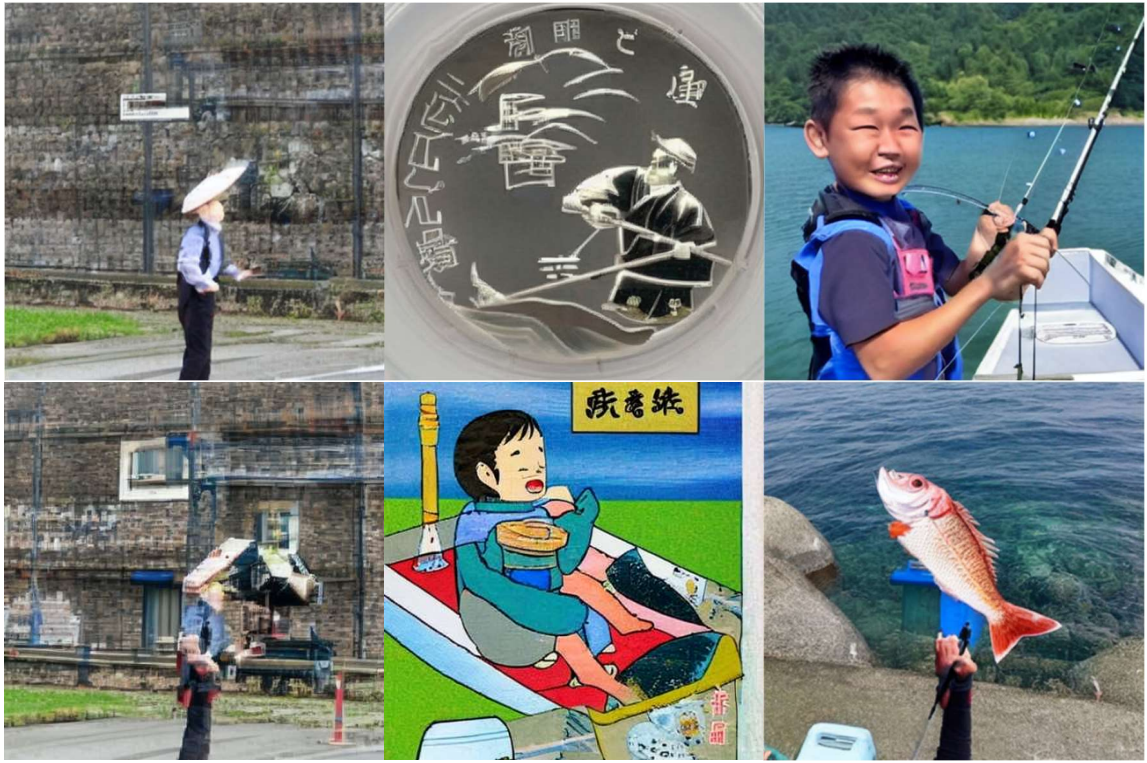

## Pokémon Diffusers

Images Generated using the SD Pokemon Diffusers Model for the book In My Garden [20]

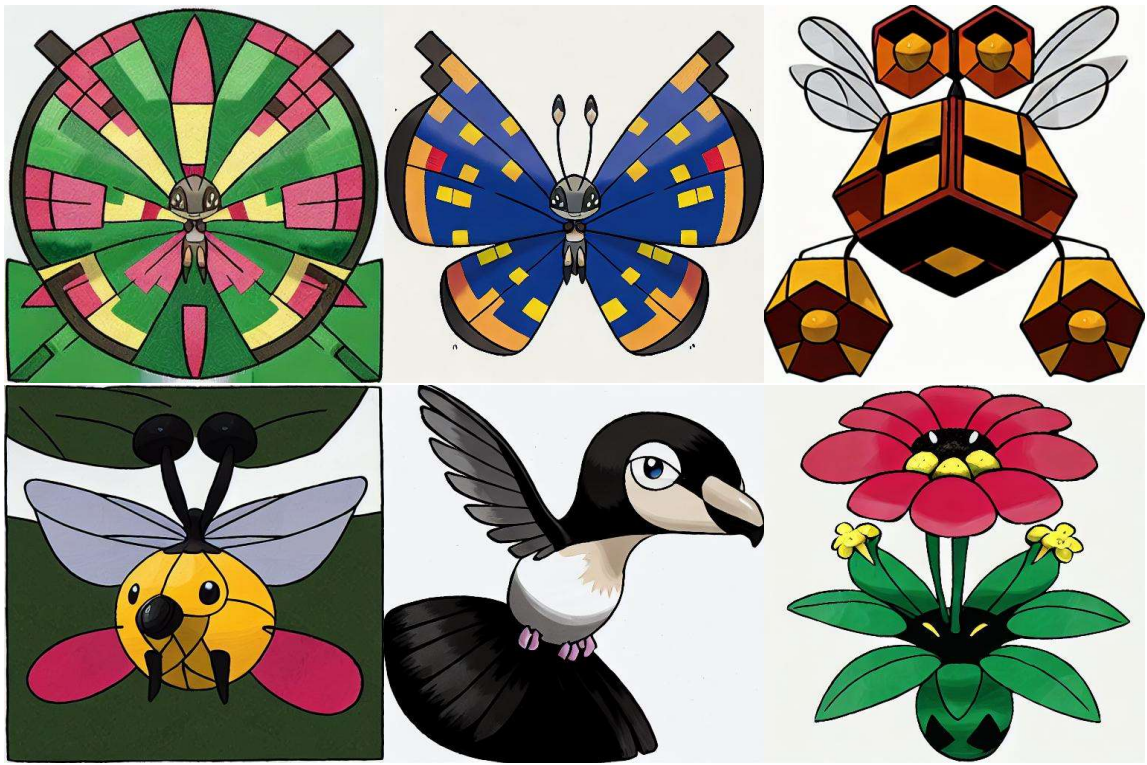

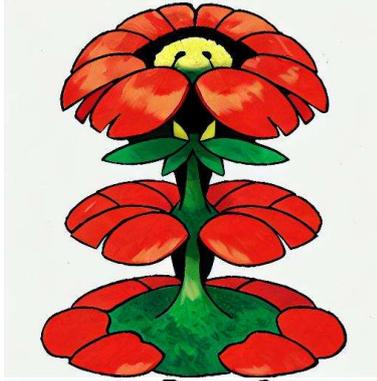 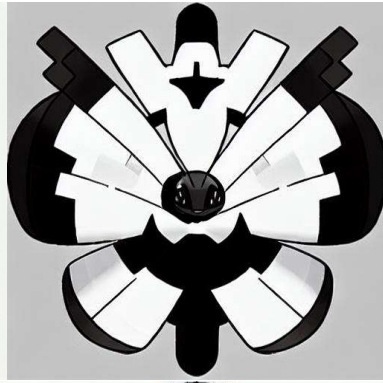 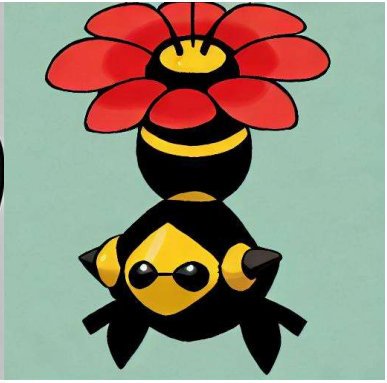
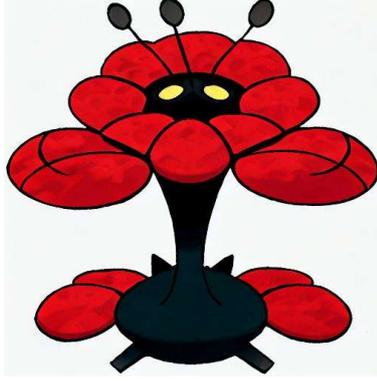 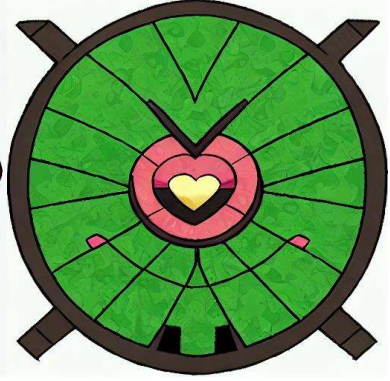

# Nitro Diffusion: Disney Style

Images Generated using the Nitro Diffusion Model in the style of Disney, for the book In My Garden [20]

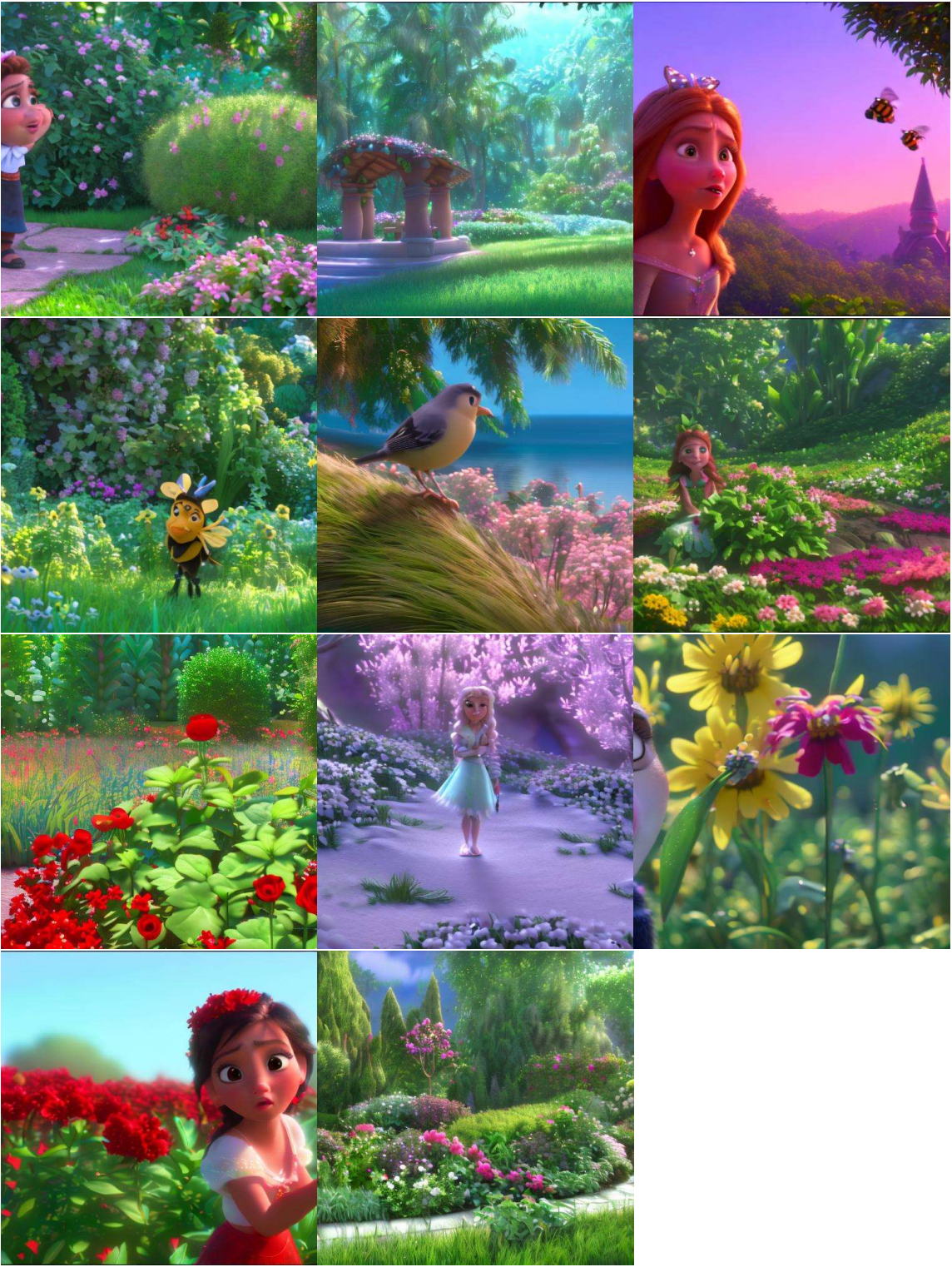